\documentclass{article}
\usepackage{ijcai22}
\usepackage{graphicx}                                                           
\usepackage{float} 
\usepackage{times}

\usepackage{soul}
\usepackage{url}
\usepackage[hidelinks]{hyperref}
\usepackage[utf8]{inputenc}
\usepackage[small]{caption}
\usepackage{caption}   

\usepackage{graphicx}
\usepackage{amsmath}
\usepackage{booktabs}
\usepackage{color}
\usepackage[linesnumbered,ruled,vlined]{algorithm2e}
\usepackage{algpseudocode}  
\usepackage{amsfonts}       
\usepackage{comment}
\usepackage[colorinlistoftodos]{todonotes}
\usepackage{subfigure}
\usepackage{CJK}
\usepackage{amsthm,amsmath,amssymb}
\usepackage{mathrsfs}
\usepackage{amssymb}
\usepackage{pifont}

\usepackage{hyperref}
\hypersetup{hidelinks,
	colorlinks=true,
	allcolors=black,
	pdfstartview=Fit,
	breaklinks=true}

\newcommand{\cmark}{\ding{51}}%
\newcommand{\xmark}{\ding{55}}%
\makeatletter

\title{PAnDR: Fast Adaptation to New Environments from Offline Experiences via Decoupling Policy and Environment Representations}

\author{
Tong Sang$^1$\textsuperscript{$\small{*}$}
\and
Hongyao Tang$^1$\footnote{Equal contribution.}\and
Yi Ma$^{1}$\and
Jianye Hao$^1$\thanks{Jianye Hao and Yan Zheng are the corresponding authors.} \and
Yan Zheng$^1$\textsuperscript{$\small{\dagger}$} \and
Zhaopeng Meng$^1$\and \\
Boyan Li$^1$\And
Zhen Wang$^2$
\affiliations
$^1$College of Intelligence and Computing, Tianjin University\\
$^2$School of Artiﬁcial Intelligence, Optics and Electronics (iOPEN)
    and School of \\ \ Cybersecurity, Northwestern Polytechnical University\\
\emails
\{2019218044, bluecontra, mayi, jianye.hao, yanzheng, mengzp, liboyan\}@tju.edu.cn, w-zhen@nwpu.edu.cn
}

\begin{document}

\maketitle

\begin{abstract}
Deep Reinforcement Learning (DRL) has been a promising solution to many complex decision-making problems. Nevertheless, the notorious weakness in generalization among environments prevent widespread application of DRL agents in real-world scenarios. Although advances have been made recently, most prior works assume sufficient online interaction on training environments, which can be costly in practical cases.
To this end, we focus on an \textit{offline-training-online-adaptation} setting,
in which the agent first learns from offline experiences collected in environments with different dynamics and then performs online policy adaptation in environments with new dynamics. In this paper, we propose Policy Adaptation with Decoupled Representations (PAnDR) for fast policy adaptation.
In offline training phase, the environment representation and policy representation are learned through contrastive learning and policy recovery, respectively. The representations are further refined by mutual information optimization to make them more decoupled and complete. With learned representations, a \textit{Policy-Dynamics Value Function} (PDVF) \cite{RaileanuGSF20PDVF} network is trained to approximate the values for different combinations of policies and environments from offline experiences. In online adaptation phase, with the environment context inferred from few experiences collected in new environments, the policy is optimized by gradient ascent with respect to the PDVF. Our experiments show that PAnDR outperforms existing algorithms in several representative policy adaptation problems.
\end{abstract}

\section{Introduction}
\label{sec:introduction}

Reinforcement learning (RL) has achieved great successes in many fields, e.g., Game Playing~\cite{BadiaPKSVGB20Agent57}, Software Testing~\cite{DBLP:conf/kbse/ZhengFXS0HMLSC19,DBLP:conf/icse/ZhengLXLMHL21}, Robotics~\cite{andrychowicz2020learning}, Industrial Supply Chain~\cite{MaHHLLTYLTM21Hierarchical} and so on.
Despite of the successes, conventional RL is well known to be lack of generalization ability~\cite{Charles2018generalization}.
To be specific, a policy well optimized in a specific environment can be catastrophically poor even in similar environments.
The poor generalization among environments and policies severely prevents more practical applications of RL algorithms.

In the literature on RL generalization~\cite{Kirk21SurveyGen},
many approaches like data augmentation~\cite{YaratsKF21DRQ} and domain randomization~\cite{PengAZA18S2R} are proposed to increase the similarity between training environments and the testing environments (i.e., the ones to generalize or adapt to).
These methods often require knowing the variations of environment and having the access to environment generation. 
In contrast, some other works aim to achieve fast adaptation in testing environments without the need for such access and knowledge.
Gradient-based Meta RL, e.g., MAML~\cite{FinnAL2017MAML}, aims to learn a meta policy which can adapt to a testing environment within few policy gradient steps.
Also in this branch, context-based Meta RL, e.g., PEARL~\cite{rakelly2019PEARL}, leverages a context-conditioned policy which allows to achieve adaptation by the generalization among contexts.
Useful context representation is learned in training environments to capture the variations; then the context of a testing environment can be inferred from few probed interaction experiences ~\cite{FuTHCFLL2021}.

Most prior works consider an \textit{online} training setting where the agent is allowed to interact with training environments arbitrarily within the interaction budget.
However, in real-world problems, online interaction is usually expensive while offline experiences are often available and relatively sufficient.
For example, in E-commerce, adapting the advertising strategies obtained from massive offline ads-users interaction data to a new user can effectively save the budgets of advertisers.
On this point, a recent work \cite{RaileanuGSF20PDVF}
studies fast adaptation under an \textit{offline-training-online-adaptation} setting,
where only offline experiences collected on training environments are available 
before online policy adaptation in new environments.
This is a more practical and general setting which is meanwhile more challenging,
since the policy learning and variation capture need to be conducted in an offline manner.
To solve this problem, Raileanu et al. \shortcite{RaileanuGSF20PDVF} propose 
an extended value function, called Policy-Dynamic Value Network (PDVF).
By definition, a PDVF additionally takes an environment representation and an policy representation as inputs, and evaluates the corresponding values.
In principle, the explicit representation allows the generalization of value estimation among environments and policies, which can be utilized to realize policy adaptation.
To be specific, given the offline experiences generated from different combinations of training policies and training environments,
the representations of environment and policy are learned from transitions and state-action sequences by dynamics prediction and policy recovery respectively.
Thereafter, a quadratic form of PDVF is approximated,
according to which the policy is optimized during online adaptation with the environment representation inferred from few interactions in the testing environment.
The first drawback is that the representations of environment and policy can have redundant and entangled information
since they are from the same interaction experiences of the corresponding combination.
It negatively affects the generalization conveyed by the representations which is vital to fast adaptation.
Another limitation is that the quadratic approximation of PDVF inevitably makes a sacrifice in expressivity for convenient closed-form optimization,
in turn this cripples the optimality of policy obtained during adaptation.

\begin{figure}
\centering
\includegraphics[width=0.4\textwidth]{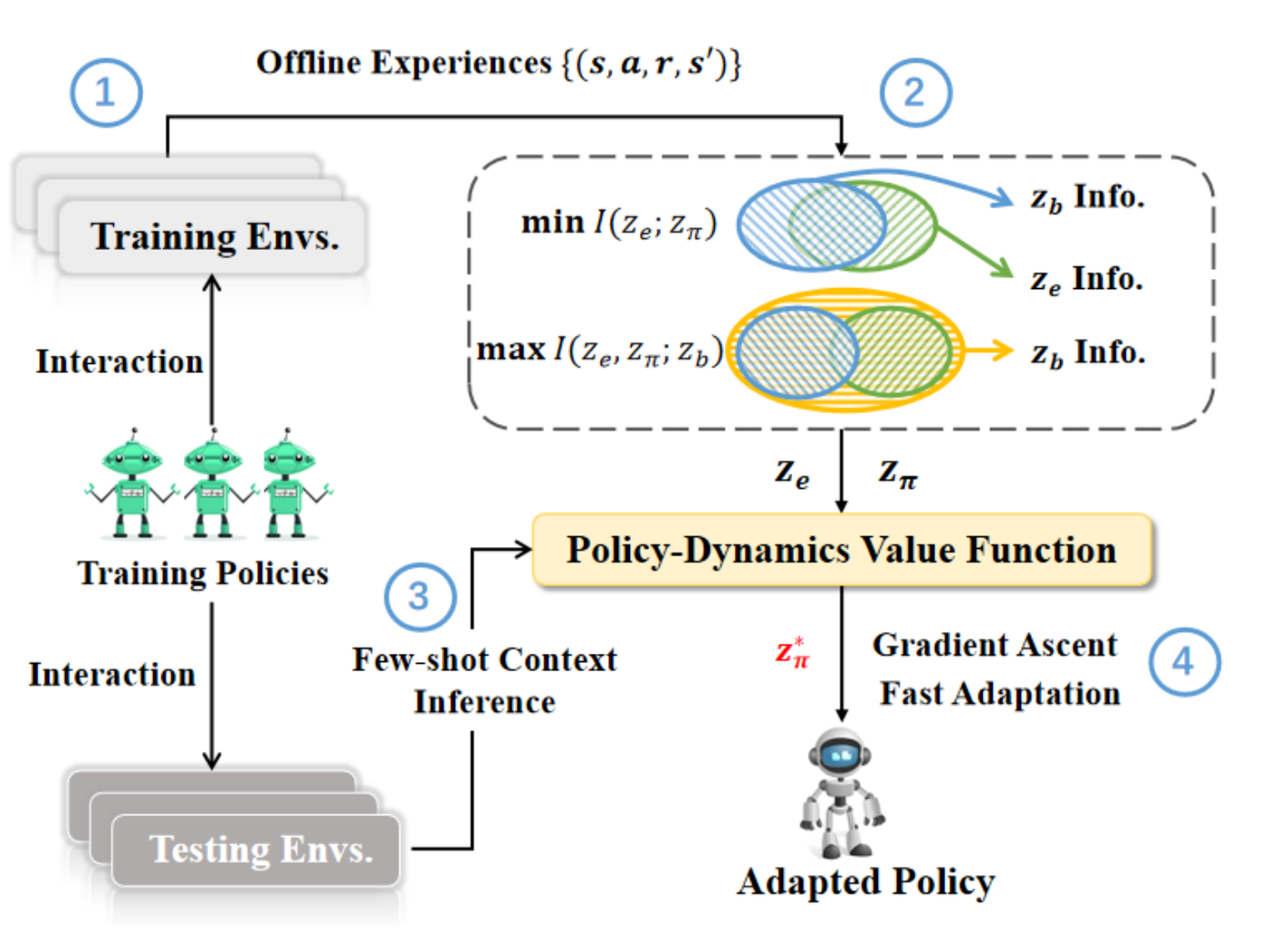}
\caption{A conceptual illustration of PAnDR for fast adaptation with offline experiences (best view in numbers).
With offline experiences collected by training policies and environments (\textcolor{blue}{\ding{172}}),
the representations for environment ($z_e$) and policy ($z_\pi$) are learned,
which are further improved by mutual information optimization with extra joint representation $z_b$  (\textcolor{blue}{\ding{173}}).
Conditioning on learned representations,
a PDVF is trained.
For adaptation, the representation of testing environment is inferred with few-shot online context (\textcolor{blue}{\ding{174}}),
based on which the policy (representation) is adapted according to PDVF (\textcolor{blue}{\ding{175}}).}
\label{fig: overall structure}
\end{figure}

In this paper, we focus on the offline-training-online-adaptation setting and follow the paradigm of \cite{RaileanuGSF20PDVF}.
We propose \textbf{P}olicy \textbf{A}daptatio\textbf{n} with \textbf{D}ecoupled \textbf{R}epresentations (PAnDR) for more effective fast adaptation of RL.
The conceptual illustration of PAnDR is shown in Fig.~\ref{fig: overall structure}.
Compared to transition-level dynamics prediction adopted by \cite{RaileanuGSF20PDVF}, we adopt trajectory-level context contrast for environment representation learning.
By this means, the variation is captured in an integral manner and the inter-environment difference is emphasized, thus offering more useful information and better robustness.
One can link our comparison here to that between reconstruction and contrastive learning, where the superiority of the latter has been widely demonstrated~\cite{He0WXG20MoCo,laskin2020curl,FuTHCFLL2021}.
Moreover, we propose a information-theoretic representation learning method to refine environment and policy representations for information decoupleness and
completeness.
This is achieved by mutual information minimization and maximization among environment and policy representations and their joint.
In addition, distinct to the quadratic form in \cite{RaileanuGSF20PDVF},
PAnDR leverages a typical multi-layer nonlinear neural network for the approximation of the PDVF. Then during online adaptation, the policy is optimized by gradient ascent. 
In this sense, PAnDR gets rid of such constraints to improve value approximation and policy adaptation.
In experiments, we evaluate PAnDR in a series of  fast adaptation tasks and analyze the contribution of each part of proposed methods.

We summarize our main contributions as follows:
1) We leverage trajectory-level contrastive learning and propose a novel mutual information based approach for better environment and policy representations.
2) We get rid of the constraint of prior quadratic approximation of PDVF and demonstrate the effectiveness of nonlinear approximation in improving online policy adaptation.
3) We show that PAnDR consistently outperforms representative existing algorithms in our experiments, sometimes with a large margin.

\section{Preliminaries}
\label{sec:preliminaries}
First, we briefly introduce the necessary preliminaries. A detailed discussion on related works is provided in Appendix~\ref{Ape:related_work}.

\subsection{Reinforcement Learning}
\label{sec:reinforcement learning}
Consider a Markov Decision Process (MDP) defined as $\mathcal{M} = \langle S, A, P, R, \gamma \rangle$, 
with the state space $S$, the action space $A$, the transition probability $P: S \times A \times S \to [0,1]$, the reward function $R: S \times A \to \mathbb{R}$ and the discount factor $\gamma \in [0,1)$.
An agent interacts with the MDP $\mathcal{M}$ by performing its policy $\pi:S \times A \to [0,1]$, 
generating the trajectories as $a_{t} \sim \pi(\cdot | s_{t})$, $s_{t+1} \sim P \left( \cdot \mid s_{t}, a_{t} \right)$ and $r_t = R \left(s_{t},a_{t} \right)$.
The value function of policy $\pi$ is defined as
$V^{\pi}(s)=\mathbb{E}_{\pi} \left[\sum_{t=0}^{\infty} \gamma^{t} r_{t} \mid s_{0}=s \right]$.
The objective of an RL agent is to optimize its policy to maximize the expected discounted cumulative rewards
$J(\pi) = \mathbb{E}_{s_{0} \sim \rho_{0} (\cdot)} [ V^{\pi}(s_0) ]$,
where $\rho_{0}$ is the initial state distribution.

\subsection{Mutual Information Optimization}
\label{sec:mutual information optimization}
Mutual Information (MI) is a widely used metric that estimates the relationship between pairs of variables in machine learning.
Given two random variables $X$ and $Y$, MI is a measure of dependency among them,
defined as $I(X; Y) = \mathbb{E}_{p(x,y)} \left[\frac{p(x, y)}{p(x)p(y)} \right]$,
with the joint probability distribution $p(x, y)$ and the marginal probability distributions $p(x),p(y)$.
MI estimation and optimization are fundamental techniques in representation learning, with Information Bottleneck~\cite{AlemiFD017IB} as a representative.

In most problems, it is intractable to compute MI directly since the exact distributions of interested variables are not known.
Thus, approximate estimation of MI is usually resorted to, which is non-trivial especially in high-dimensional space.
For the general purpose of MI optimization, lower-bound and upper-bound approximations of MI are often used as surrogate objectives~\cite{PooleOOAT19MIBounds}.
Among the options, 
the variational approximation is a common way to compute lower bounds of MI:
\begin{equation*}
    \begin{aligned}
        I(X; Y) \geq - \mathbb{E}_{p(x)}[\log\ p(x)] + \mathbb{E}_{p(x, y)}[\log\ q(x\mid y)].
    \end{aligned}
    \label{eq:lower bound of MI}
\end{equation*}
The lower bound holds since the KL divergence is non-negative.
For the upper bound of MI, a recent work called Contrastive Log-ratio Upper Bound (CLUB)~\cite{cheng2020club} provides the following form:
\begin{equation*}
\begin{aligned}
I(X; Y) \leq \mathbb{E}_{p(x, y)}[\log q(x\mid y)] - \mathbb{E}_{p(x)p(y)}[\log q(x\mid y)],
\end{aligned}
\label{eq:upper bound of MI}
\end{equation*}
where $q(x\mid y)$ is approximated, e.g., by maximum likelihood estimation (MLE), of actual conditional distribution $p(x\mid y)$ when $p$ is not known.

\begin{figure*}[t]
\centering
\includegraphics[width=0.7\textwidth]{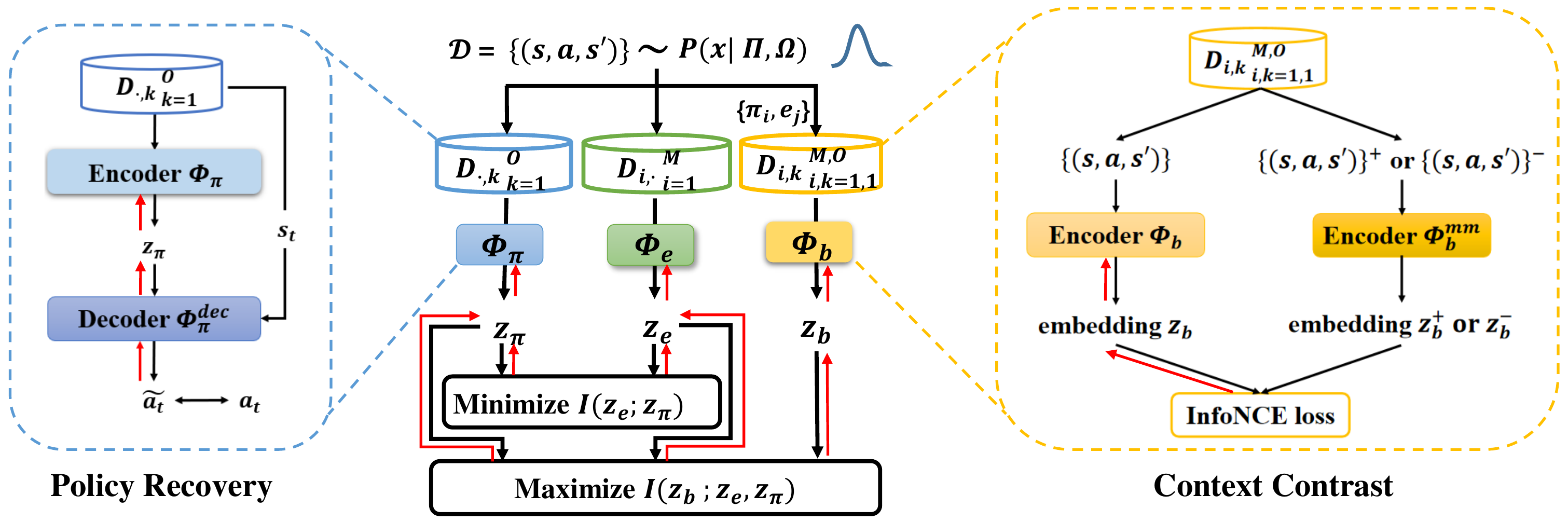}
\caption{Illustration of learning environment representation $z_e$ and policy representation $z_{\pi}$ from the offline experiences $\mathcal{D}$ collected by training environments $\Omega$ and policies $\Pi$.
With different views of $\mathcal{D}$, i.e., $D_{\cdot,k}$, $D_{i,\cdot}$ and $D_{i,k}$,
the representations for environment, policy and their joint ($z_b$) are obtained by encoders $\phi_e, \phi_{\pi}, \phi_b$ respectively, which are learned by policy recovery (\textit{left}) and context contrast (\textit{right}).
Thereafter,
the representations of environment and policy are refined through mutual information based optimization, 
i.e.,
$\min I(z_{e};z_{\pi})$ for representation decoupleness and $\max I(z_{b};  z_{e},z_{\pi})$ for representation completeness.
Gradient traces of model parameters are plotted in red.}
\label{fig:PAnDR_encoder}
\end{figure*}

\section{Methodology}
\label{sec:methodology}

In this section, we propose our framework for fast adaptation, named Policy Adaptation with Decoupled Representation (PAnDR).
We follow the paradigm of~\cite{RaileanuGSF20PDVF} and consider the offline-training-online-adaptation setting with a set of training environments $\Omega_{\text{train}} = \{\mathcal{M}_{i}\}_{i=1}^{M}$ and a set of testing environments $\Omega_{\text{test}} = \{\mathcal{M}_{j}\}_{j=1}^{N}$.
Given a set of training policies $\Pi_{\text{train}}=\{\pi_{k}\}_{k=1}^{O}$,
an offline experience buffer is established as $\mathcal{D}=\bigcup_{i=1,k=1}^{M,O} D_{i,k}$,
where $D_{i,k}$ contains the $s,a,r$-trajectories generated by $\pi_k$ in $\mathcal{M}_i$.
$M,N,O$ denote the sizes of corresponding set.
In this paper, we call a set of $(s,a,r)$ generated in an environment as the \textit{context} of it, and a set of $(s,a)$ collected by a policy as its \textit{behavior},
denoted by $c$ and $b$ respectively.

With offline experience buffer $\mathcal{D}$, PAnDR learns the representations of environment and policy through context contrast and policy recovery, respectively (Sec.~\ref{sec:self-supervised representation learning}). 
To eliminate the redundant information entangled in the two representations while retaining necessary information, we further introduce a novel approach by mutual information minimization and maximization to refine the representations (Sec.~\ref{sec:decoupling and coupling of representations}).
Thereafter, we train a nonlinear neural network to approximate PDVF conditioning on both environment and policy representations and introduce the online adaptation with gradient-based policy optimization (Sec.~\ref{sec:value learning and fast adaptation}).

\subsection{Self-supervised Representation Learning}
\label{sec:self-supervised representation learning}

The first part of PAnDR is to learn effective representations for environment and policy from offline experiences, replying on which the environment generalization and policy adaptation are carried out.
We learn these representations in a self-supervised manner via context contrast and policy recovery,
to capture the pivotal information of environment variation and policy behavior.
We introduce the details below.

\paragraph{Learning Environment Representation via Context Contrast.}
\label{sec:environment representation learning}

By following the general idea of self-supervised Contrastive Learning~\cite{He0WXG20MoCo}, 
we make use of the contrast of contexts collected in different training environments to extract context embedding as environment representation.
For a training environment $\mathcal{M}_i$, we use the contexts sampled from $D_{i,\cdot}=\bigcup_{k=1}^{O} D_{i,k}$
as anchor sample $c_i$ and positive sample $c_i^{+}$;
in turn, the context sampled from the offline experiences generated in other environments, i.e., $D_{i^{\prime},\cdot}$ for any $i^{\prime} \ne i$, is taken as negative sample $c_i^{-}$.

With the contrastive context samples described above, we
train a context encoder $\phi_{e}$ that maps context to embedding $z_e$
by ensuring the representation of similar sample pairs (anchors and their positives) stay close to each other while dissimilar ones (anchors and their negatives) are far apart.
Formally, the context encoder is optimized with InfoNCE loss similar to \cite{He0WXG20MoCo,laskin2020curl}:
\begin{equation}
\begin{aligned}
\mathcal{L}_{\text{CC}}(\phi_{e},W) = - \mathbb{E}_{c,c^{+},\{c^{-}\} \sim \mathcal{D}}
 & \Big[ 
 \ z_e^{\top} W z_e^{+} \\
- \log \big( \exp(z_e^{\top} W z_e^{+}) + & \sum_{\{c^{-}\}}  \exp(z_e^{\top} W z_e^{-}) \big)
\Big],
\end{aligned}
\label{eq:CL loss of env_encoder}
\end{equation}
where $W$ denotes the parameter matrix updated along with $\phi_e$.
Note that $z_e$ is differentiable with respect to $\phi_e$;
while $z_e^{+},z_e^{-}$ are the embeddings for positive and negative context samples calculated by a momentum averaged version of $\phi_{e}$.

Compared to learning environment representation by transition-level dynamics prediction as adopted in~\cite{RaileanuGSF20PDVF},
PAnDR captures environment variations in an integral manner and emphasizes inter-environment differences by context-based contrastive learning.
Thus, the environment representation can be more effective and robust, as similarly demonstrated in~\cite{FuTHCFLL2021}.

\paragraph{Learning Policy Representation via Policy Recovery.}
\label{sec:policy representation learning}

For a training policy $\pi_k$, we learn the embedding of the behavior data sampled from $D_{\cdot, k}=\bigcup_{i=1}^{M} D_{i,k}$ as policy representation.
As shown in the left of Fig.~\ref{fig:PAnDR_encoder}, a behavior $b$ is fed in the behavior encoder $\phi_{\pi}$,
producing the embedding $z_{\pi}$.
With the purpose of making $z_{\pi}$ contain the essential policy information,
we utilize a $z_{\pi}$-conditioned policy decoder $\phi_{\pi}^{\text{dec}}$ which takes as input a state $s$ and the embedding $z_{\pi}$, then deterministically outputs a prediction of the actual action made by $\pi$.
We call this \textit{policy recovery} (or Behavioral Cloning~\cite{GroverAGBE18MAPR}).
With the offline experiences,
$\phi_{\pi}, \phi_{\pi}^{\text{dec}}$ are updated by minimizing the $l_{2}$ prediction loss:
\begin{equation}
\mathcal{L}_{\text{PR}}(\phi_{\pi},\phi_{\pi}^{\text{dec}}) = \mathbb{E}_{s,a,b \sim \mathcal{D}} \big[ \|a -  \phi_{\pi}^{\text{dec}} (s, \phi_{\pi}(b) ) \|_{2}^{2} \big].
\label{eq:recovery loss of policy_encoder}
\end{equation}

\subsection{MI-based Representation Refinement}
\label{sec:decoupling and coupling of representations}

A consequent concern on the representations obtained above is the redundant and entangled information in them. This is because the representations for both environment and policy are learned from the offline experiences.
The entangled information exists in both representations will negatively affect the generalization among environments and policies,
thus hinders value approximation and policy adaptation in turn.

To address this problem, we propose a novel information-theoretic approach to refine the learned representations.
The key idea is 1) to eliminate the redundant information contained in environment and policy representations for the decoupleness,
while 2) to ensure the essential information in their joint representation are completely retained in the two representations.
This is achieved by two ways of mutual information (MI) optimization, as illustrated in the middle part of Fig.~\ref{fig:PAnDR_encoder}.

\paragraph{MI Minimization for Decoupleness.}
\label{sec:mi_min_for_decoupleness}

Firstly, for the decoupleness,
we are interested in reducing the entangled information in the representations of environment and policy.
To be specific,
we minimize the MI between the embeddings $z_e$ and $z_{\pi}$ that are obtained from context and behavior data sampled from the offline experiences,
i.e., $\min I(z_{e};z_{\pi})$.
We then convert the MI minimization into minimizing the CLUB upper bound of MI, as introduced in Sec.~\ref{sec:mutual information optimization}.
With the approximation conditional distribution $q_{\psi_{1}}$ parameterized by $\psi_{1}$,
our loss function for representation decoupleness (RD) is formulated as:
\begin{equation}
\begin{aligned}
\mathcal{L}_{\text{RD}} (\phi_{e},\phi_{\pi})  = \ & \mathbb{E}_{p(z_{e},z_{\pi})} \big[ \log q_{\psi_{1}}(z_{e}\mid z_{\pi}) \big] \\ 
- & \mathbb{E}_{p(z_{e})p(z_{\pi})} \big[\log q_{\psi_{1}}(z_{e}\mid z_{\pi}) \big].
\end{aligned}
\label{eq: min MI loss}
\end{equation}
Concretely, as to any $D_{i,k}$, 
we use the embedding pairs of $c_i, b_k \sim D_{i,k}$ for the samples from joint distribution $p(z_{e},z_{\pi})$;
and use those of $c_i \sim D_{i, \cdot}, b_k \sim D_{\cdot,k}$ for the marginal distributions $p(z_{e}),p(z_{\pi})$.
Note that $\psi_{1}$ is not updated by $\mathcal{L}_{\text{RD}}$, while the approximate conditional distribution $q_{\psi_{1}}$ is trained through MLE by minimizing the following loss:
\begin{equation}
\begin{aligned}
\mathcal{L}_{\text{MLE}} (\psi_{1})  = & - \mathbb{E}_{p(z_{e},z_{\pi})} \big[ \log q_{\psi_{1}}(z_{e}\mid z_{\pi}) \big], 
\end{aligned}
\label{eq:q_theta_mle}
\end{equation}
where the joint distribution is sampled in the same way as in Eq.~\ref{eq: min MI loss}.
The updates defined in Eq.~\ref{eq: min MI loss} and Eq.~\ref{eq:q_theta_mle} are performed iteratively:
$q_{\psi_{1}}$ approximates the conditional distribution with respect to the embeddings given by $\phi_{e},\phi_{\pi}$;
and in turn $q_{\psi_{1}}$ is used in the estimation of CLUB.
We refer the readers to the original paper of CLUB~\cite{cheng2020club} for better understanding.

\paragraph{MI Maximization for Completeness.}
\label{sec:mi_max_for_completeness}

Secondly, we expect the information contained in the joint representation of environment and policy can be completely retained in both environment and policy representations.
This is also necessary especially when the decoupleness described above is required,
in case that the vital information of environment variation or policy behavior is overly reduced during MI minimization.
This is achieved by maximizing the MI between the concatenation of the embeddings $z_e,z_{\pi}$ and a target environment-policy joint representation $z_b$ which contains complete information, i.e., $\max I(z_e,z_{\pi};z_b)$.
We resort to extracting the target representation $z_b$ also from the offline experiences.
Similar but distinct to the environment representation, $z_b$ is learned by context-behavior joint contrast,
i.e., only the context-behavior data sampled from the same $D_{i,k}$ are regarded as mutually positive samples.
We omit the similar formulations here and an illustration is shown in the right part of Fig.~\ref{fig:PAnDR_encoder}.
Thereafter,
we maximize the variational lower bound of $I(z_e,z_{\pi};z_b)$, i.e., minimize the loss function for representation completeness (RC) below:
\begin{equation}
\begin{aligned}
    \mathcal{L}_{\text{RC}}(\phi_{e}, \phi_{\pi}, \psi_{2}) =
    - \mathbb{E}_{p(z_{e},z_{\pi},z_{b})} \big[\log q_{\psi_{2}}(z_{b}|z_{e},z_{\pi}) \big],
\end{aligned}
\label{eq:max_mi}
\end{equation}
where $q_{\psi_{2}}$ is the variational distribution parameterized by $\psi_{2}$,
and the joint distribution are sampled by $c_i,b_k \sim D_{i,k}$.
Note that the negative entropy term $- \mathcal{H}(z_b) = \mathbb{E}_{p(z_b)}[\log p(z_b)]$ irrelevant to the optimization of $z_{e},z_{\pi}$ is neglected in Eq.~\ref{eq:max_mi}.

Therefore, combining the self-supervised learning losses and the MI-based refinement losses, the total loss function for the representations of environment and policy is:
\begin{equation}
    \mathcal{L}_{\text{Total}}(\phi_e, \phi_{\pi}) = \mathcal{L}_{\text{CC}}
    + \mathcal{L}_{\text{PR}}
    + \alpha  \mathcal{L}_{\text{RD}}
    + \beta \mathcal{L}_{\text{RC}},
\label{eq:final loss of encoders}
\end{equation}
where $\alpha, \beta$ are the balancing weights.
The pseudocode of overall representation training is in Alg.~\ref{Alg:Training Encoders} in Appendix.

\subsection{Value Approximation and Policy Adaptation}
\label{sec:value learning and fast adaptation}

So far, we have introduced the training process of environment and policy representations on the offline experiences.
With the trained representations,
we are ready to introduce the approximation of PDVF~\cite{RaileanuGSF20PDVF} and the policy optimization during online adaptation.

\paragraph{Value Function Network Approximation.}
\label{sec:value function approximation}

As described in Section \ref{sec:introduction}, to evaluate the values of different environment-policy configurations,
PDVF additionally conditions on the representations of environment and policy.
Distinct from the quadratic approximation adopted in ~\cite{RaileanuGSF20PDVF}, 
PAnDR leverages a typical multi-layer nonlinear neural network, denoted by $\mathbb{V}_{\theta} (s,z_{e},z_{\pi})$ with parameter $\theta$.
This setting frees the function expressivity constrained by quadratic form.
We suggest that this is important because the value approximation for multiple environments and policies is much more difficult than the conventional situation (e.g., $V^{\pi}$).
Achieving better value approximation is crucial to the following policy adaptation.

For any offline experience $D_{i,k}$,
we can obtain the corresponding representations $z_e,z_{\pi}$,
conditioning on which
$\mathbb{V}_{\theta}$ is trained by Monte Carlo method~\cite{SuttonB98}:
\begin{equation}
    \mathcal{L}_{V}(\theta) = \mathbb{E}_{s,G(s),c,b \sim \mathcal{D}} \big[ \big( G(s) - \mathbb{V}_{\theta} (s,z_{e},z_{\pi} ) \big)^2 \big], 
\label{eq:loss of value function}
\end{equation}
where $G(s)=\sum_{t=0}^{T} \gamma^{t} r_t$ letting $s_0=s$ is the Monte Carlo return.
The pseudocode is provided in Alg. \ref{Alg:Training Policy-Dynamics Value Function} in Appendix.

\begin{figure*}
\centering
\subfigure[Spaceship-Charge]{
\includegraphics[width=0.23\textwidth]{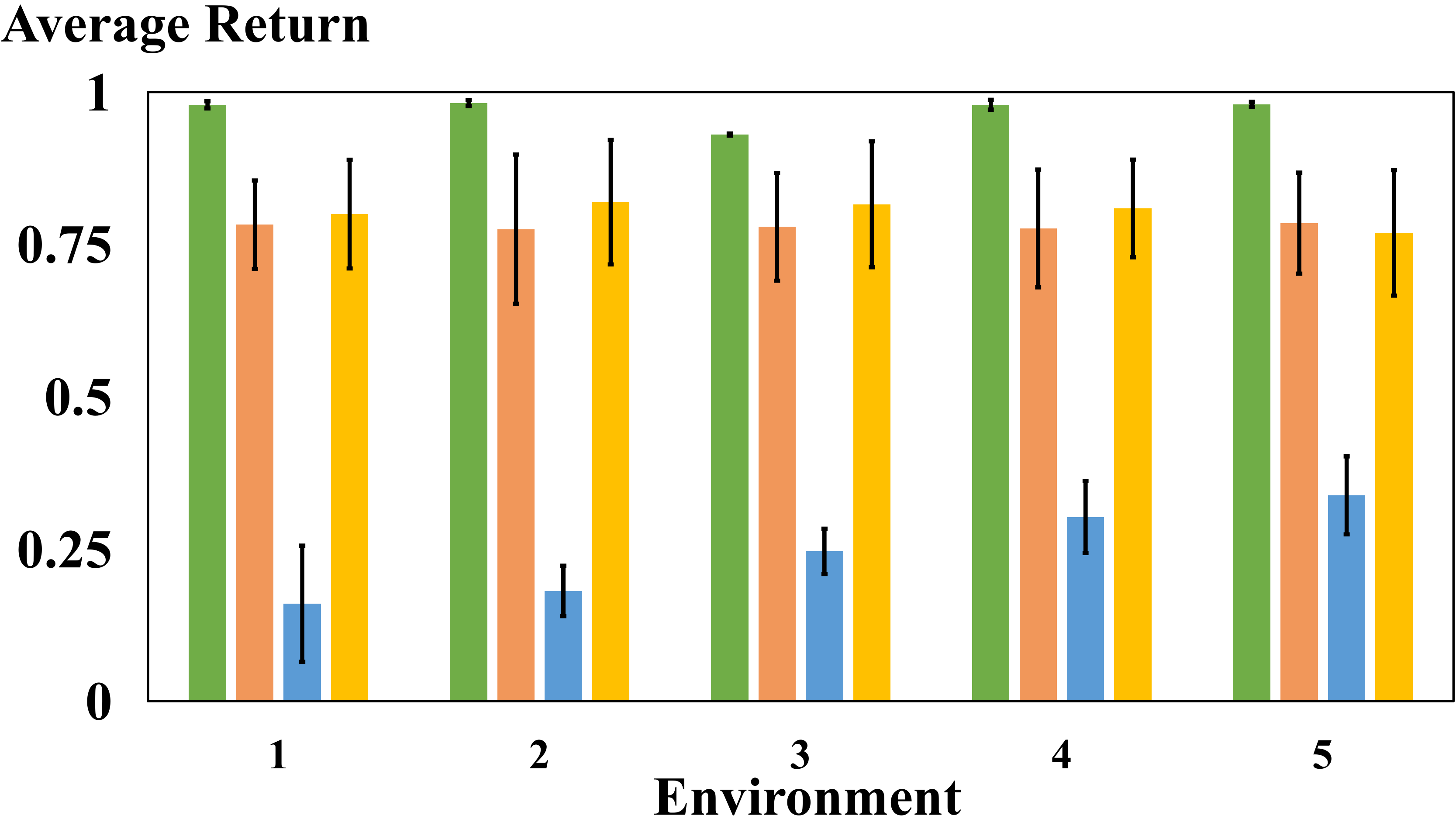}}
\subfigure[Ant-Wind]{
\includegraphics[width=0.23\textwidth]{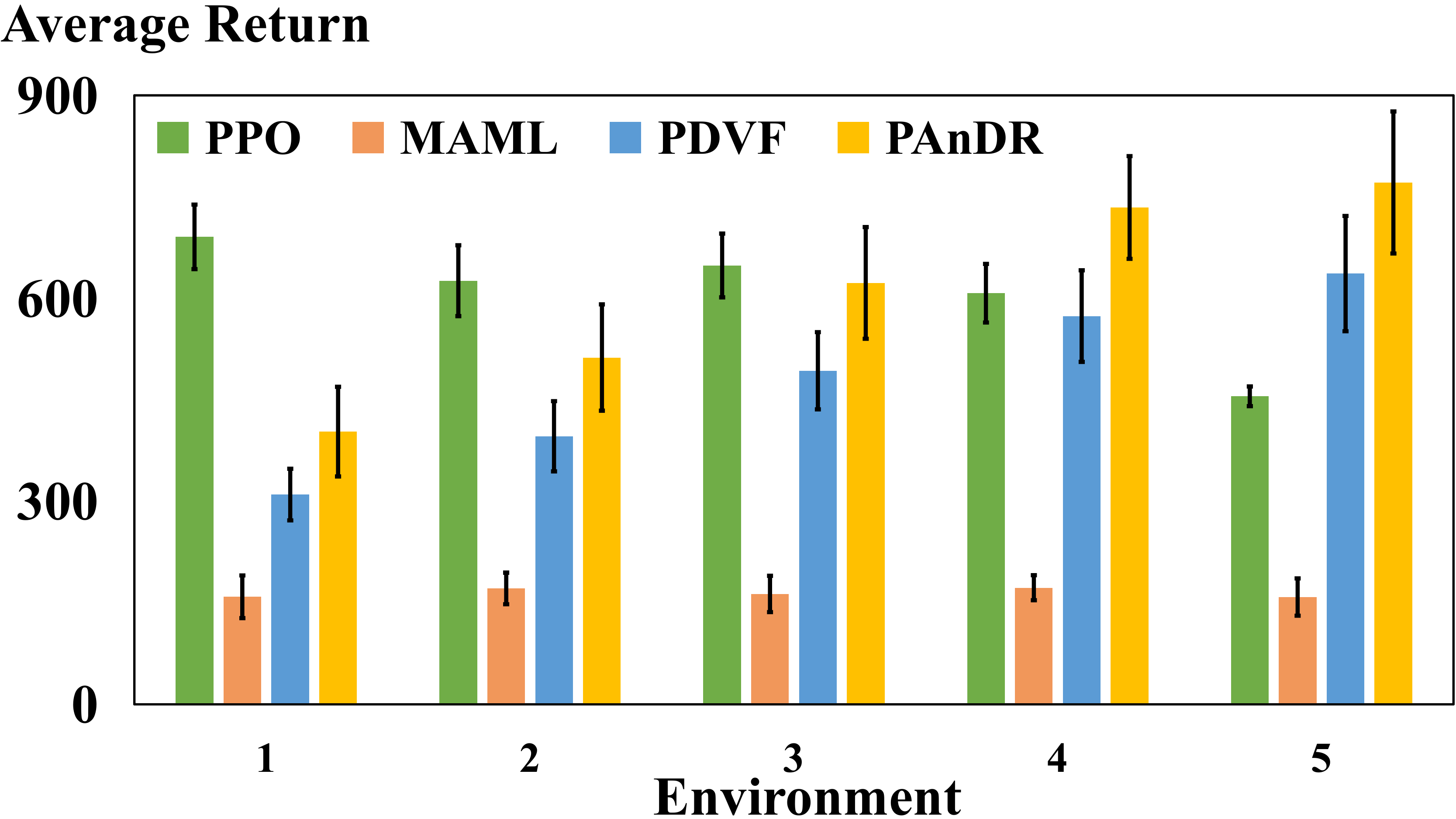}}
\subfigure[Swimmer-Fluid]{
\includegraphics[width=0.23\textwidth]{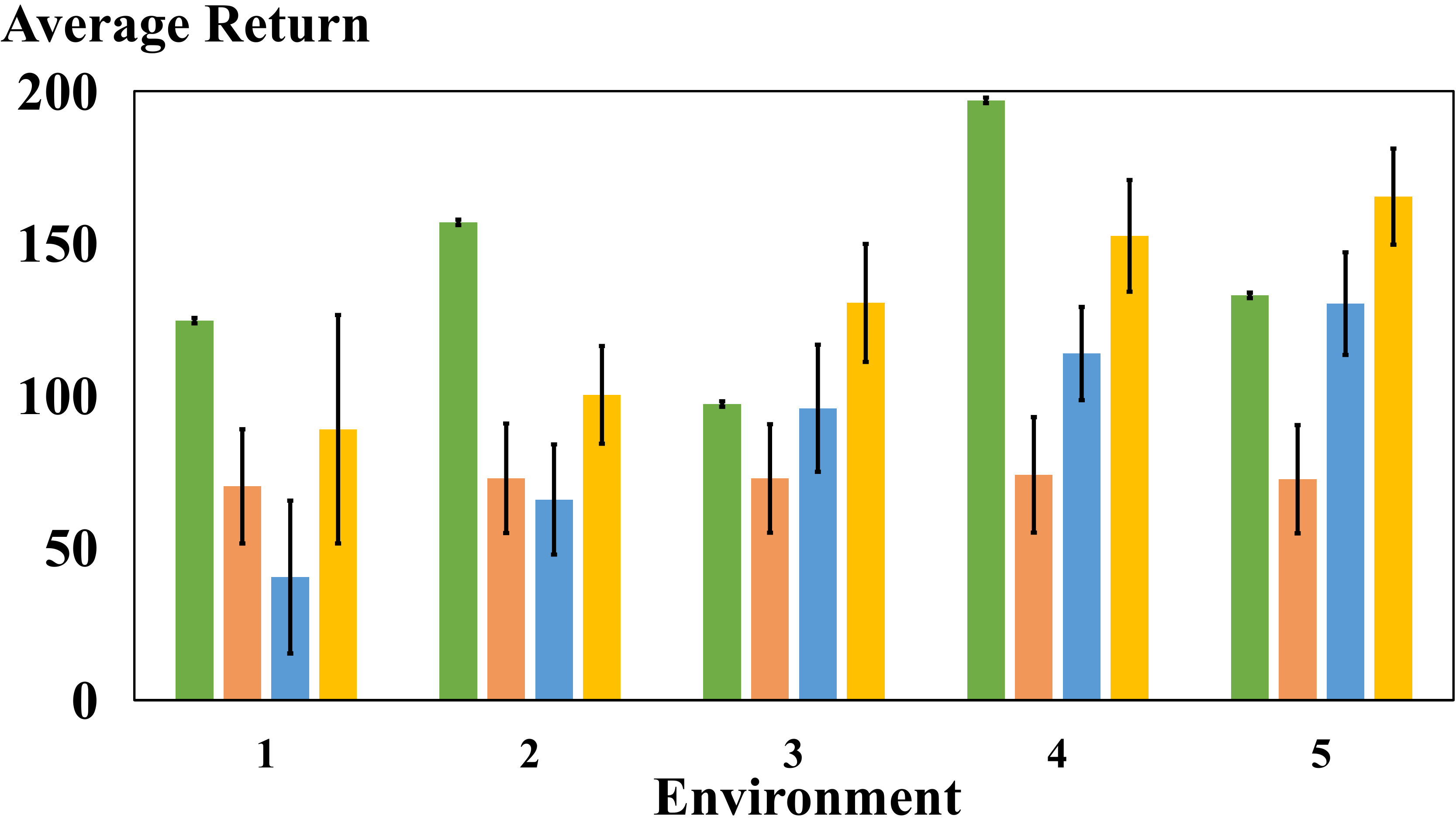}}
\subfigure[HalfCheetah-Mass]{
\includegraphics[width=0.23\textwidth]{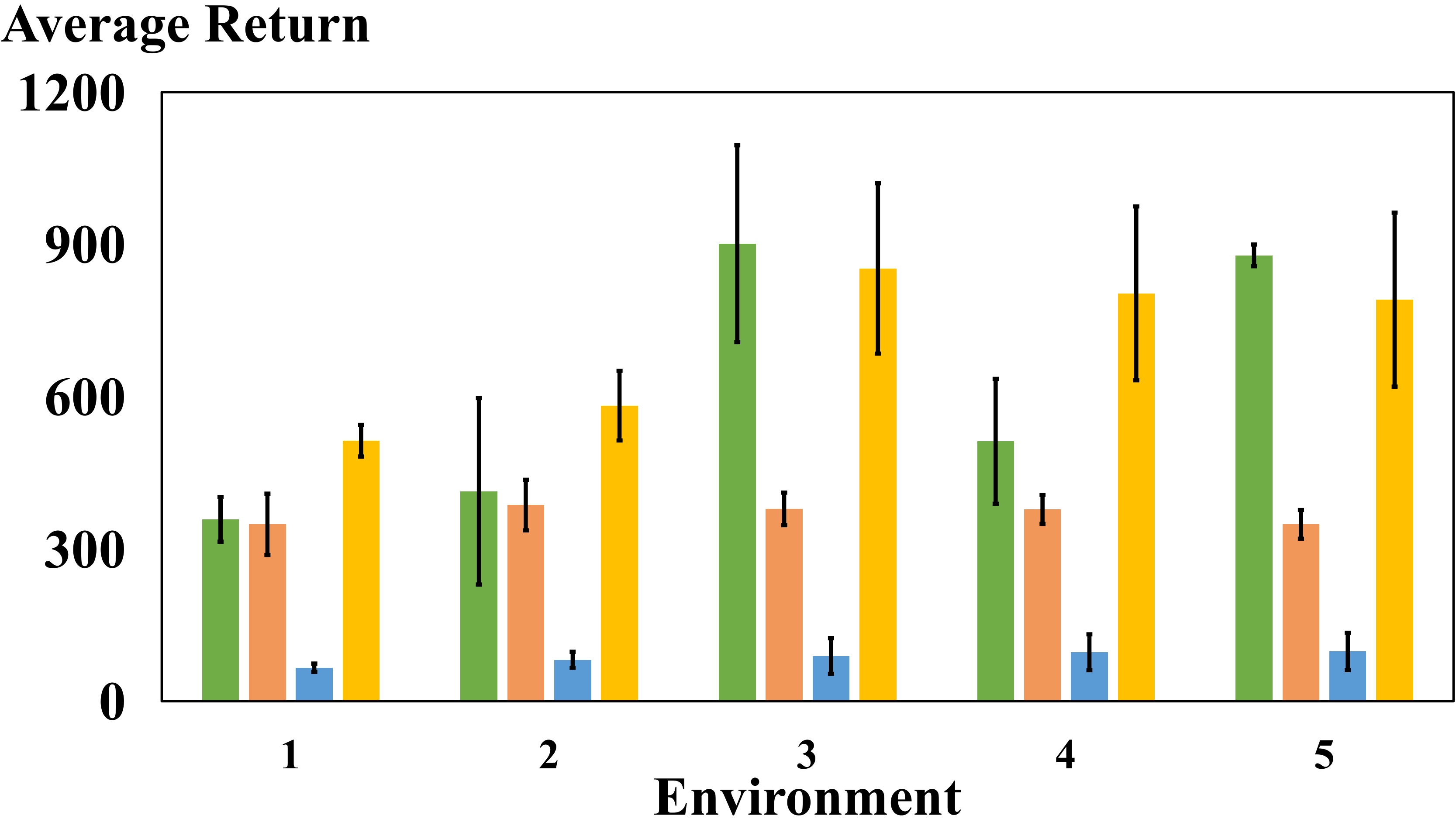}}
\vspace{-0.3cm}
\caption{Adaptation performance of different algorithms in four domains. Experimental results of algorithm generalization in new environment. Our results are even better than PPO in some environments, which proves that our method has a better generalization ability by learning in the training environment. For the PDVF algorithm, we use the source code provided by the authors.}
\label{fig:new env test}
\end{figure*}

\paragraph{Policy Adaptation via Representation Gradient Ascent.}
\label{sec:policy_adaptation}

After training $\mathbb{V}_{\theta}$ network well, in principle we can evaluate the values for any environment-policy configuration to a certain degree.
For online adaptation,
we first sample a training policy from $\Pi_{\text{train}}$ and use it to interact with the testing environment for a few steps.
Then the embedding of testing environment $z_{e}^{\text{on}}$ can be inferred with the collected context.
Next, starting from an arbitrary policy representation $z_{\pi}^{0}$ (randomly initialized by default),
by gradient ascent (GA), we can optimize the policy representation along the direction of increasing values of $\mathbb{V}_{\theta}$ with learning rate $\eta$:
\begin{equation}
    z_{\pi}^{t+1}= z_{\pi}^{t} + \eta \nabla_{z_{\pi}^{t}} \mathbb{V}_{\theta} (s,z_{e}^{\text{on}}, z_{\pi}^{t}).
\label{eq:gradient ascent of policy embedding}
\end{equation}
After a certain number of GA,
the obtained policy representation 
can be used to interact with the testing environment with the help of the $z_{\pi}$-conditioned policy decoder $\phi_{\pi}^{\text{dec}}$ (described in Sec.~\ref{sec:policy representation learning}),
which is trained offline to predict policy's action.
We adopt the best-performing policy representation as $z_{\pi}^{\ast}$ encountered during the GA process for our adapted policy in our experiments.

It is worth noting that the GA optimization itself is zero-shot, i.e.,
without the overhead of interacting with the testing environment. 
The pseudocode is in Alg.~\ref{Alg:Adaptation to New Environment} in Appendix.

\begin{table}
	\centering
	\scalebox{0.7}{
		\begin{tabular}{c|cc}
			\toprule
			Domain & Train/Test Env. No. & Variation\\
			\midrule
			\textbf{Spaceship-Charge} & 15 / 5 & The strength of electric field \\
			\textbf{Ant-Wind} & 15 / 5  & Direction of wind \\
            \textbf{Swimmer-Fluid} & 15 / 5 & Direction of the fluid \\
            \textbf{HalfCheetah-Mass} & 8 / 5 & Joint damping coefficient \\
			\bottomrule
		\end{tabular}
	}
\vspace{-0.1cm}
\caption{Domains in our experiments for fast adaptation. }

\label{table:Env Setting}
\end{table} 

\section{Experiment}
\label{sec:Experiment}

In our experiments, we mainly aim at investigating the following questions:
(\textbf{Q1}) Whether PAnDR outperforms existing algorithms for fast adaptation?
(\textbf{Q2}) How do contrastive environment representation and MI-based refinement contribute to the performance of PAnDR?
(\textbf{Q3}) Whether nonlinear neural approximation of PDVF and gradient-based policy adaptation improve the quadratic counterpart?

\subsection{Setup}
\label{Experiment: Setup}

We conduct experiments in four continuous control domains.
All the experimental details
can be found in Appendix~\ref{Ape:Additional Experimental Details}.

\paragraph{Domains and Variations.}
Table \ref{table:Env Setting} shows the fast adaptation domains used in our experiments,
where the first 3 domains are from \cite{RaileanuGSF20PDVF} and the last one is from \cite{lee2020CADM}.
In each domain, we have a variety of environments controlled by corresponding variations.
These variations determine the dynamics and reward functions of environment.
Therefore, fast adaptation is expected to be achieved against these changes.
The number of training and testing environments are provided in Table \ref{table:Env Setting}.

\paragraph{Offline Experiences and Adaptation Budget.}
We follow the same data setting as \cite{RaileanuGSF20PDVF}.
For offline experiences, a policy is optimized by PPO~\cite{schulman2017PPO} in each training environments.
The obtained policies are training policies that serve as collectors.
In general, such collector policies can be obtained from other ways.
For each domain, 50-episode interaction trajectories are collected by each combination of training environment and policy, e.g., $15*15*50$ episodes of experience for Ant-Wind.
For online adaptation,
50 steps of interaction are collected in the testing environment by one of the training policies.
For PAnDR, 100 steps of GA is used by default.

\paragraph{Evaluation Criteria.}
The main evaluation criterion is \textit{adaptation performance},
calculated as the undiscounted cumulative rewards of the policy obtained after online adaptation.
Moreover, for PAnDR, we analyze the evolvement of adaptation performance against the number of GA adaptation steps (in Sec.~\ref{Experiment:Ablation}).
The results reported in our experiments are means and error bars of a standard deviation calculated with 7 independent trials.

\subsection{Performance Comparison}
\label{Experiment:comparison}

In this section,
we compare the adaptation performance between PAnDR and representative existing algorithms to answer \textbf{Q1} within the domains introduced above.

\paragraph{Baselines.}
We consider the following baselines.
1) PPO: We use the PPO policy which is trained online with $3e6$ steps in testing environments as a \textit{powerful} baseline.
2) MAML~\cite{FinnAL2017MAML}:
We use MAML as a representative of gradient-based Meta RL.
Note MAML (as well as most Meta RL algorithms) performs online meta training.
This means in principle it is \textit{not applicable} to the offline training experiences.
3) PDVF~\cite{RaileanuGSF20PDVF}: Since we propose PAnDR based on the same paradigm of PDVF,
we use PDVF as a closely related baseline.
Moreover, to our knowledge, PDVF is the state-of-the-art fast adaptation algorithm in offline-training-online-adaptation setting. Note that in this section, we use PDVF to denote the method name of~\cite{RaileanuGSF20PDVF} rather than the extended value function.
For all above baselines, we use the official source codes provided by PDVF paper~\cite{RaileanuGSF20PDVF}.
For PAnDR, we also implement upon these source codes with almost the same configuration of hyperparameters and structures for a valid comparison. 

\paragraph{(A1) Results.}

As shown in Fig. \ref{fig:new env test}, for the simple Spaceship domain, the overall performance of PAnDR is significantly better than PDVF and slightly outperfroms MAML. Further in the three more difficult MuJoCo domains, the leading edge of PAnDR is much more evident and consistent compared with the two baseline methods,
while MAML loses its competitiveness.
We consider that this is because the preferable representations obtained via self-supervised learning and the MI-based refinement are more effective and robust than those of PDVF. 
Moreover, with less-constrained approximation, 
the more accurate value estimation achieved improves the optimality of policy obtained during adaptation.


It is worth noting that, perhaps somewhat surprisingly, PAnDR shows competitive and even better performance when comparing with the powerful PPO baseline trained in the testing environments.
Instead of requiring millions of interactions to optimize a policy in the testing environments, PAnDR only needs limited interacting experiences for fast adaptation.
This further reveals the superiority and potential of PAnDR.

\begin{figure}[t]
\vspace{-0.5cm}
\centering
\subfigure[Ant-Wind]{
\includegraphics[width=0.22\textwidth]{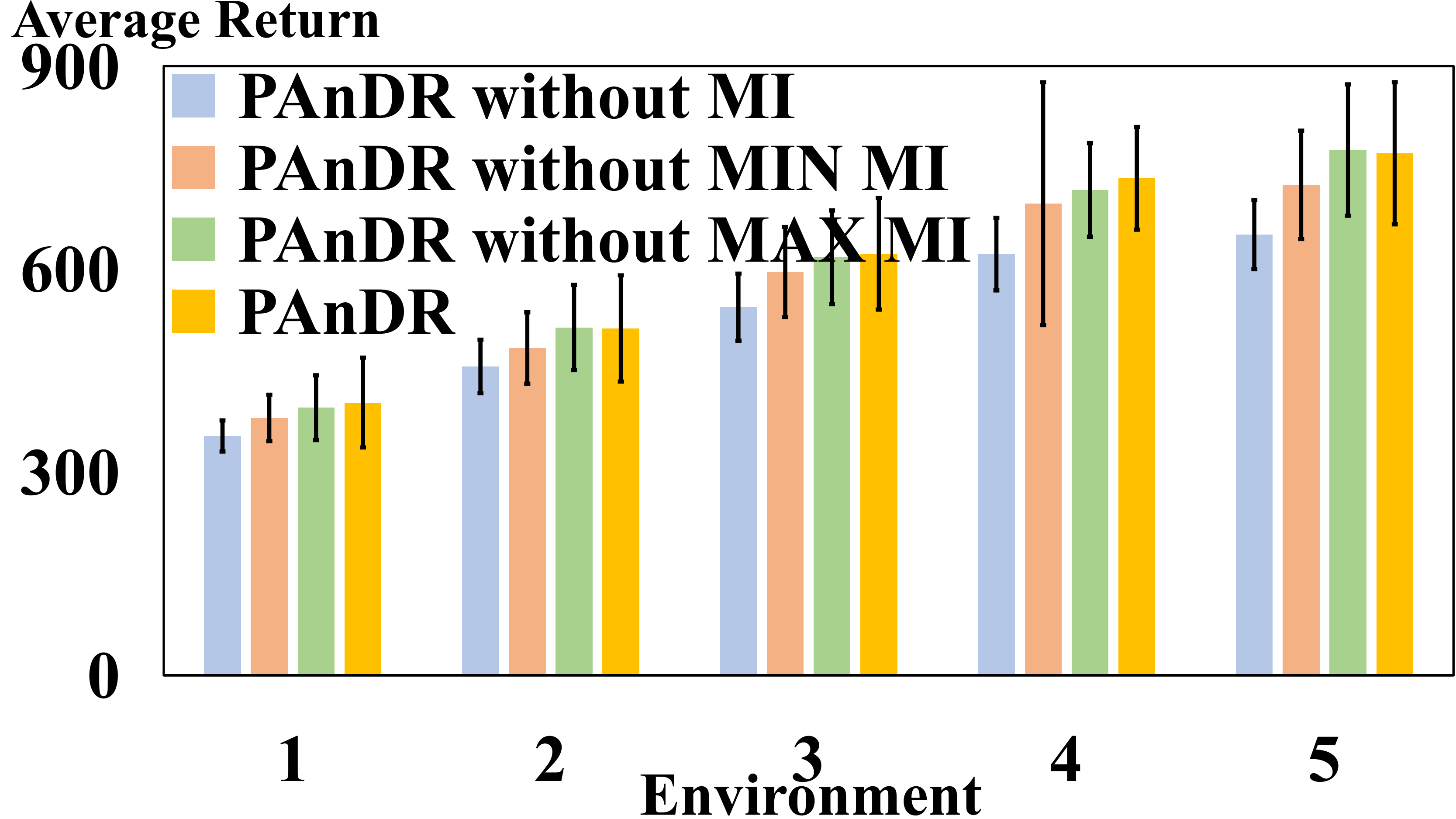}}
\subfigure[Spaceship-Charge]{
\includegraphics[width=0.22\textwidth]{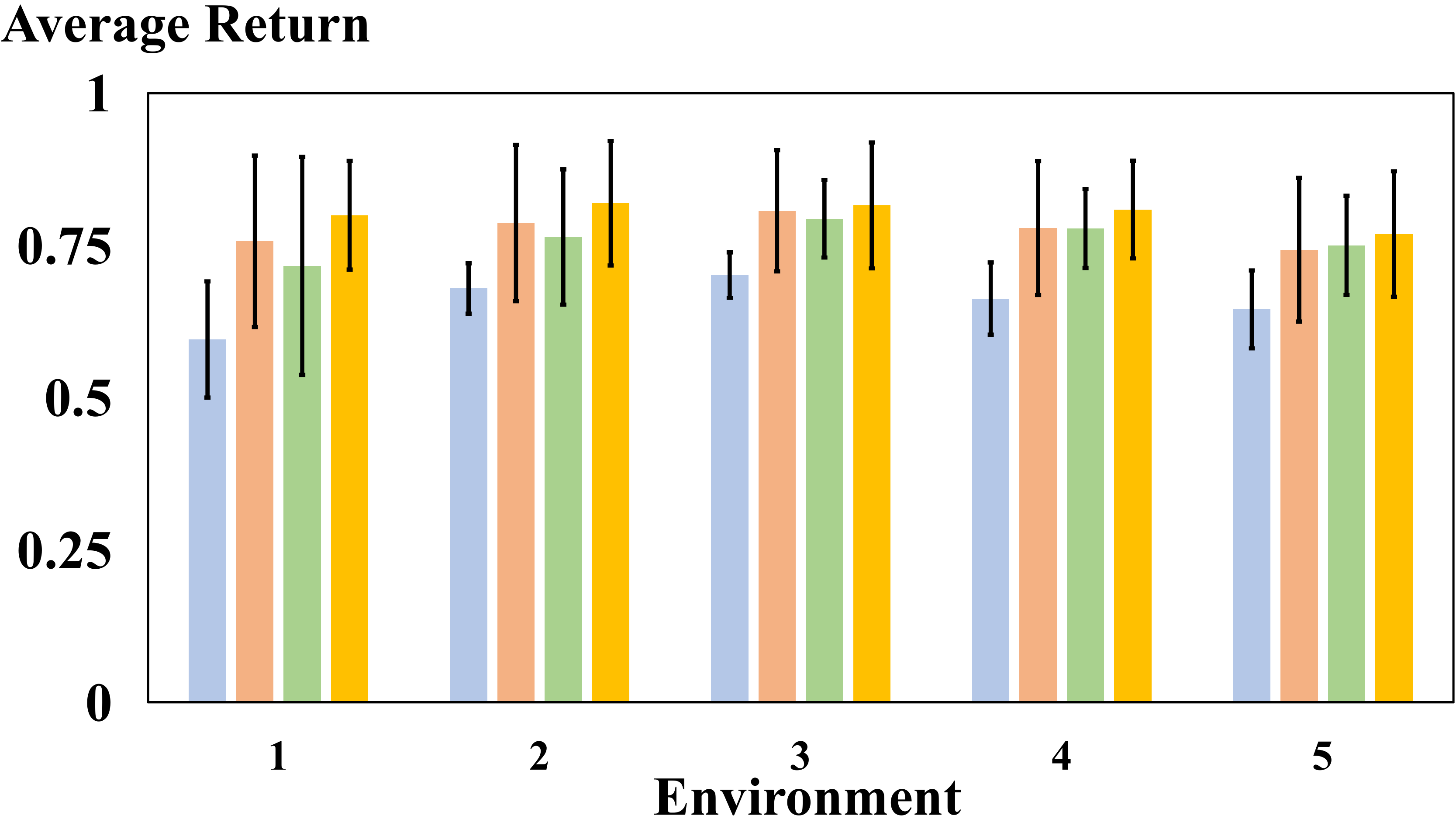}}
\vspace{-0.3cm}
\caption{Ablations for MI-based representation refinement.
}
\vspace{-0.2cm}
\label{fig:albation mi}
\end{figure}

\subsection{Ablation Study and Analysis}
\label{Experiment:Ablation}
In this section, we give the answers to \textbf{Q2} and \textbf{Q3}. 
We provide main results below, and the complete results along with detailed analysis are given in Appendix \ref{Ape:complete_exp_results}.

\paragraph{(A2-\uppercase\expandafter{\romannumeral1}) MI-based Representation Refinement.}

To verify the contribution of the MI minimizing and maximizing in representation refinement of PAnDR, we compare PAnDR with its several variants including: 1) PAnDR without MAX MI: removes MI maximizing refinement; 2) PAnDR without MIN MI: removes MI minimizing refinement; 3) PAnDR without MI: removes both MI maximizing and minimizing refinements.
As shown in Fig.\ref{fig:albation mi}, both MI minimizing and maximizing positively contribute to the improvement of PAnDR's overall performance.
It demonstrates that both MI-based refinement components are indispensable in PAnDR.

\paragraph{(A2-\uppercase\expandafter{\romannumeral2}) Context Contrast for Environment Representation.}

\begin{figure}[t]
\vspace{-0.5cm}
\centering
\subfigure[Ant-Wind]{
\includegraphics[width=0.22\textwidth]{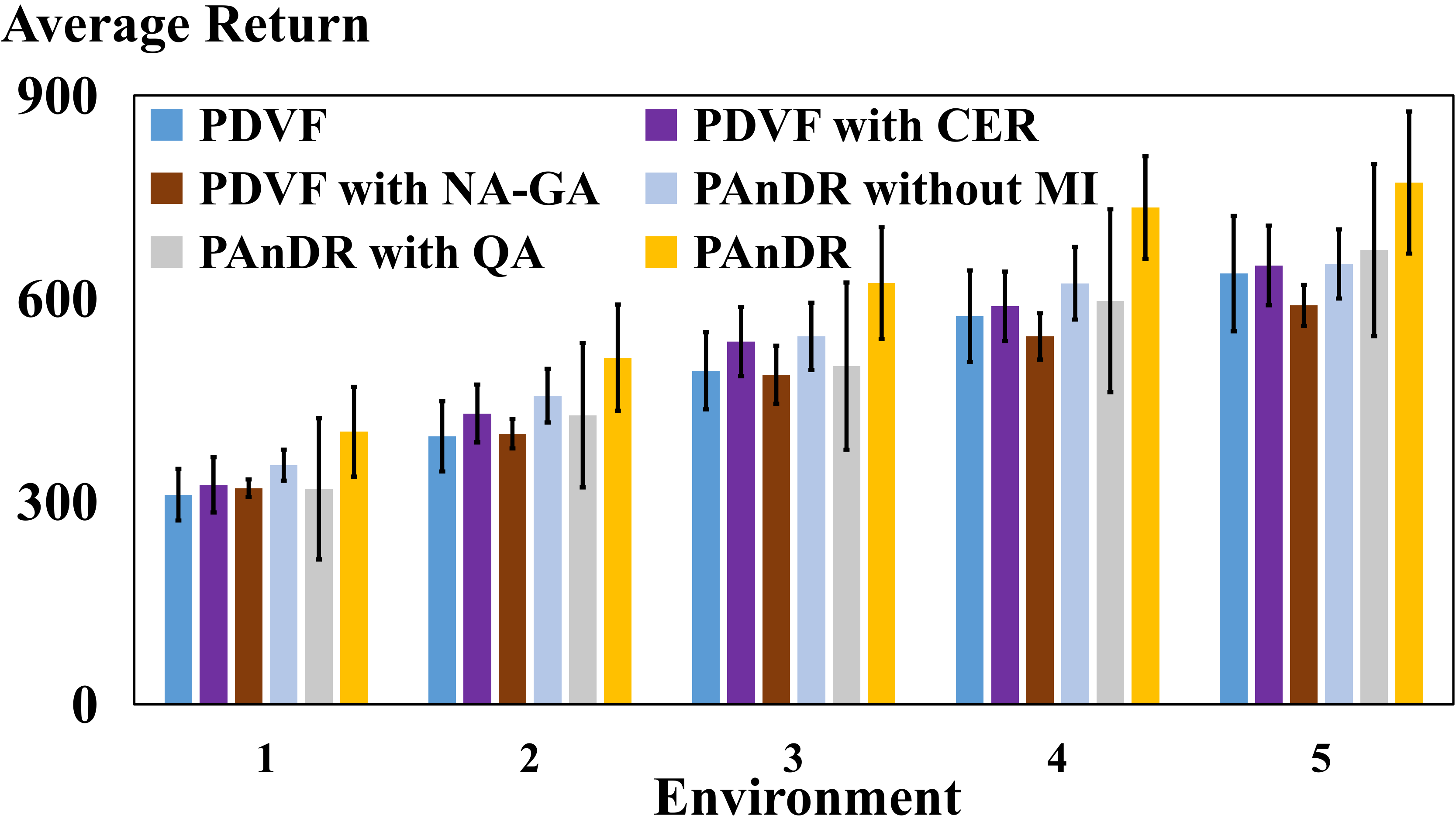}
}
\hspace{-0.2cm}
\subfigure[Spaceship-Charge]{
\includegraphics[width=0.22\textwidth]{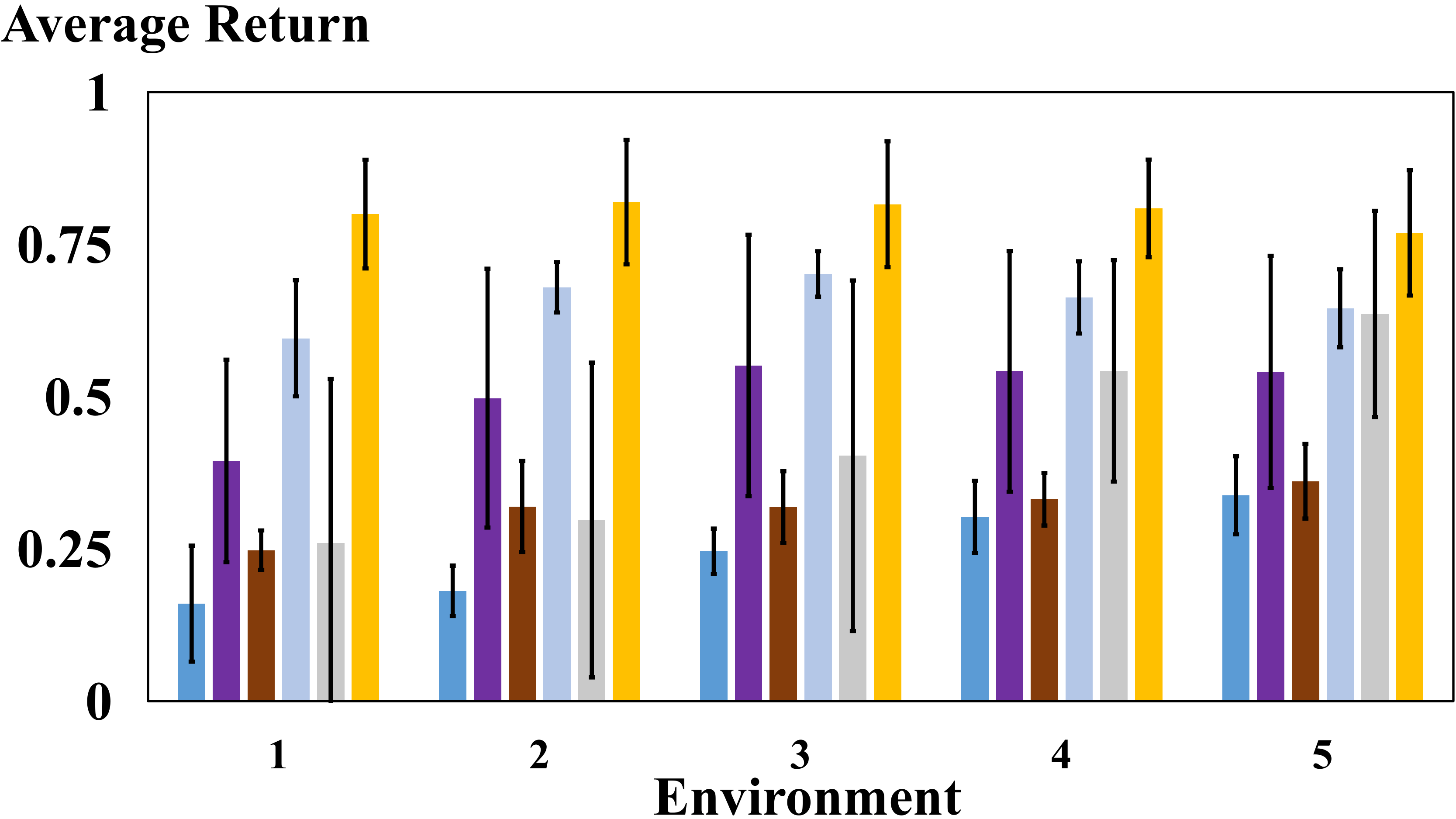}
}
\vspace{-0.2cm}
\caption{Efficacy analysis for contrast environment representation (CER) and neural approximation of PDVF with GA adaptation (NA-GA) in contrast to quadratic approximation (QA).}
\label{fig:Struction Ablation}
\end{figure}

Fig.\ref{fig:Struction Ablation}, presents the efficacy of context contrast in comparison with dynamics prediction upon both
types of $\mathbb{V}_{\theta}$
in PDVF and PAnDR. 
The associated variants
are shown in Table \ref{table:alg_properties}. As in Fig.\ref{fig:Struction Ablation}, PDVF with CER has superior performance than original PDVF
while PAnDR without MI outperforms PAnDR with NA-GA.
Thus, context contrast is more favorable in both cases of $\mathbb{V}_{\theta}$ compared with dynamics prediction.

\paragraph{(A3-\uppercase\expandafter{\romannumeral1}) $\mathbb{V}_{\theta}$ Approximation \& GA Adaptation.}

With the variants in Table~\ref{table:alg_properties},  
we compares: PAnDR  v.s. PAnDR with QA, and PDVF v.s. PDVF with NA-GA. The results in Fig.\ref{fig:Struction Ablation}
verify that there is a clear advantage of using NN nonlinear approximation of $\mathbb{V}_{\theta}$ along with GA optimization over the quadratic counterpart.

\paragraph{(A3-\uppercase\expandafter{\romannumeral2}) Adaptation Performance against GA Steps.}
Since there is no longer a closed-form solution to optimization available for $\mathbb{V}_{\theta}$ with NN nonlinear approximation,
we investigate how consecutive steps of GA affect the adpatation performance of PAnDR.
The results for various choices of GA step number are provided in Table~\ref{table:Para Tune Ant}-\ref{table:certain step spaceship} in Appendix~\ref{Ape:Tuning Parameter Experiment}. 
Moreover, we provide detailed analysis for useful insights.

\begin{table}[t]
\vspace{-0.2cm}
	\centering
	\scalebox{0.60}{
		\begin{tabular}{c|ccc}
			\toprule
			Alg. / Variant & Env. Repre. & $\mathbb{V}_{\theta}$ Approx. \& Policy Opt. & If Use MI \\
			\midrule
			\textbf{PDVF} & DP & Quadratic \& Close-form & \xmark \\
            \textbf{PDVF with CER} & CL & Quadratic \& Close-form  & \xmark \\
            \textbf{PDVF with NA-GA} &	DP & NN Nonlinear \& GA & \xmark \\
            \textbf{PAnDR without MI} &	CL & NN Nonlinear \& GA & \xmark \\
            \textbf{PAnDR with QA} &	CL & Quadratic \& Close-form &	\cmark\\
            \textbf{PAnDR} & CL & NN Nonlinear \& GA & \cmark \\

			\bottomrule
		\end{tabular}
	}
\vspace{-0.1cm}
	\caption{Comparison of the variants with respect to three algorithmic factors  (Fig.~\ref{fig:Struction Ablation}). Abbreviations are used, i.e., Dynamics Prediction (DP), Contrastive Learning (CL).}
\label{table:alg_properties}
\end{table}

\section{Conclusion and Discussion}
\label{sec:conclusion}
This paper focus on fast adaptation problem with only offline experiences on training environments available.
Following the paradigm of~\cite{RaileanuGSF20PDVF},
we propose a new algorithm, PAnDR, that consists of effective environment and policy representations
and better policy optimization during online adaptation.
In essence, policy adaptation in PAnDR (and PDVF) is a process of policy representation optimization and decoding.
Both the two parts are not well studied at present.
Optimization in policy representation space is non-trivial as there is a lack of knowledge on the landscape of the space; 
while the decoding of such latent representation can be invalid without carefully handling the uncertainty~\cite{Pascal21Blackbox}.
We consider that they are significant and there is much room for improvement.
In addition, advances in environment and policy representation learning are expected to further improve the performance of PAnDR.

\section*{Acknowledgments}
This work is supported by the National Science Fund for Distinguished Young Scholarship of China (No. 62025602), the National Key R\&D Program of China (Nos. 2018AAA0100900), the National Natural Science Foundation of China (Nos. U1803263, 62106172 and 11931015), the Key Technology Research and Development Program of Science and Technology-Scientific and Technological Innovation Team of Shaanxi Province (Grant No. 2020TD-013), the XPLORER PRIZE, the New Generation of Artificial Intelligence Science and Technology Major Project of Tianjin (Grant No.: 19ZXZNGX00010), and the Science and Technology on Information Systems Engineering Laboratory (Grant No. WDZC20205250407).

\bibliographystyle{named}
\bibliography{ijcai22}

\clearpage

\appendix
\section*{Appendix}
\label{Ape}

\section{Complete Experimental Results}
\label{Ape:complete_exp_results}

\subsection{Performance on Training Environments}
\label{Ape: Training Environment Result}
\begin{figure}[h]
\centering
\subfigure[Ant-Wind]{
\includegraphics[width=0.35\textwidth]{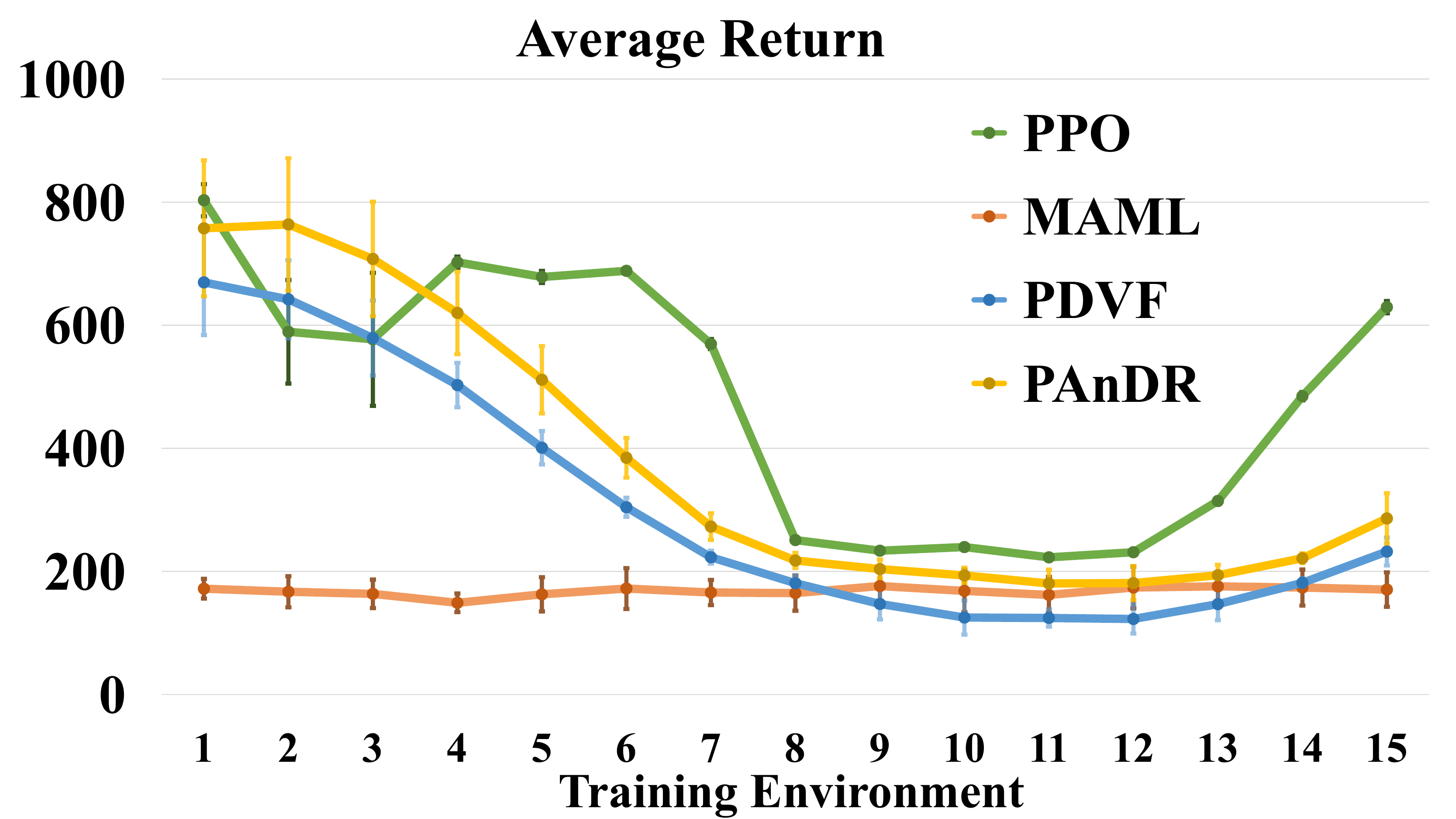}}
\quad
\subfigure[HalfCheetah-Mass]{
\includegraphics[width=0.35\textwidth]{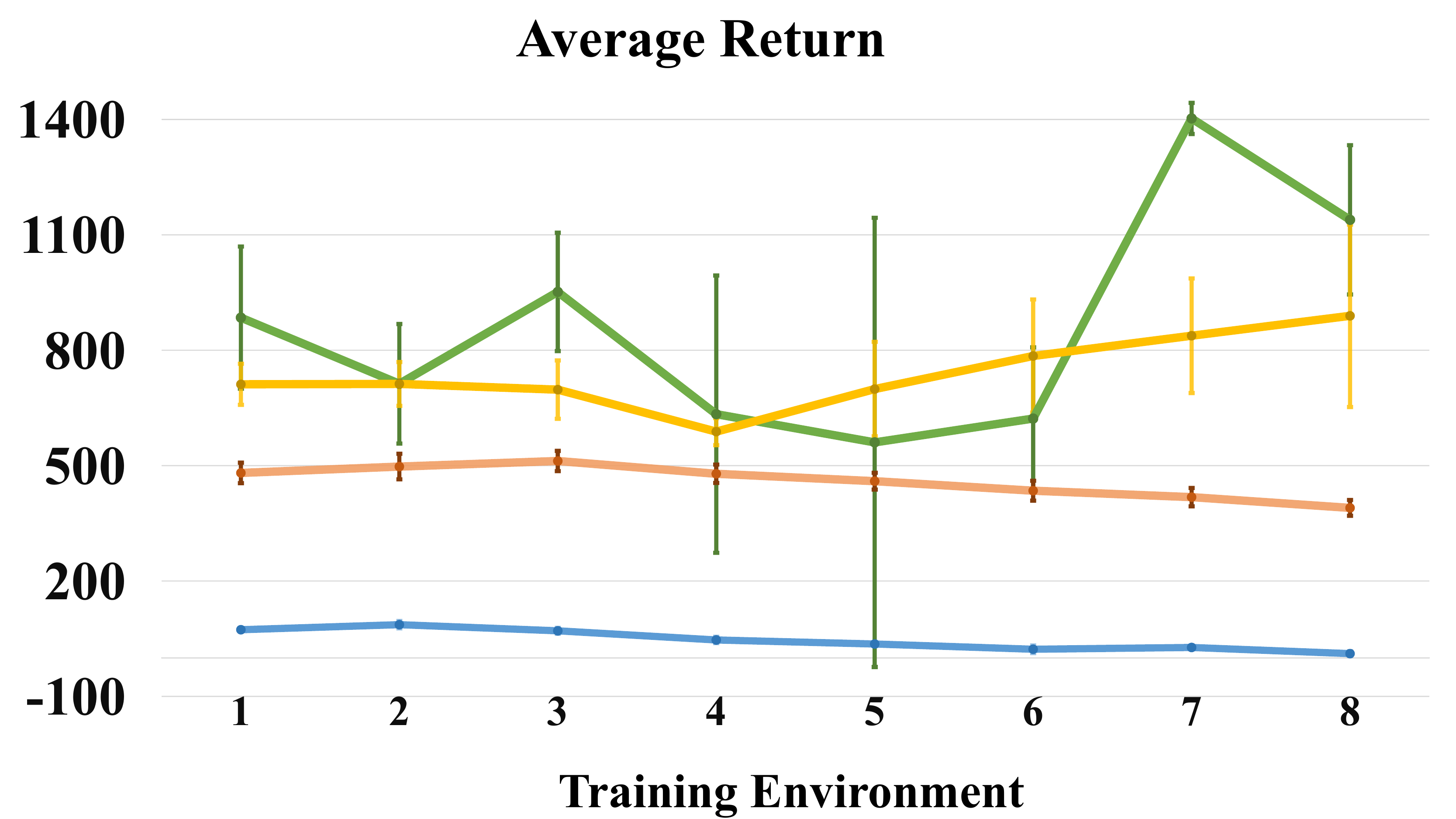}}
\caption{Experimental results of algorithm generalization in training environment.}
\label{fig:train env}
\end{figure}

Although the aim of meta-RL is the fast adaptation to new testing environments, we also present the results by testing the trained agents on the original training environments, as shown in Fig.\ref{fig:train env}. Obviously, PPO achieves the best performance in most of the environments thanks to re-learning in each environment. Although our method is worse than PPO in most training environments, its overall performance is clearly better than MAML and PDVF. In a few environments, PAnDR even surprisingly outperforms PPO algorithm, which further reveals the superiority and potential of PAnDR.

We find that PAnDR performs poorly in environments 8, 9, 10, 11 and 12. This is because the learning of our algorithm is based on the offline dataset collected by PPO. Due to the poor performance of PPO, the collected offline experience is of low quality, and we cannot extract abundant and effective environment and policy information from it, thus limiting the performance ceiling of our algorithm

The reason for the poor performance of PDVF in the testing environment of HalfCheetah-Mass showed in Fig.\ref{fig:new env test} can also be summarized from Fig.\ref{fig:train env}. PDVF can't get satisfactory performances even in the training environments, so it can't be expected to generalize in the testing environments and perform well. 

\subsection{In-depth Analysis of MI-based Refinement}
\label{ape: MI ablatio studies}

\begin{figure}[t]
\centering

\subfigure[Swimmer-Fluid]{
\includegraphics[width=0.23\textwidth]{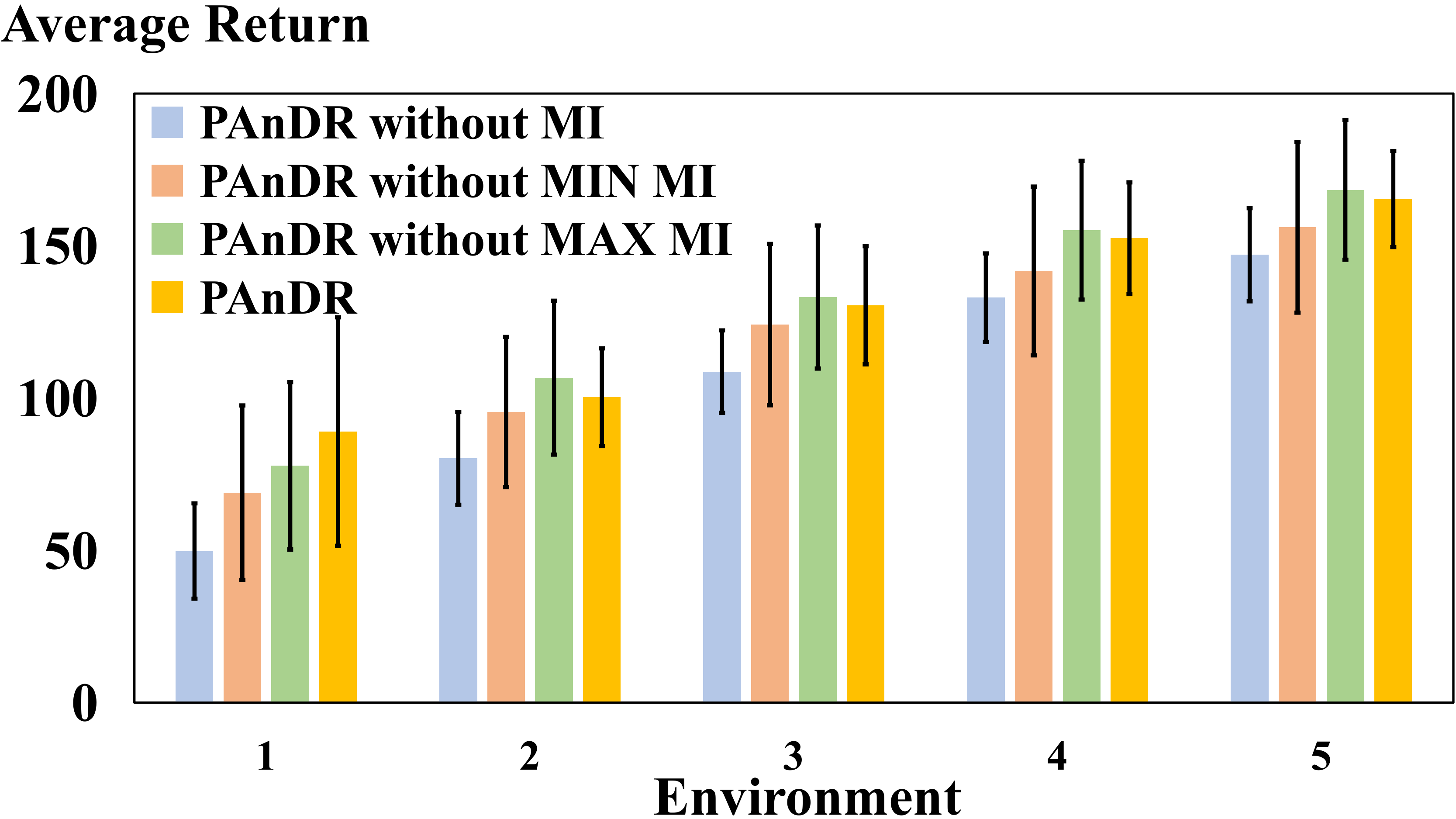}}
\subfigure[HalfCheetah-Mass]{
\includegraphics[width=0.23\textwidth]{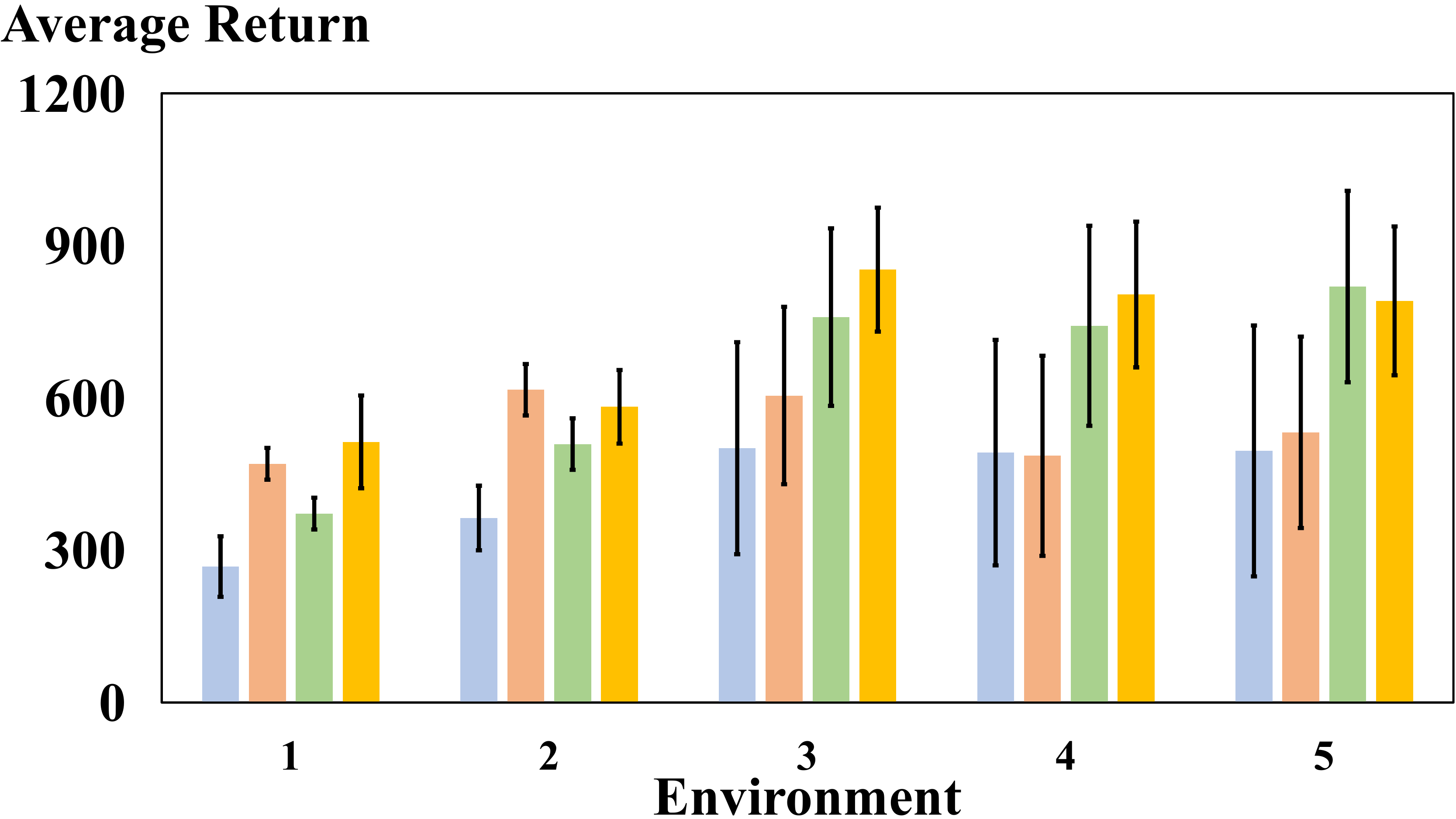}}
\caption{Ablations for MI-based representation refinement in remain environments.}
\label{fig:Remain Albation Mi}
\end{figure}

Here we give the results on Swimmer and HalfCheetah in Fig.\ref{fig:Remain Albation Mi}. As mentioned in Section \ref{Experiment:Ablation}, both MI minimizing and maximizing positively contribute to the improvement of PAnDR’s overall performance in all environment, which proves that both MI-based refinement components are indispensable in PAnDR.  
We also observe a phenomenon from the experiments that the results of using only MI-maximization refinement are slightly worse than those of using only MI-minimization refinement in most of the environments. This may mainly due to following two factors. On the one hand, additional environment-policy joint representations are introduced into the refinement process compared with that of MI-minimization. An inferior joint representation may result in limited or sometimes even negative improvements to the policy and environment representations during the refinement. On the other hand, in contrast to MI-minimization, the current policy representation space is not further compressed in MI-minimization refinement. As a result, the policy representation may still contains residual redundant information, which can negatively affect the subsequent value estimation and online adaptation.



\begin{figure}[t]
\centering
\subfigure[Swimmer-Fluid]{
\includegraphics[width=0.23\textwidth]{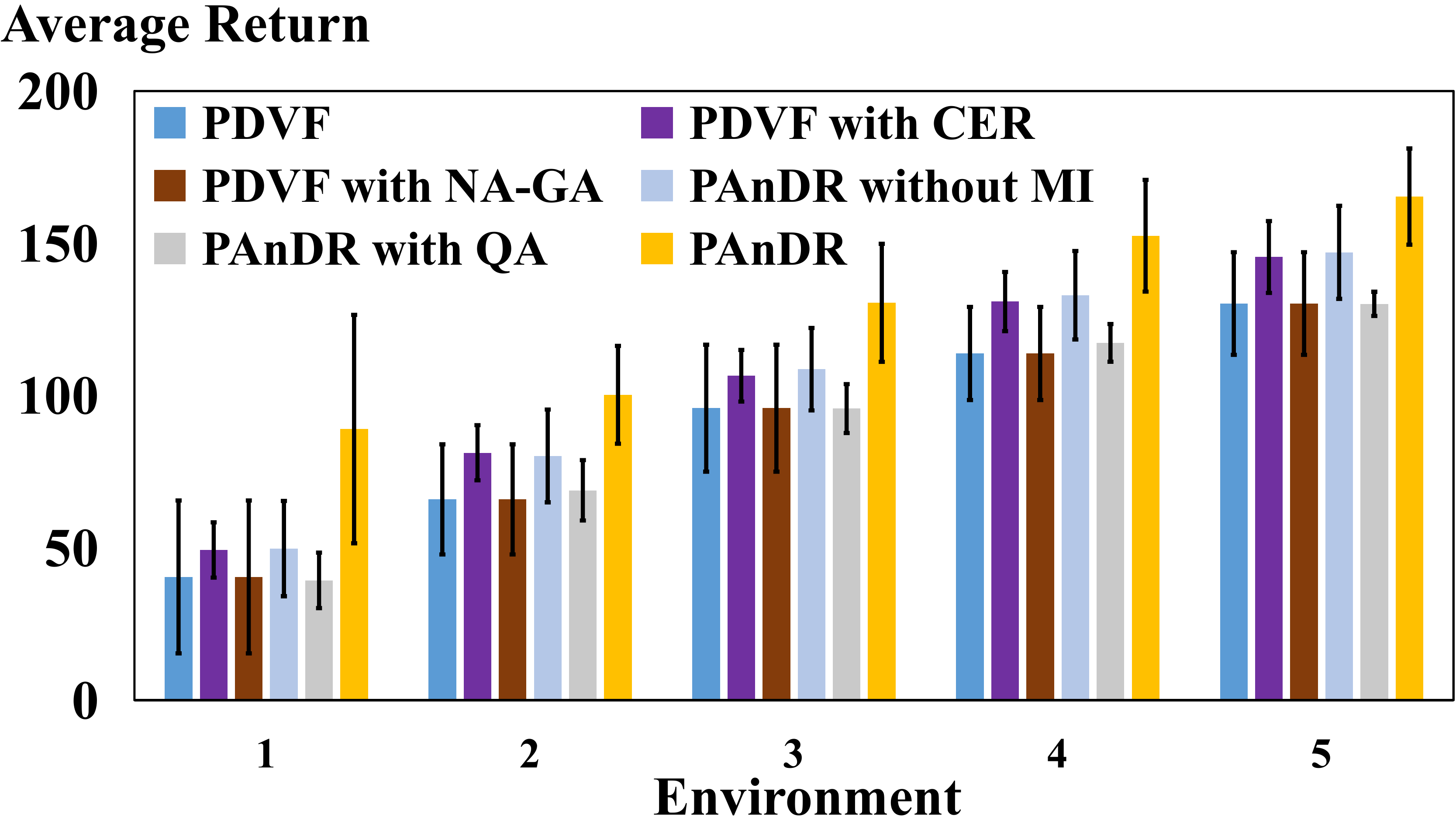}}
\subfigure[HalfCheetah-Mass]{
\includegraphics[width=0.23\textwidth]{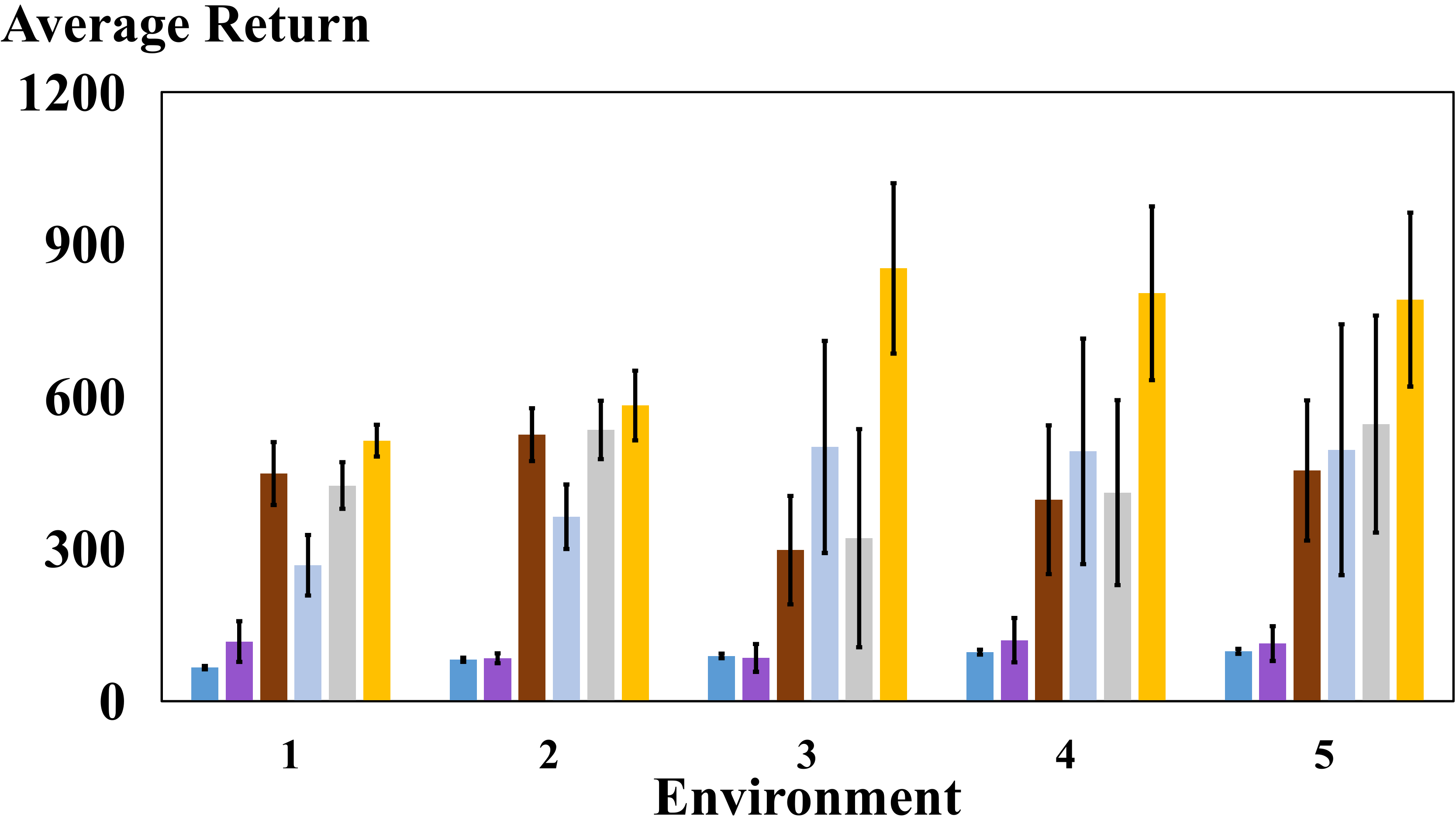}}
\vspace{-0.2cm}
\caption{Efficacy analysis for contrast environment representation (CER) and neural approximation of PDVF with GA adaptation (NA-GA) in contrast to quadratic approximation (QA) in remain environments.}
\label{fig:Remain Ablation Structure}
\end{figure}

\subsection{Adaptation Performance against GA Steps.}
\label{Ape:Tuning Parameter Experiment}
During the online adaptation phase, we use the gradient ascent (GA) to optimize the policy representation.  For this purpose, we conduct experiments of different numbers of steps of gradient ascent, the results are shown in Table \ref{table:Para Tune Ant} to Table \ref{table:Para Tune HalfCheetah}. It can be seen that as the number of steps increases, the performance of the optimized policy also rises gradually.  

In Fig.\ref{fig::Para Tune Space 3}, we give the distribution of policy performance at each step when the number of $steps=200$ in the SpaceShip environment, each policy is randomly sampled as described in Alg.\ref{Alg:Adaptation to New Environment}.  It can be concluded from the figures that different initial policies have different performance trends as the number of GA steps increases. For example, as showed in Fig.\ref{fig::Para Tune Space 3} Policy 3 Env.1, when the selected initial policy is inferior, a better result can potentially be obtained by gradient ascent. To the opposite, when the selected initial policy is an approximated optimal one, GA with multiple steps may mislead the optimization towards the wrong direction so that the policy's performance is even worse as showed in Fig.\ref{fig::Para Tune Space 3} Policy 3 Env.4. However, the more the number of gradients ascending, the more computation required, and Finally, considering the balance between the additional computation cost of more GA steps and the optimization of policy, we choose to stop the gradient ascend when the step reaches to 100 during the adaptation.

However, evaluating the updated policy at each step and then adopting  the best-performing one will lead to heavy computational overhead.  To alleviate this problem, we evaluate the updated policy every $K$ steps. We give the results of different $K$ as shown in Table \ref{table:certain step Ant} to Table \ref{table:certain step HalfCheetah}. We can see that the results of evaluating policy at different GA frequency are relatively acceptable when $K \leq 5$, which can reduce computational cost by more than 50\%.

\begin{table}[t]
	
	\centering
	\scalebox{0.6}{
		\begin{tabular}{c|ccccc}
			\toprule
			Test Env. / Freq. & 10 & 20 & 50 & 100 & 200 \\
			\midrule
			\textbf{1} & $371.0 \pm 43$ & $388.1 \pm 45$& $404.9 \pm 43$ & $416.5 \pm 45$ & $428.3 \pm 71$  \\

            \textbf{2} & $472.3 \pm 54$ & $494.9 \pm 58$ & $513.5  \pm 58$ & $528.2 \pm 59$ & $540.6  \pm67$ \\
            \textbf{3} & $555.5 \pm 67$ & $589.4 \pm 69$ & $612.7 \pm 72$ & $622.3 \pm76$ & $648.0  \pm89$ \\
            \textbf{4} & $639.1 \pm70$ & $665.3 \pm 79$ & $694.8 \pm 86$ & $715.3 \pm88$ & $730.7  \pm94$\\
            \textbf{5} & $668.9 \pm82$ & $694.5 \pm 92$ & $727.2 \pm104$ & $752.1 \pm109$ & $797.2 \pm97$ \\

			\bottomrule
		\end{tabular}
	}
\caption{Policy performance against GA steps on Ant}
\label{table:Para Tune Ant}
\end{table} 

\begin{table}[t]

	\centering
	\scalebox{0.6}{
		\begin{tabular}{c|ccccc}
			\toprule
			Test Env. / Freq. & 10 & 20 & 50 & 100 & 200 \\
			\midrule
			\textbf{1} & $0.556 \pm0.17$ & $0.630\pm0.17$ & $0.661\pm0.15$  & $0.698\pm0.14$  &  $0.793\pm0.09$  \\

            \textbf{2} & $0.718\pm0.12$  & $0.762\pm0.10$ & $0.819\pm0.09$& $0.848\pm0.08$ & $0.858\pm0.08$ \\
            \textbf{3} & $0.797\pm0.08$ & $0.829\pm0.006$ & $0.861\pm0.06$ & $0.868\pm0.06$ & $0.872\pm0.06$  \\
            \textbf{4} & $0.753\pm0.09$ & $0.801\pm0.06$ & $0.837\pm0.06$ & $0.853\pm0.06$ & $0.869\pm0.06$ \\
            \textbf{5} & $0.681\pm0.11$ & $0.732\pm0.11$ & $0.755\pm0.10$ & $0.794\pm0.10$ &  $0.808\pm0.11$ \\

			\bottomrule
		\end{tabular}
	}
	\caption{Policy performance against GA steps on Spaceship}
\label{table:Para Tune Space}
\end{table} 

\begin{table}[t]
	
	\centering
	\scalebox{0.6}{
		\begin{tabular}{c|ccccc}
			\toprule
			Test Env. / Freq. & 10 & 20 & 50 & 100 & 200 \\
			\midrule
			\textbf{1} & $52.2\pm13$ & $54.0\pm13$ & $62.6\pm14$  & $71.7\pm14$  &   $80.0\pm8$ \\

            \textbf{2} &  $83.1\pm12$ & $85.1\pm12$ & $93.1\pm13$ & $103.4\pm14$ & $105.7\pm14$ \\
            \textbf{3} & $109.4\pm12$ & $112.6\pm12$ & $121.2\pm13$ & $127.8\pm14$ &  $132.8\pm15$ \\
            \textbf{4} & $132.3\pm11$ & $135.0\pm11$ & $141.9\pm12$ & $149.1\pm13$ & $154.3\pm14$\\
            \textbf{5} & $144.4\pm10$ & $146.8\pm10$ & $154.1\pm11$ & $161.5\pm13$ & $167.4\pm14$ \\

			\bottomrule
		\end{tabular}
	}
\caption{Policy performance against GA steps on Swimmer}
\label{table:Para Tune Swimmer}
\end{table} 

\begin{table}[htbp]

	\centering
	\scalebox{0.6}{
		\begin{tabular}{c|ccccc}
			\toprule
			Test Env. / Freq. & 10 & 20 & 50 & 100 & 200 \\
			\midrule
			\textbf{1} & $385.3\pm46$ & $423.3\pm48$ & $478.2\pm45$  & $508.7\pm46$  & $529.4\pm62$   \\

            \textbf{2} & $503.0\pm58$  & $503.8\pm47$ & $632.9\pm42$ & $638.6\pm37$ & $656.8\pm60$ \\
            \textbf{3} & $404.4\pm72$  & $591.9\pm55$ & $694.5\pm53$ & $813.9\pm46$ &  $816.19\pm105$\\
            \textbf{4} & $536.4\pm106$ & $631.0\pm104$ & $711.4\pm109$ & $724.5\pm106$ & $807.3\pm138$ \\
            \textbf{5} & $461.3\pm107$ & $577.5\pm89$ & $657.7\pm99$ & $731.4\pm108$ & $803.9\pm147$\\

			\bottomrule
		\end{tabular}
	}
	\caption{Policy performance against GA steps on HalfCheetah}
\label{table:Para Tune HalfCheetah}
\end{table}

\begin{table}[t]

	\centering
	\scalebox{0.55}{
		\begin{tabular}{c|cccccc}
			\toprule
			Test Env. / Freq. &  1 & 2 & 5 & 10 & 20 & Last \\
			\midrule
			\textbf{1} &  $428.3 \pm 71$ &$407.4\pm41$ &$386.4\pm37$ & $369.5\pm34$ & $351.4\pm34$  & $265.8\pm36$\\
            \textbf{2} & $540.6  \pm67$ &$519.1\pm50$ & $485.0\pm52$ & $466.2\pm54$ & $452.7\pm55$ & $335.7\pm52$ \\
            \textbf{3} & $648.0  \pm89$ &$626.0\pm61$ & $604.5\pm57$ &$587.7\pm62$ & $565.5\pm64$ & $402.6\pm73$ \\
            \textbf{4} & $730.7  \pm94$ & $707.7\pm66$ & $679.2\pm71$ & $663.8\pm75$ & $649.1\pm76$ & $463.6\pm89$ \\
            \textbf{5} &  $797.2 \pm97$ & $773.3\pm65$ &$739.7\pm69$ & $716.9\pm73$ & $680.1\pm78$ & $504.8\pm99$ \\
			\bottomrule
		\end{tabular}
	}
\caption{Policy evaluation at different GA frequency on Ant}
\label{table:certain step Ant}
\end{table}

\begin{table}[ht]
	
	\centering
	\scalebox{0.5}{
		\begin{tabular}{c|cccccc}
			\toprule
			Test Env. / Freq. &  1 & 2 & 5 & 10 & 20 & Last \\
			\midrule
			\textbf{1} & $0.793\pm0.09$ & $0.724\pm0.14 $&$0.630\pm0.16$ & $0.567\pm0.16$ & $0.520\pm0.16$  & $0.203\pm0.15$\\

            \textbf{2} & $0.858\pm0.08$ &$0.833\pm0.09$&$0.729\pm0.14$ & $0.684\pm0.14$ & $0.664\pm0.14$ & $0.187\pm0.13$\\
            \textbf{3} &$0.872\pm0.06$ & $0.858\pm0.06$ &$0.827\pm0.07$ & $0.800\pm0.07$ & $0.770\pm0.09$ & $0.17\pm0.11$ \\
            \textbf{4} &$0.869\pm0.06$ & $0.836\pm0.06$ &$0.822\pm0.07$ & $0.809\pm0.07$ & $0.770\pm0.09$ & $0.211\pm0.11$\\
            \textbf{5} &$0.808\pm0.11$ & $0.789\pm0.10$ &$0.767\pm0.11$ & $0.745\pm0.12$ & $0.707\pm0.12$ & $0.21\pm0.13$\\

			\bottomrule
		\end{tabular}
	}
\caption{Policy evaluation at different GA frequency on Spaceship}
\label{table:certain step spaceship}
\end{table}

\begin{table}[ht]
	
	\centering
	\scalebox{0.55}{
		\begin{tabular}{c|cccccc}
			\toprule
			Test Env. / Steps & 1& 2&5 & 10 & 20 & Last \\
			\midrule
			textbf{1} & $80.0\pm8$ & $73.9\pm13$ &$54.5\pm13$ & $52.8\pm13$ & $51.0\pm12$  & $42.0\pm12$ \\

            \textbf{2} &$105.7\pm14$ & $104.7\pm14$ &$86.0\pm14$  & $85.2\pm14$ & $83.0\pm14$ & $74.0\pm13$ \\
            \textbf{3} &$132.8\pm15$ &  $129.9\pm13$ &$110.7\pm14$ & $109.1\pm14$ & $108.0\pm14$ & $98.9\pm13$ \\
            \textbf{4} &$154.3\pm14$ & $151.4\pm13$ &$136.2\pm14$ & $135.1\pm13$ & $133.4\pm12$ & $123.4\pm13$ \\
            \textbf{5} &$167.4\pm14$ & $164.0\pm13$ &$149.6\pm13$ & $148.1\pm12$ & $145.8\pm11$ & $134.8\pm10$\\

			\bottomrule
		\end{tabular}
	}
\caption{Policy evaluation at different GA frequency on Swimmer}
\label{table:certain step swimmer}
\end{table}

\begin{table}[htbp]
	
	\centering
	\scalebox{0.5}{
		\begin{tabular}{c|cccccc}
			\toprule
			Test Env. / Steps & 1 &2 & 5 & 10 & 20 & Last \\
			\midrule
			\textbf{1} & $529.4\pm62$ & $524.5\pm45$ &$521.7\pm50$ & $494.4\pm49$ & $486.9\pm46$  & $314.0\pm92$ \\
            \textbf{2} & $656.8\pm60$ & $642.1\pm67$ &$636.7\pm62$ & $613.8\pm62$ & $591.4\pm70$ & $312.2\pm96$ \\
            \textbf{3} & $816.19\pm105$ & $781.5\pm56$ &$681.6\pm74$ & $645.7\pm74$ & $578.8\pm83$ & $272.2\pm97$ \\
            \textbf{4} & $807.3\pm138$ & $742.5\pm94$ &$577.4\pm104$ & $519.6\pm108$ & $410.2\pm116$ & $93.6\pm126$ \\
            \textbf{5} & $803.9\pm147$ & $754.5\pm101$ &$463.6\pm152$ & $392.6\pm147$ & $280.9\pm129$ & $-91.2\pm146$ \\

			\bottomrule
		\end{tabular}
	}
\caption{Policy evaluation at different GA frequency on HalfCheetah}
\label{table:certain step HalfCheetah}
\end{table}

\subsection{Remaining Ablation Analysis}
The remaining results of \textbf{Q2} and \textbf{Q3} are showed in Fig.\ref{fig:Remain Albation Mi} and Fig.\ref{fig:Remain Ablation Structure}. The conclusions are consistent with the ones in Section \ref{Experiment:Ablation}: (1) Context contrast is more favorable in designing $\mathbb{V}_{\theta}$ both in \cite{RaileanuGSF20PDVF} and PAnDR compared with dynamics prediction; (2) NN nonlinear approximation of $\mathbb{V}_{\theta}$ along with GA optimization has clear advantage over the quadratic counterpart.

\section{Additional Experimental Details}
\label{Ape:Additional Experimental Details}

\subsection{Other Algorithms Experimental settings}

For PDVF, we use the source codes from here\footnote{https://github.com/rraileanu/policy-dynamics-value-functions}. In PDVF we follow the parameters settings used in the original paper. We first tried to use the source code directly for training. After running three seeds, we find that except for the Ant-Wind, PDVF cannot achieve the performance neither on Swimmer-Fluid nor Spaceship reported in the original paper. Moreover, the source codes take a lot of time collecting new data during the second stage for data augmentation. On the same device, the training time of PDVF is about three times that of PAnDR, which is obviously unacceptable. Therefore, in the subsequent experiments, we abandon the second stage of PDVF and use the results of the first stage training of 7 seeds.

For MAML, we use the codes from here\footnote{https://github.com/tristandeleu/pytorch-maml-rl}.
In MAML, in the meta training phase, the update number is set to 500 and during each update we sample 20 trajectories from each task; in the meta adaptation phase, the update number is set to 10 and during each update we sample 20 trajectories from each task.

\subsection{Equipment and Software}
All experiments are performed on a Linux server with Ubuntu 18.04 containing 256G of RAM, Intel(R) Xeon(R) CPU E5-2680 v4 @ 2.40GHz CPU processor with 56 logical cores, the server contains 4 Nvidia-2080Ti GPUs. Our codes are implemented with Python 3.7, Torch 1.7.1, gym 0.15.7, and MuJoCo 2.0.2.

\subsection{Hyperparameters Setting}
\label{Ape:Hyperparameters Setting}

Before training the encoder networks, we need first to train the PPO agents on the training environments, and then use these agents to collect data. The hyperparameters setting of the PPO algorithm is shown in Table \ref{table:Hyperparameters Setting PPO}.

Table \ref{table:Hyperparameters Setting Encoder} shows the training parameters of the encoder networks. The large difference between the weights of maximizing mutual information and minimizing mutual information is due to the different ways of calculating the upper and lower bounds of MI. So we give a larger weight to the MI minimizing weight to make the smaller loss work as well.

Table \ref{table:Hyperparameters Setting Value} gives the main parameters of the value approximation and policy adaptation stage.

\section{Algorithm Pseudocode}
\label{Ape:Algorithm Pseudocode}
Algorithm \ref{Alg:Training Encoders} describes the training process of encoders. First, we trained the environment representation and policy representation by context contrast and policy recovery, respectively. Then the MI loss is introduced to further refine the encoder networks.


\IncMargin{1em}
\begin{algorithm} \SetKwData{Left}{left}\SetKwData{This}{this}\SetKwData{Up}{up} \SetKwFunction{Union}{Union}\SetKwFunction{FindCompress}{FindCompress} \SetKwInOut{Input}{input}\SetKwInOut{Output}{output}
	
	\Input{Environment encoder $\phi_{e}$, policy encoder $\phi_{\pi}$, policy decoder $\phi_{\pi}^{\text{dec}}$, policy-environment encoder $\phi_{b}$, training set $D$ and hyper-parameters $\alpha$, $\beta$.} 
	\BlankLine 
 	\If(){use maximum mutual information}{Divide $D$ according to the environment and policy to get $D_{i,k}$\; Sample anchor, positive and negative from $D_{i,k}$\; Training $\phi_{b}$ with contrastive loss.}
	 \For{each training step}{ 
	 	Sample anchor and positive from $D_{i,\cdot}$, sample negative from  $D_{j,\cdot}, j\neq i$ \; 
	 	Calculate environment encoder contrastive loss $\mathcal{L}_{\text{CC}}$ with equation \ref{eq:CL loss of env_encoder}\;
        Sample data from $D_{\cdot, k}=\bigcup_{i=1}^{M} D_{i,k}$\;
        Calculate policy recovery loss $\mathcal{L}_{\text{PR}}$ with equation \ref{eq:recovery loss of policy_encoder}\;
 	    \If(){use minimize mutual information}{Sample data from D \;
        Using encoder $\phi_{e}$ to get environment embedding $z_{e}$\;
        Using encoder $\phi_{\pi}$ to get policy embedding $z_{\pi}$\;
        Calculate minimize mutual information loss $\mathcal{L}_{RD}$ using equation \ref{eq: min MI loss}\;
 		}
 		\If(){use maximum mutual information}{Sample data from D \;
        Using encoder $\phi_{e}$ to get environment embedding $z_{e}$\;
        Using encoder $\phi_{\pi}$ to get policy embedding $z_{\pi}$\;
        Using encoder $\phi_{b}$ to get policy-environment joint embedding $z_{b}$\;
        Calculate maximize mutual information loss $\mathcal{L}_{RC}$ using equation \ref{eq:max_mi}\;
 		}
 	 }
 	Update environment encoder $\phi_{e}$, policy encoder $\phi_{\pi}$ and decoder  $\phi_{\pi}^{\text{dec}}$  with equation \ref{eq:final loss of encoders}\; 
    \caption{PAnDR: Representation Learning and Refinement} 
    \label{Alg:Training Encoders} 
 	 \end{algorithm}
 \DecMargin{1em} 
 
Algorithm \ref{Alg:Training Policy-Dynamics Value Function} and \ref{Alg:Adaptation to New Environment} correspond to the training process of the value function and the online adaptation process of the policy representation, respectively. In Algorithm \ref{Alg:Training Policy-Dynamics Value Function}, we formulate the training of value function as a prediction problem using Monte Carlo return as the label.  In Algorithm \ref{Alg:Adaptation to New Environment}, we describe the optimization of the policy representation by gradient ascent along the direction of increasing values of $\mathbb{V}_{\theta}$. 

\IncMargin{1em}
\begin{algorithm} \SetKwData{Left}{left}\SetKwData{This}{this}\SetKwData{Up}{up} \SetKwFunction{Union}{Union}\SetKwFunction{FindCompress}{FindCompress} \SetKwInOut{Input}{input}\SetKwInOut{Output}{output}
	\Input{ Environment encoder $\phi_{e}$, policy encoder $\phi_{\pi}$, training set $D$ and value estimation network $\mathbb{V}_{\theta}$ .} 
	\BlankLine 
	
	 \For{each training step}{ 
	 	Sample episode $\tau={(s_{i},a_{i},s_{i+1})_{i=0}^{T}}$ and episode reward $G(s_{0})$\; 
	    Input $\tau$ into environment encoder $\phi_{e}$ to get environment embedding $z_{e}$\;
        Input $\tau$ into policy encoder $\phi_{\pi}$ to get policy embedding $z_{\pi}$\;
        Update $\mathbb{V}_{\theta}$ with equation \ref{eq:loss of value function}\;
 	 }

 	 	  \caption{PAnDR: Value Function Network Approximation}
 	 	  \label{Alg:Training Policy-Dynamics Value Function} 
 	 \end{algorithm}
 \DecMargin{1em} 

\IncMargin{1em}
\begin{algorithm} \SetKwData{Left}{left}\SetKwData{This}{this}\SetKwData{Up}{up} \SetKwFunction{Union}{Union}\SetKwFunction{FindCompress}{FindCompress} \SetKwInOut{Input}{input}\SetKwInOut{Output}{output}
	
	\Input{ Environment encoder $\phi_{e}$, policy encoder $\phi_{\pi}$, policy decoder  $\phi_{\pi}^{\text{dec}}$, value estimate network $\mathbb{V}_{\theta}$ and a initial policy $\pi$.} 
	\BlankLine 
	
 	Sample a trajectory $\tau={(s_{i},a_{i},s_{i+1})_{i=0}^{K}}$ using policy $\pi$\; 
    Input $\tau$ into environment encoder $\phi_{e}$ to get environment embedding  $z_{e}^{\text{on}}$\;
    Input $\tau$ into policy encoder $\phi_{\pi}$ to get policy embedding $z_{\pi}^{0}$\;
    \For{$t=0,...,N$}{ 
    Update policy embedding: $z_{\pi}^{t+1}= z_{\pi}^{t}+ \eta \nabla_{z_{\pi}^{t}} \mathbb{V}_{\theta} (s, z_{e}^{\text{on}}, z_{\pi}^{t} )$\;}
    Using policy decoder $\phi_{\pi}^{\text{dec}}$ based on $z_{\pi}^{*}$  to output action to interact with new environment.
  	  \caption{PAnDR: Policy Adaptation via Representation Gradient Ascent}
  	  \label{Alg:Adaptation to New Environment} 
 	 \end{algorithm}
 \DecMargin{1em}

\begin{table}[b]
	
	\centering
	\scalebox{0.75}{
		\begin{tabular}{c|c}
			\toprule
			\textbf{Hyperparameters} &\textbf{ Value}\\
			\midrule
			Batch size & 32 \\

            Training steps & 1e6\\
            $\epsilon$ & 0.2 \\
            Learning rate & 3e-4 \\
            $\gamma$ & 0.99 \\
            Number of trajs. collected under each Env. & 50 \\
			\bottomrule
		\end{tabular}
	}
\caption{Hyperparameters Setting of PPO}
\label{table:Hyperparameters Setting PPO}
\end{table}

\begin{table}

	\centering
	\scalebox{0.8}{
		\begin{tabular}{c|c}
			\toprule
			\textbf{Hyperparameters} & \textbf{Value}\\
			\midrule
            Training episodes & 3000\\
            $\phi_{\pi}$ learning rate & 0.01 \\
            $\phi_{e}$ learning rate & 0.001 \\
            $\psi_{1}$ learning rate & 0.005 \\
            $\psi_{2}$ learning rate & 0.005 \\
            The weight of MIN MI loss $\alpha$& 1000 \\
            The weight of MAX MI loss $\beta$& 1 \\ 
            The dimension of $z_{e}$ & 8 \\
            The dimension of $z_{\pi}$ & 8 \\
            The dimension of $z_{b}$ & 8 \\
			\bottomrule
		\end{tabular}
	}
\caption{Hyperparameters Setting of Encoder Networks}
\label{table:Hyperparameters Setting Encoder}
\end{table}

\begin{table}
	
	\centering
	\scalebox{0.8}{
		\begin{tabular}{c|c}
			\toprule
			\textbf{Hyperparameters} & \textbf{Value}\\
			\midrule
            Training episodes & 3000\\
            Learning rate & 0.005 \\
            Batch size & 128 \\
            The steps for gradient ascent in adaptation & 100 \\
			\bottomrule
		\end{tabular}
	}
\caption{Hyperparameters Setting of Value Approximation and Policy Adaptation}
\label{table:Hyperparameters Setting Value}
\end{table}

\section{Related Work}
\label{Ape:related_work}

\subsection{Meta-RL}
\label{Related Work:Meta-RL}
Meta-RL is meta-learning in the field of RL that the training and testing tasks are drawn from the same family of problems while having different components such as reward probabilities, environmental dynamics and so on. Common meta-RL methods are mainly divided into three classes. The first class is the model-based meta-RL, \cite{DuanSCBSA2016RL2}  uses a RNN structure to record historical information in the agent model and incoporate task-related knowledge into the model parameters. The second class is the optimization-based meta-RL methods. A representative method is  \cite{FinnAL2017MAML} that use base-learners to collect data on different subtasks and then use a meta-learner to learn from these data. However, both model-based and optimization-based method requires on-policy learning, of which the sampling efficiency is poor. The third class is the context-based methods which propose to train an encoder in multi different environments to extract the environmental information. The context of new environments is inferred using a small amount of data collected during the initial limited interactions. Then the context is used as the input of the policy and value networks to help the agent quickly adjust its policy and adapt to the new environments. The encoder is expected to capture the different task information. In \cite{rakelly2019PEARL,FakoorCSS20MQL}, the encoder networks are learned based on the reward signal. In \cite{FuTHCFLL2021}, the encoder is trained on the trajectories through contrastive learning. However, training context encoders using reward signals maybe unstable while directly using trajectories may introduce redundant information as the encoder learned based on trajectory may contains policy information. Recently, \cite{RaileanuGSF20PDVF} uses a predictive approach to try to extract environment context and policy context separately, but they are not completely decoupled. In contrast to these methods, our PAnDR can extract and refine policy and environment 
embeddings to obtain the decoupled representations.

\subsection{Representation Learning}
\label{Related Work:Representation Learning}
Due to the high complexity of the policy space in RL, it is difficult to directly optimize the policy in this space. Therefore, a possible way is to map the original high-dimensional policy space to a low-dimensional representation space and optimizing the policy in this representation space. In this way, one can greatly reduce the difficulty of searching for the optimal policy in the policy space. Some recent works focus on the policy representation and use the policy representation as auxiliary input to the value function. For example, PDVF \cite{RaileanuGSF20PDVF} used this idea in meta-RL. It's trained by using state and policy representations to predict corresponding action.  \cite{Jean2020PVN} uses the actions that policy sampled in probing states as policy representations. There are also some articles \cite{FaccioK2021PBVFs,Tang2020PeVFA} that use policy network parameters as policy representations. Then given the policy representation, by gradient ascent, the policy representation can be updated along the direction of increasing the values of value function. In this way, one can find an approximated optimal policy without the overhead of interacting with the environment. However, how to characterize the policy in a reasonable and efficient way and whether the policy representation can improve the meta-RL is still an open challenge. In contrast to policy representation, environment representation is usually applied in context-based meta-RL. There are two common ways to extract envionment representations. One way is to predict future information based on the information of current steps~\cite{RaileanuGSF20PDVF,guo2020bootstrapMETA}. The other way is to extract representations based on contrastive learning \cite{FuTHCFLL2021}. The goal of contrastive learning is to learn an embedding space in which
similar sample pairs stay close to each other while dissimilar
ones are far apart. A metric is then given to measure the similarity of the feature representations of these samples and a discriminative encoder is trained to group data with similar embedding while distinguish data with dissimilar embedding. Then the learned representations are used for downstream tasks. PAnDR also use contrastive learning to extract representations. We follow the training framework and the infoNCE loss used in CURL~\cite{laskin2020curl}. But we take a different approach than CURL to construct positive and negative examples. Details are given in Sec.~\ref{sec:methodology}.

\begin{figure*}[t]
\centering
\subfigure[Policy 1 Env. 1 to Env. 5]{
\includegraphics[width=0.19\textwidth]{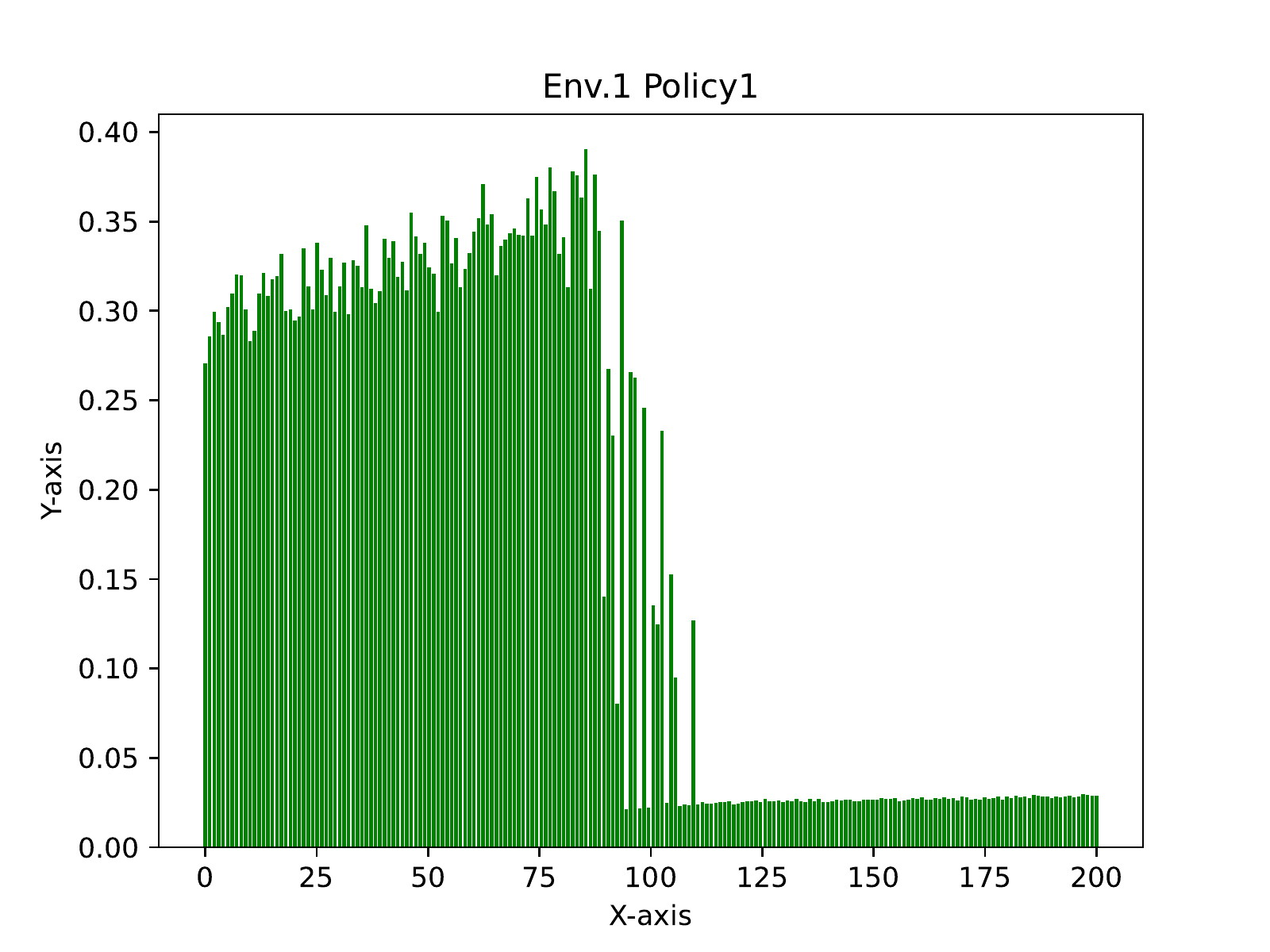}
\includegraphics[width=0.19\textwidth]{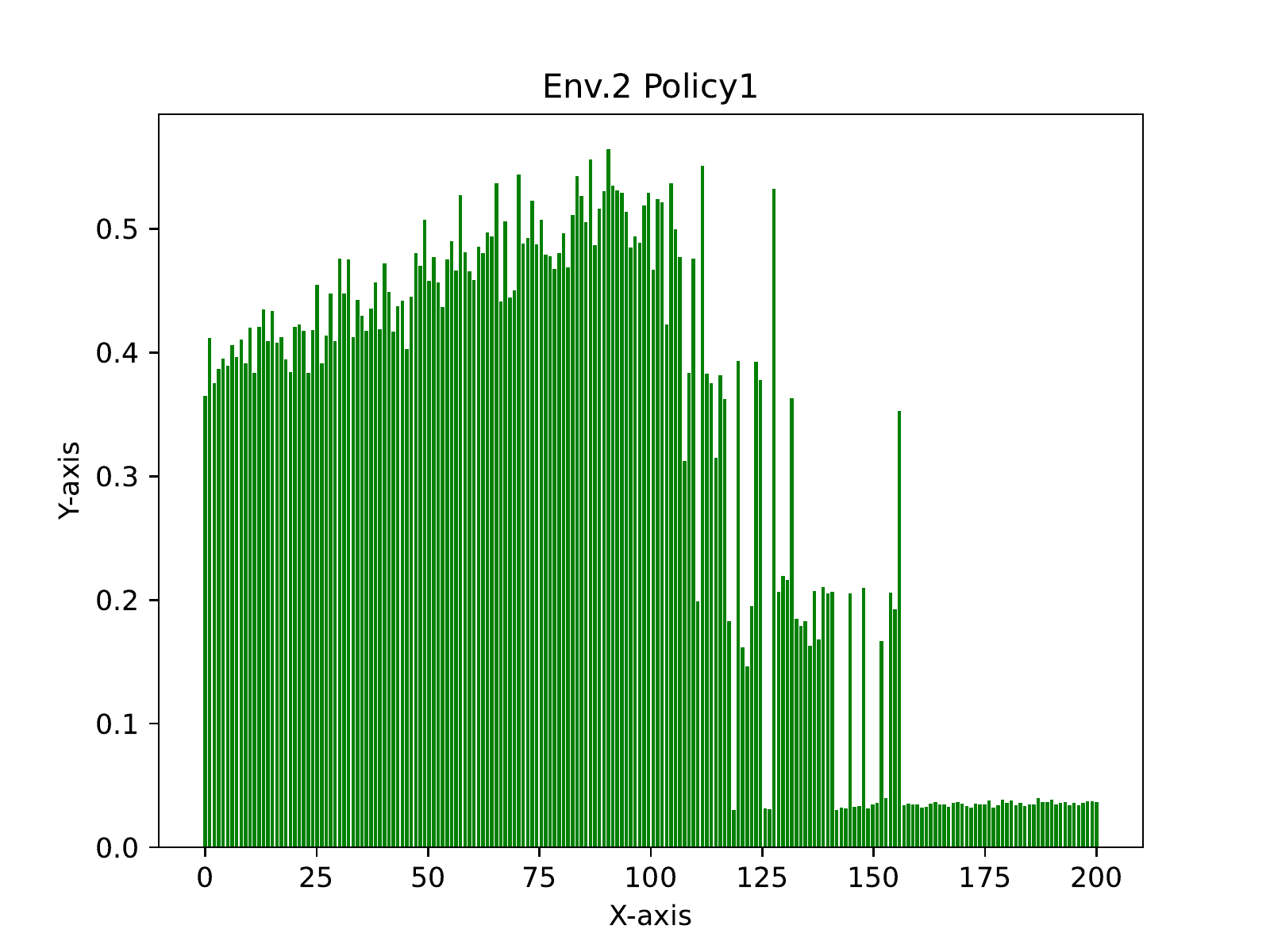}
\includegraphics[width=0.19\textwidth]{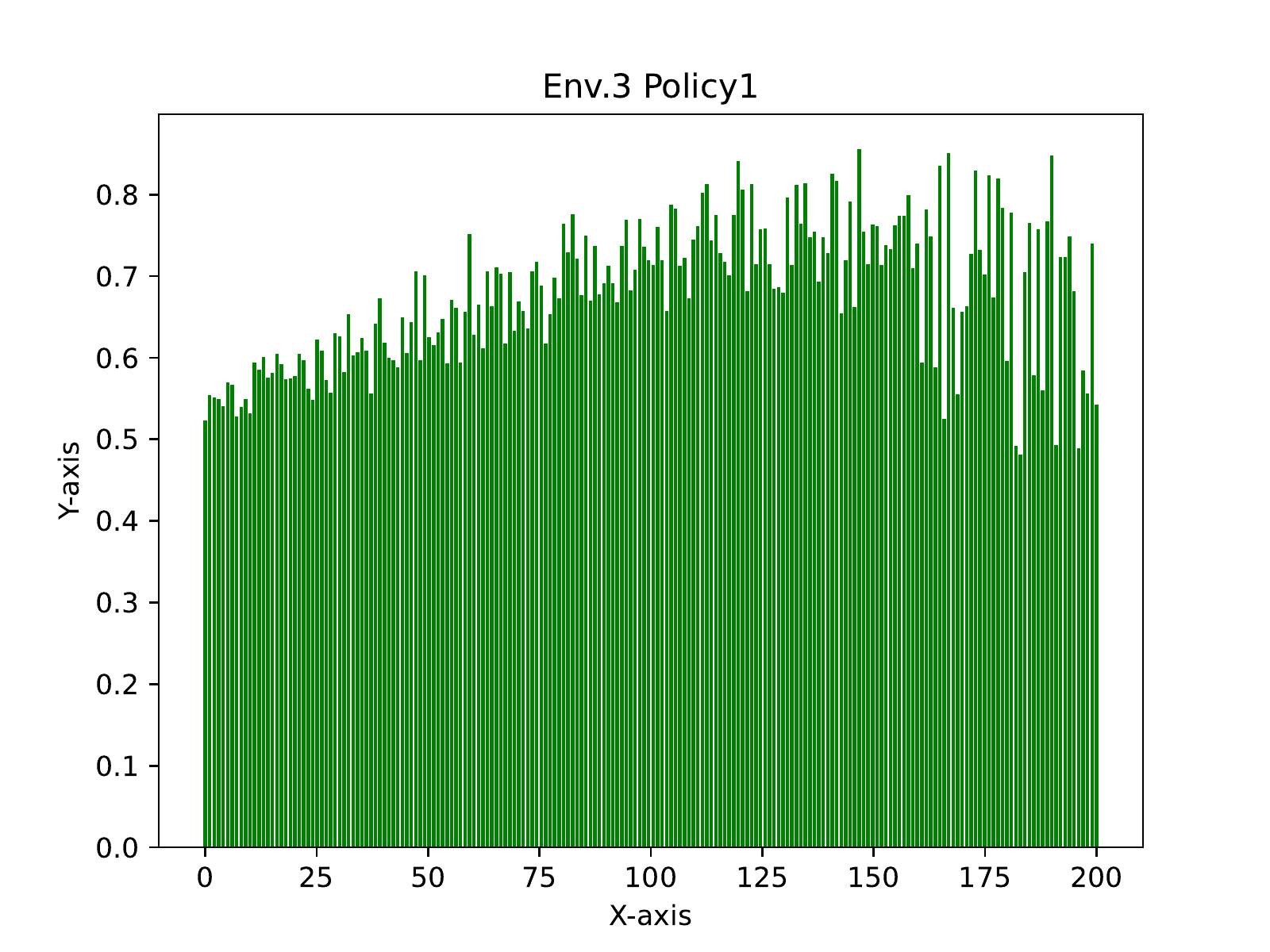}
\includegraphics[width=0.19\textwidth]{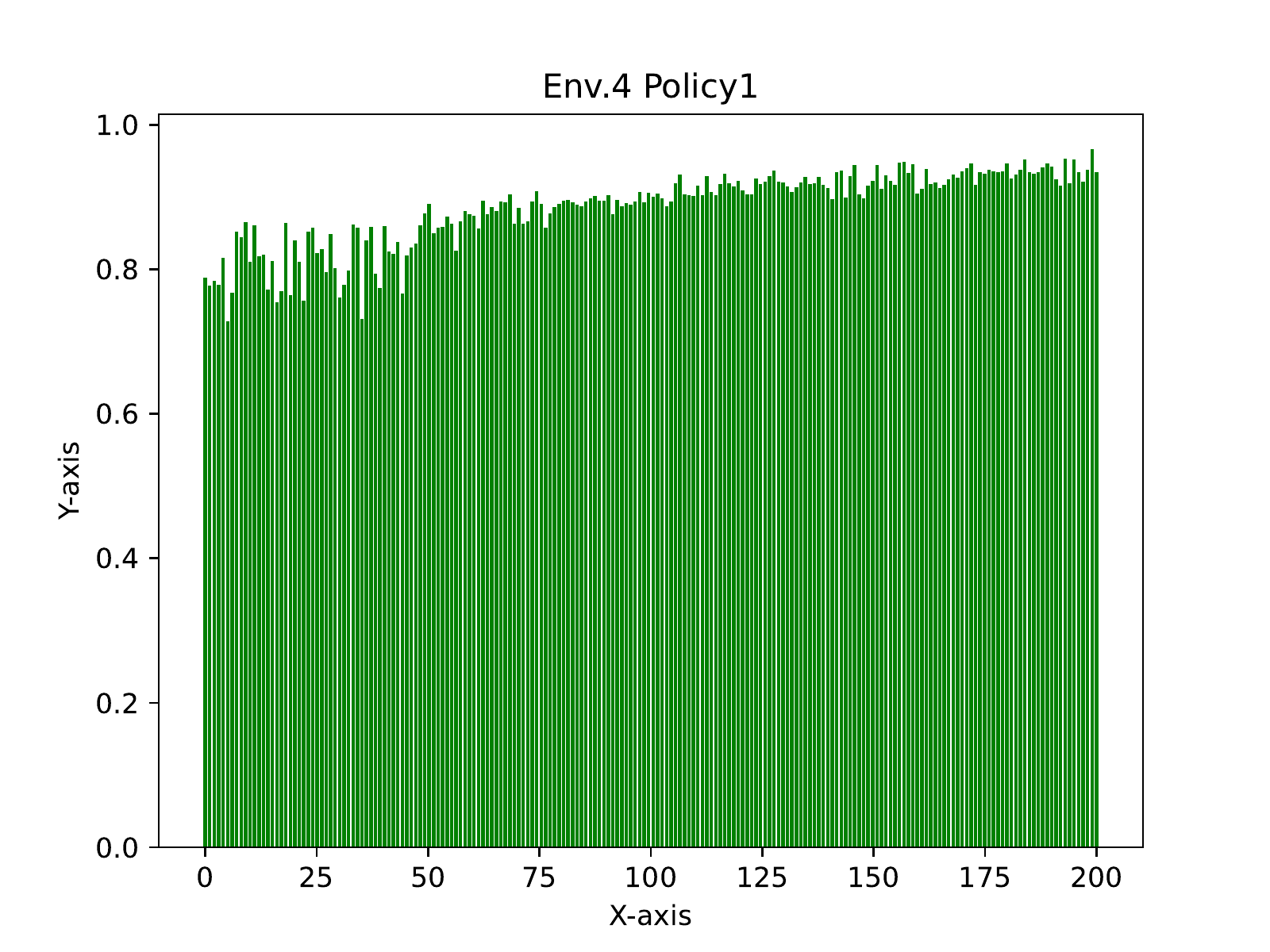}
\includegraphics[width=0.19\textwidth]{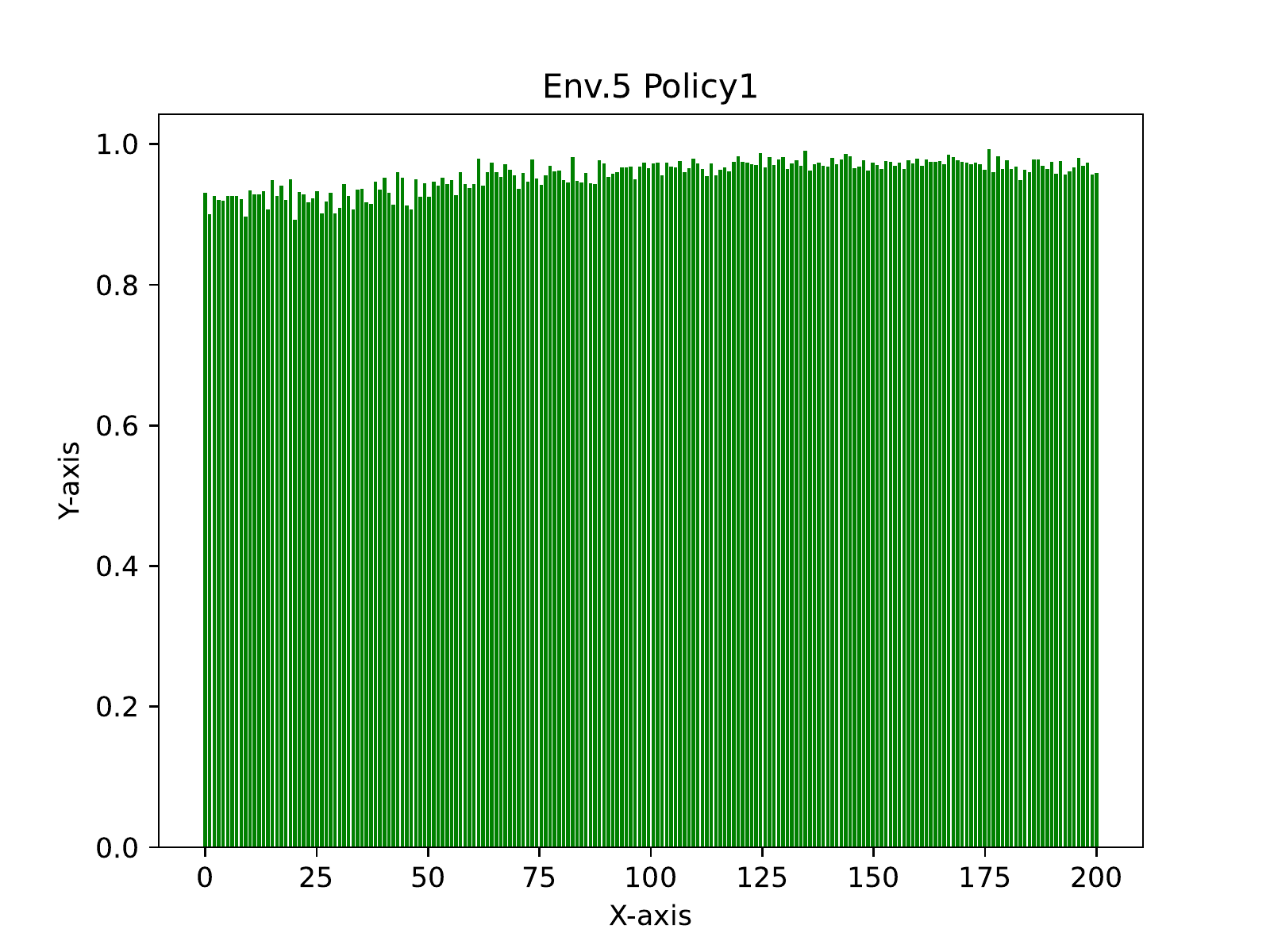}
}
\quad

\subfigure[Policy 2 Env. 1 to Env. 5]{
\includegraphics[width=0.19\textwidth]{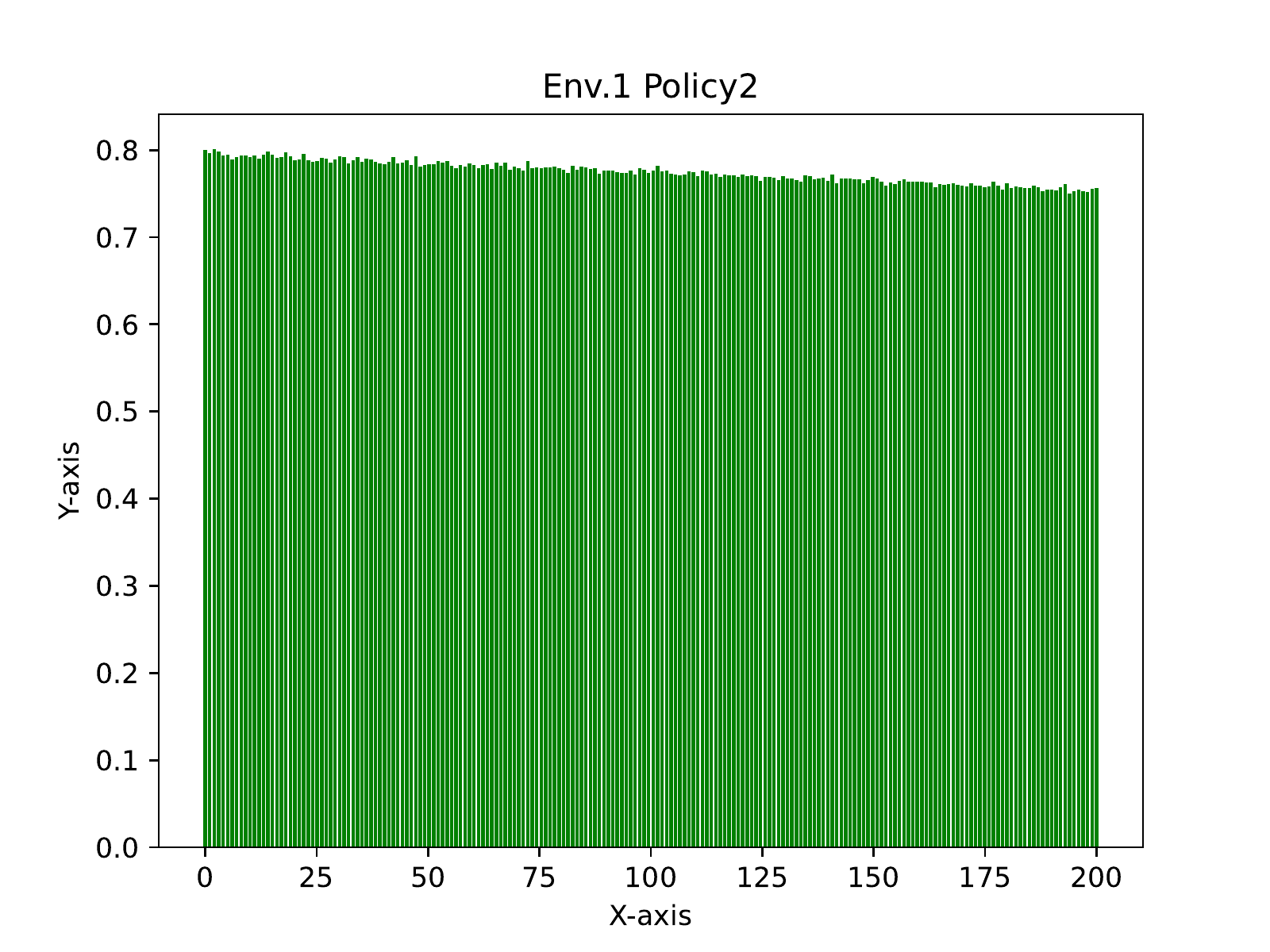}
\includegraphics[width=0.19\textwidth]{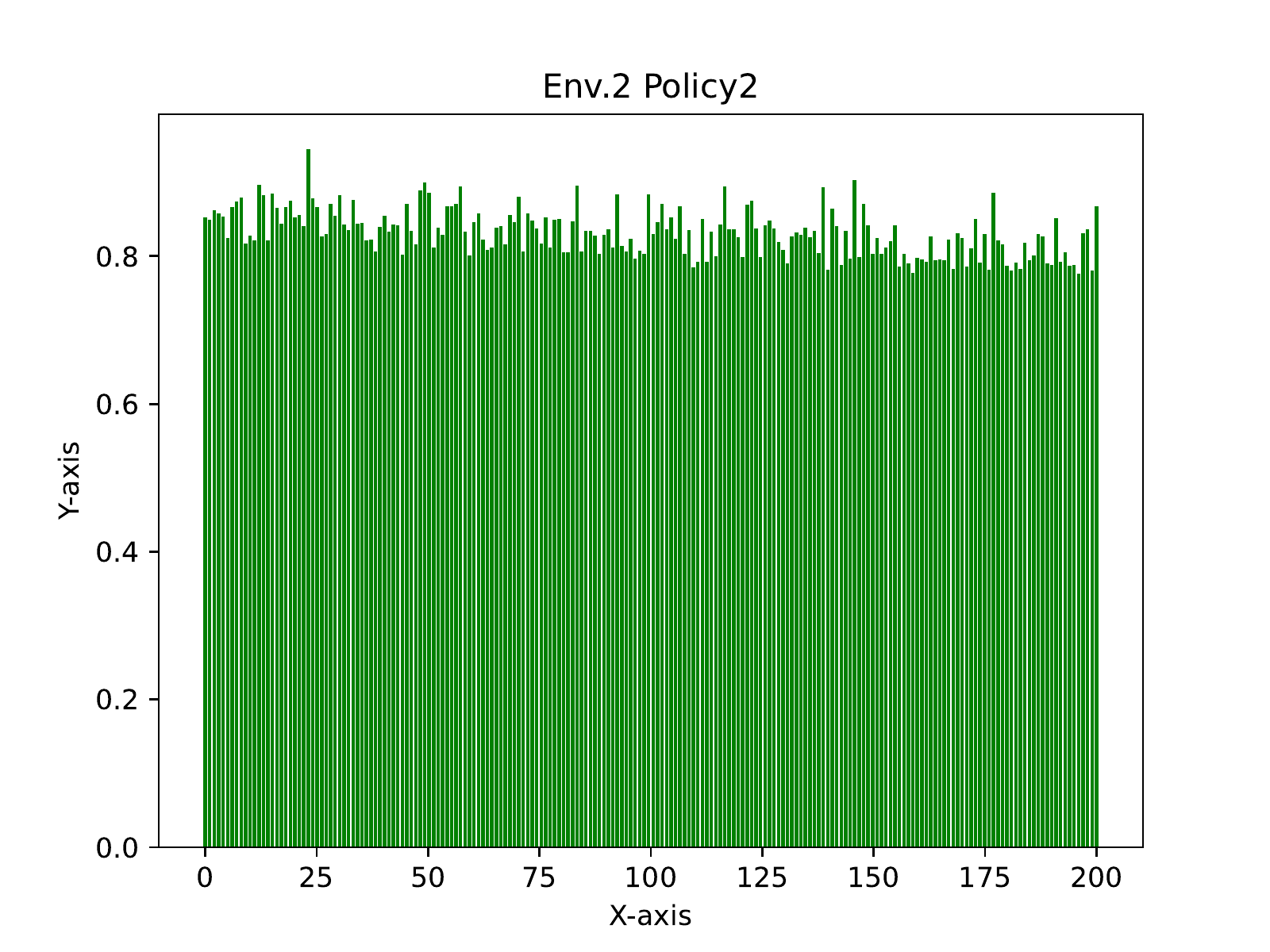}
\includegraphics[width=0.19\textwidth]{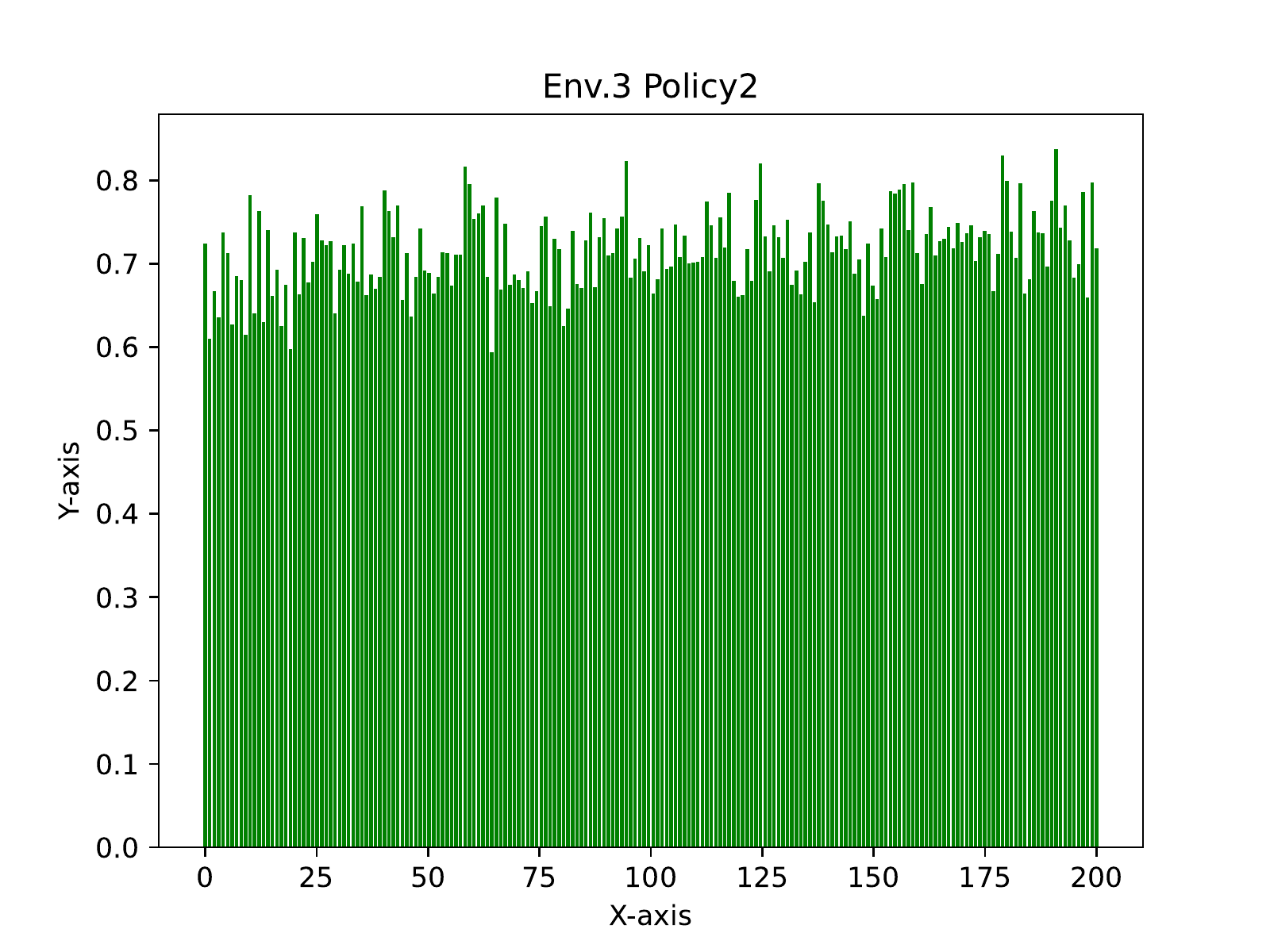}
\includegraphics[width=0.19\textwidth]{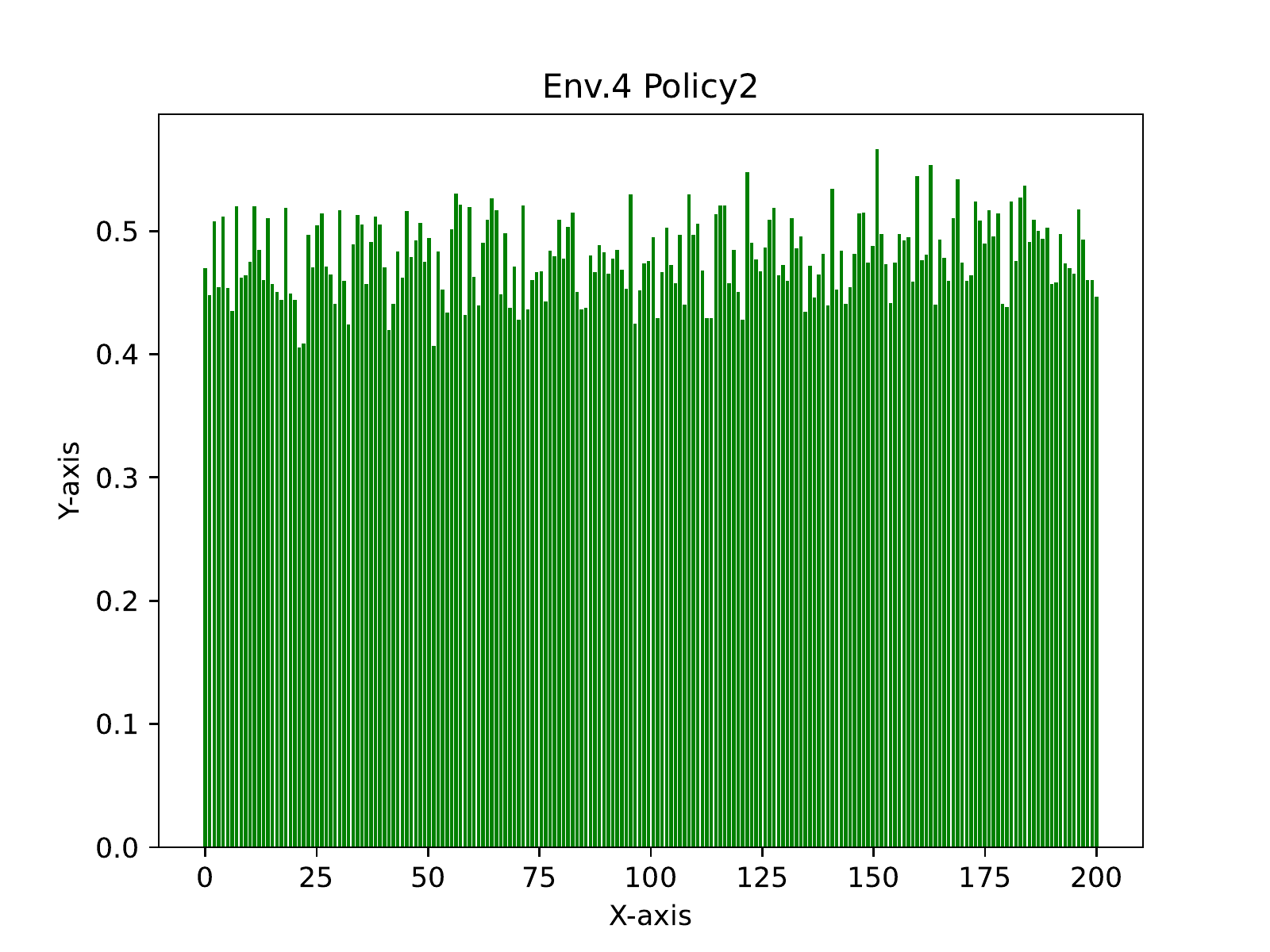}
\includegraphics[width=0.19\textwidth]{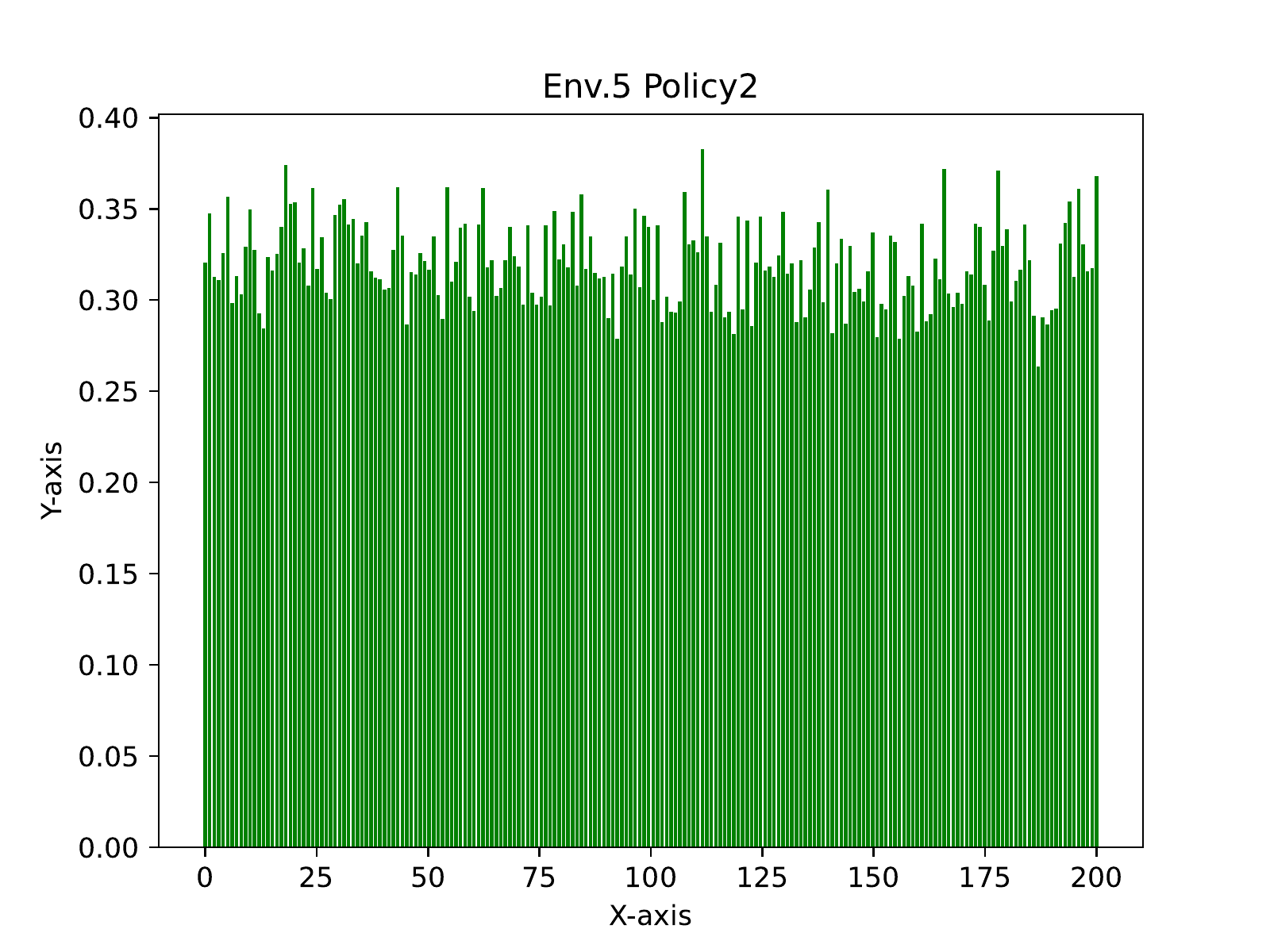}
}
\quad

\subfigure[Policy 3 Env. 1 to Env. 5]{
\includegraphics[width=0.19\textwidth]{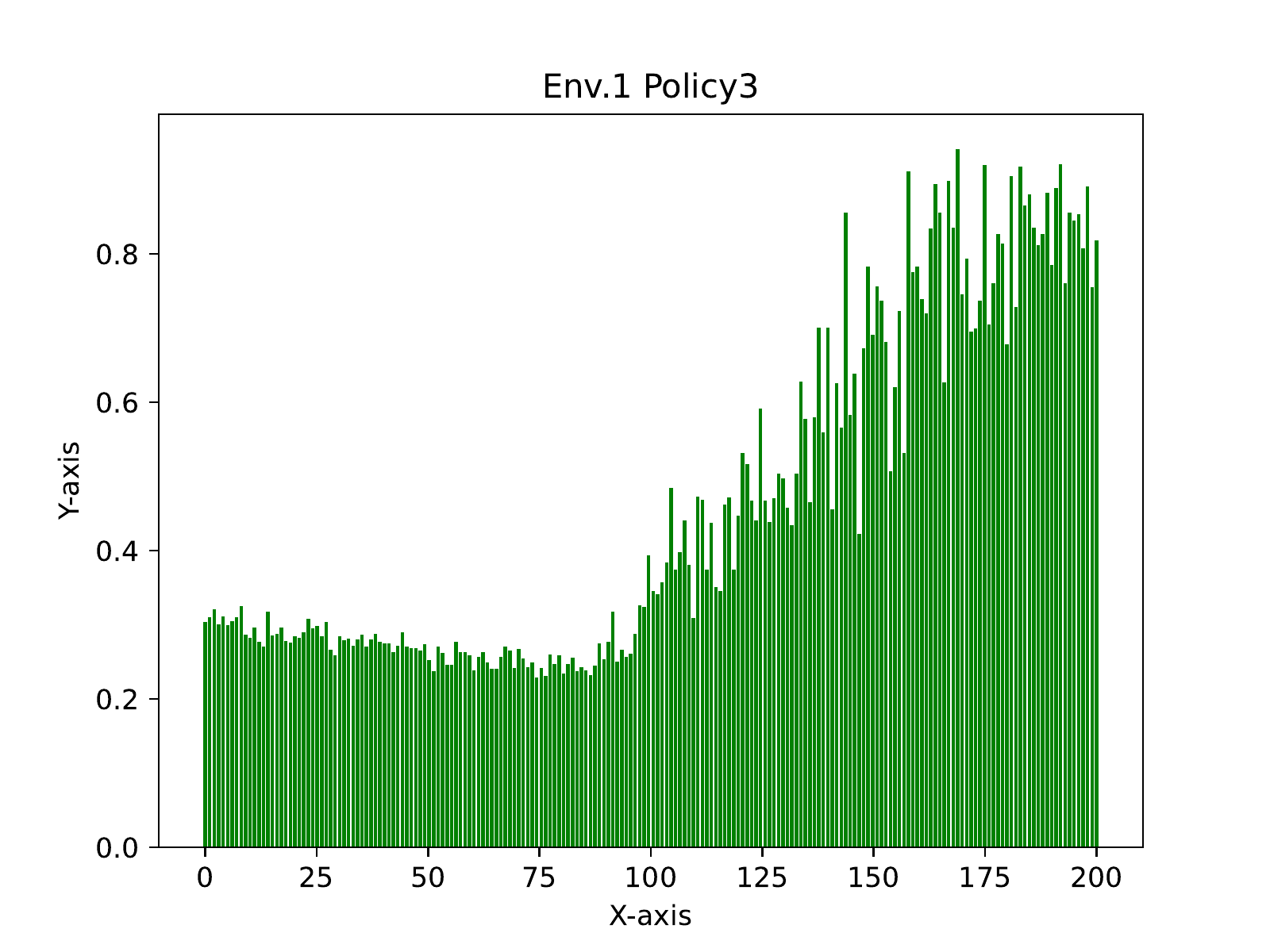}
\includegraphics[width=0.19\textwidth]{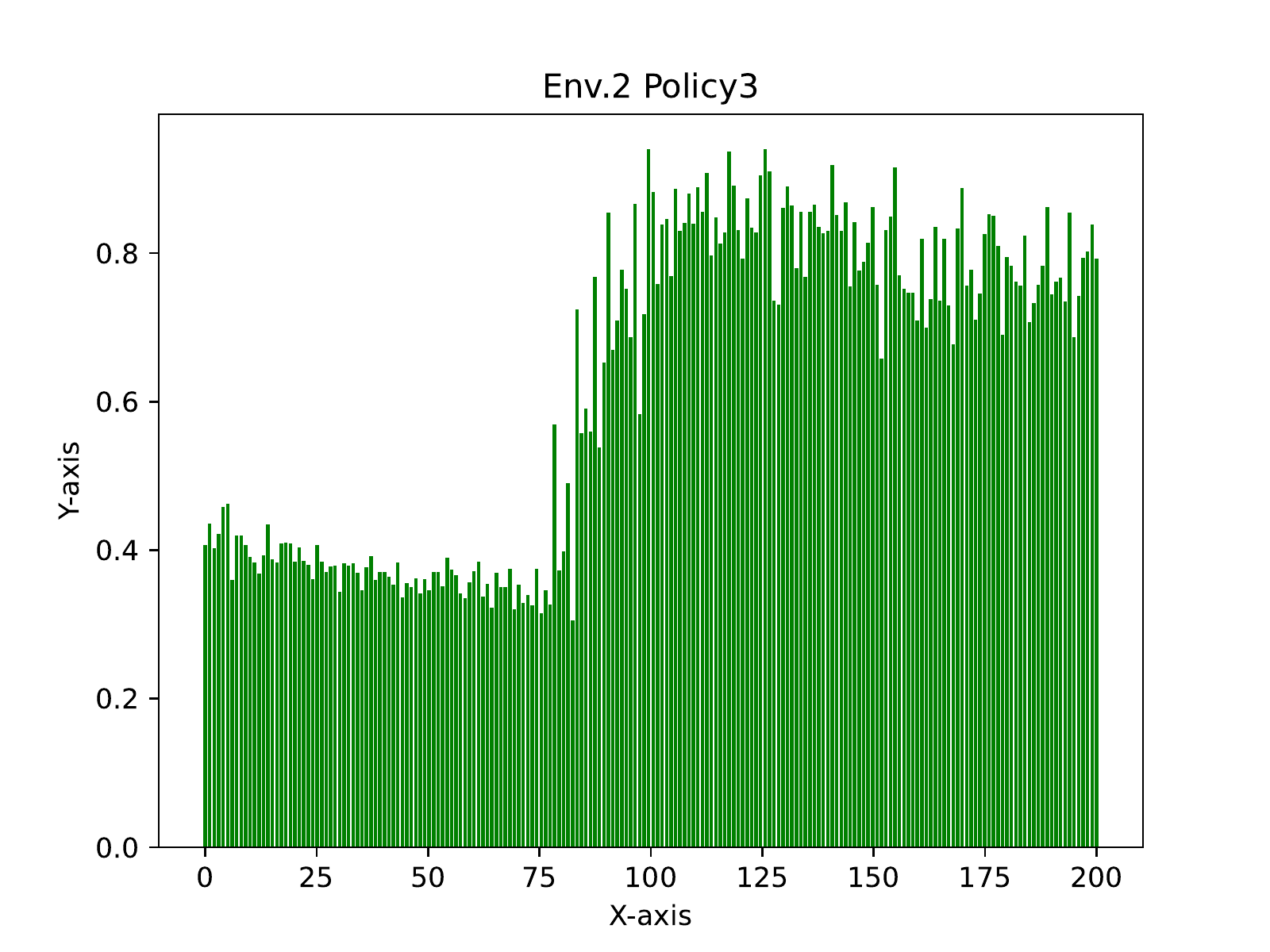}
\includegraphics[width=0.19\textwidth]{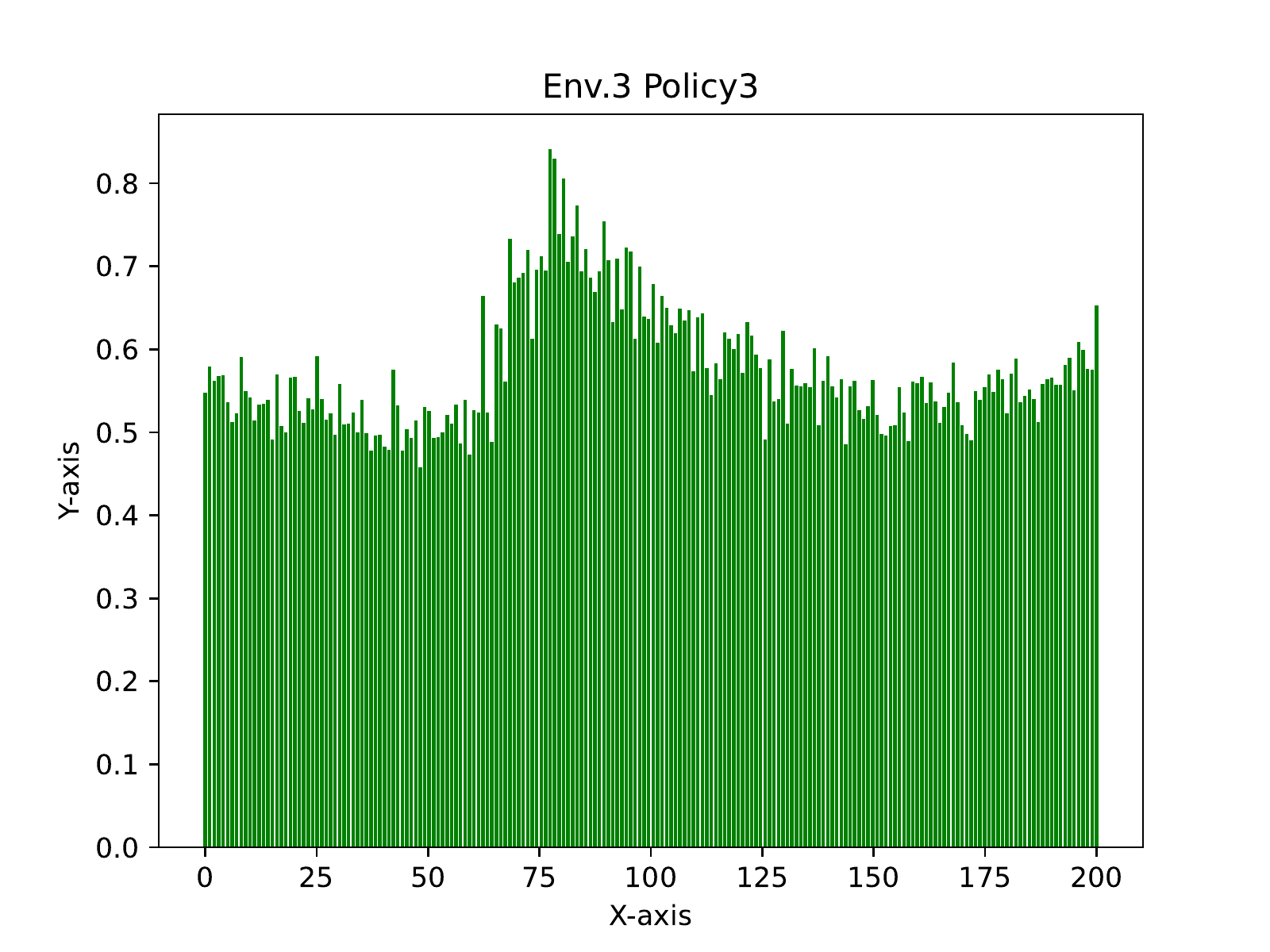}
\includegraphics[width=0.19\textwidth]{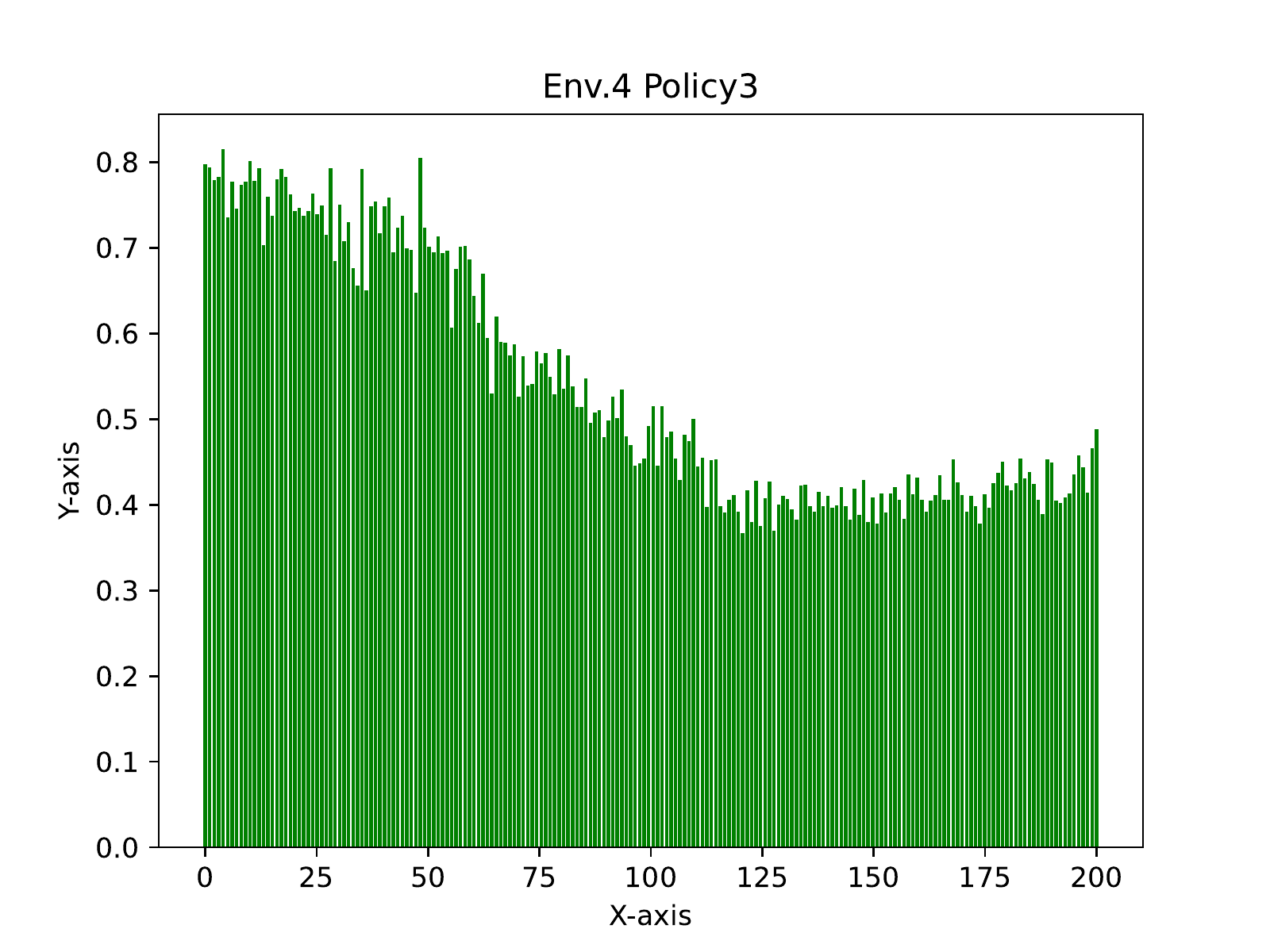}
\includegraphics[width=0.19\textwidth]{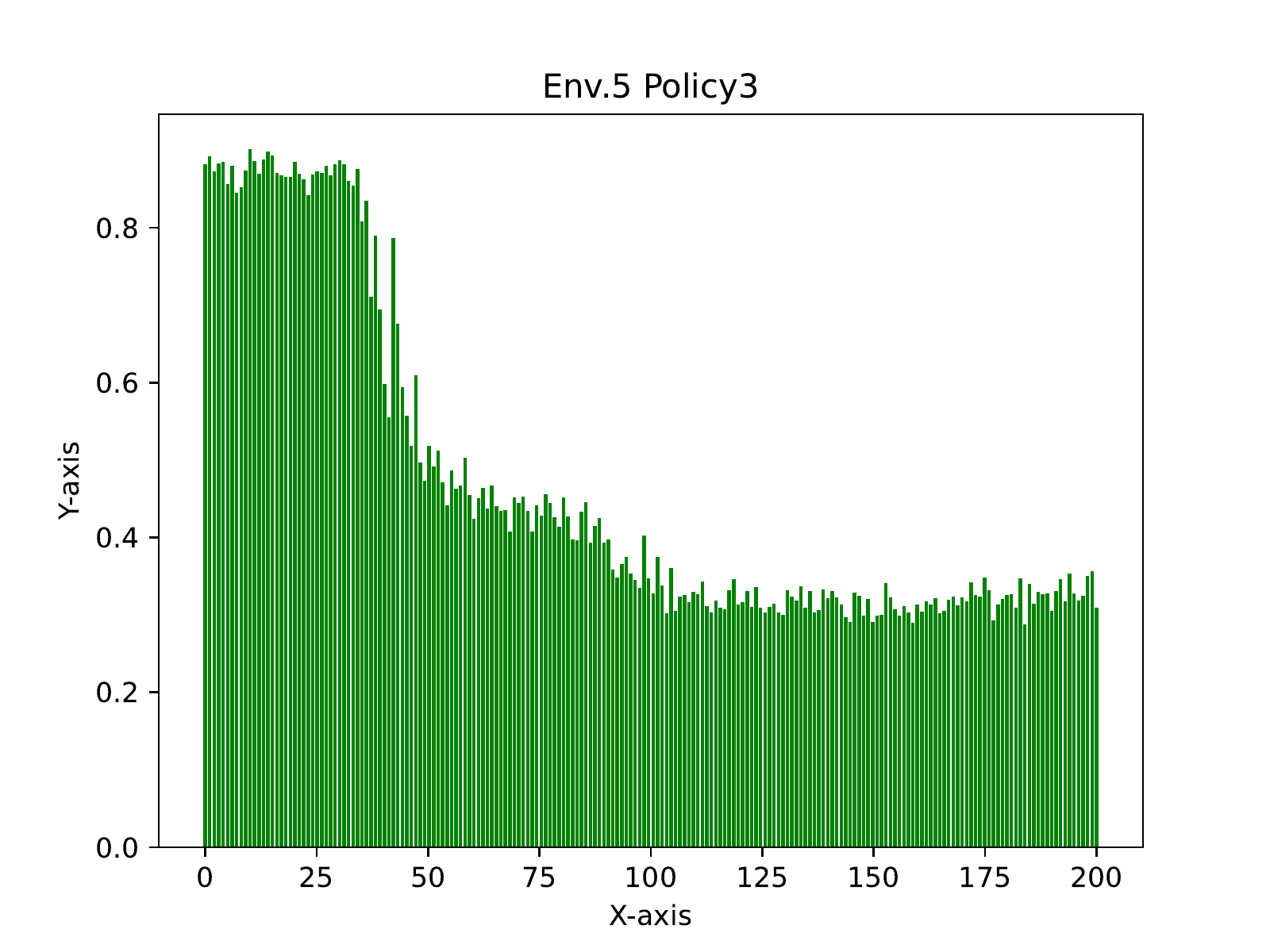}
}
\quad

\subfigure[Policy 4 Env. 1 to Env. 5]{
\includegraphics[width=0.19\textwidth]{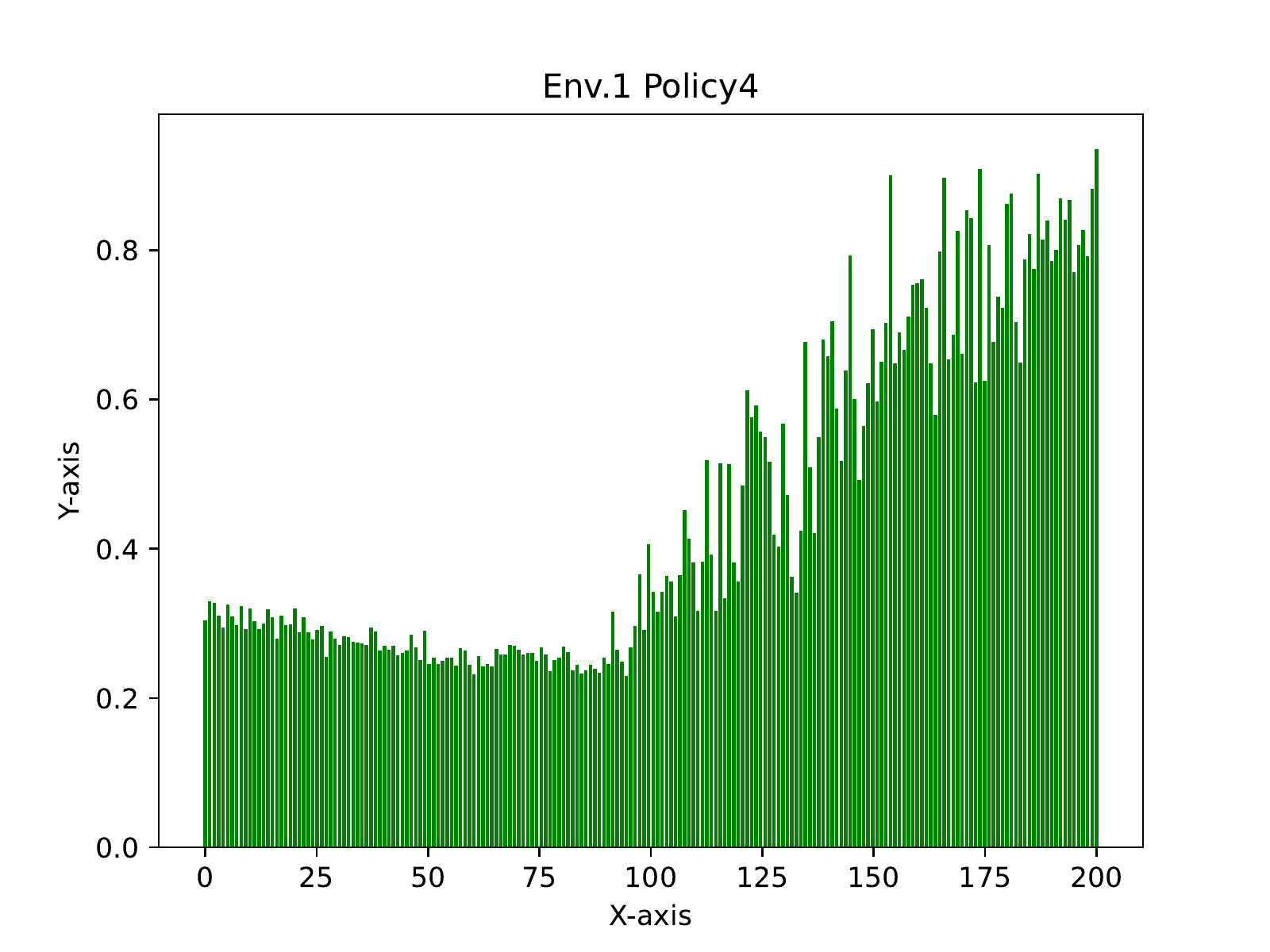}
\includegraphics[width=0.19\textwidth]{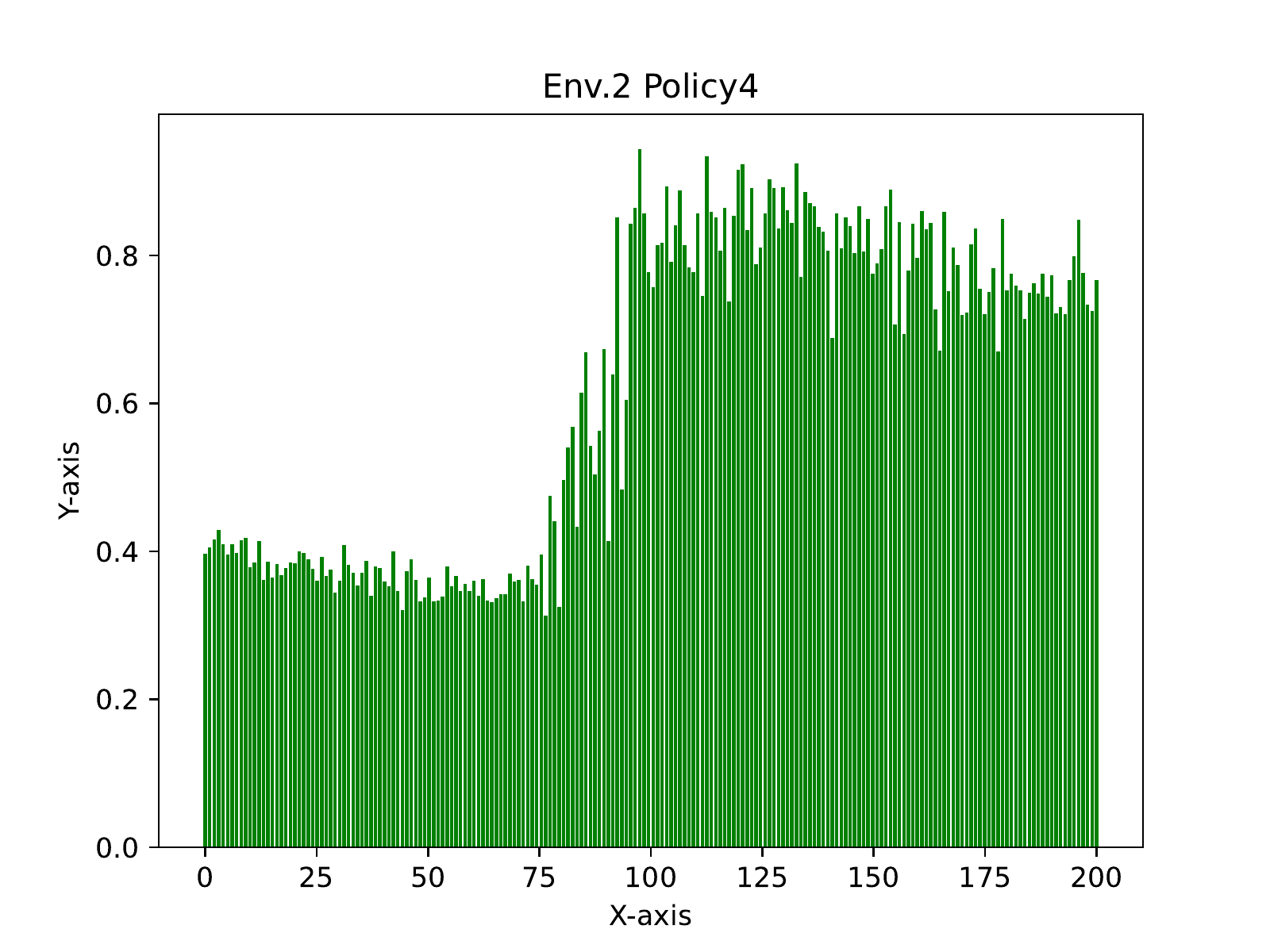}
\includegraphics[width=0.19\textwidth]{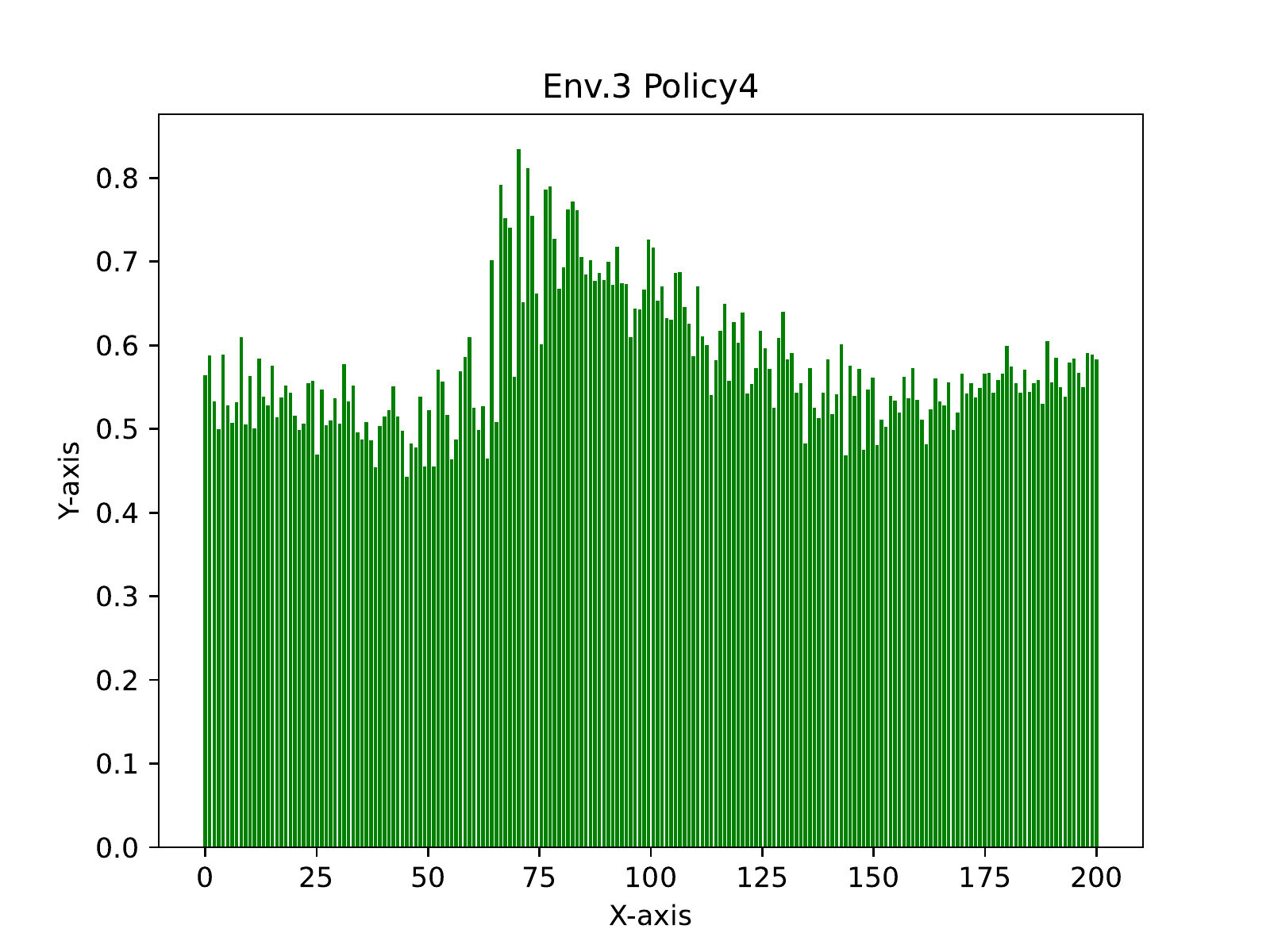}
\includegraphics[width=0.19\textwidth]{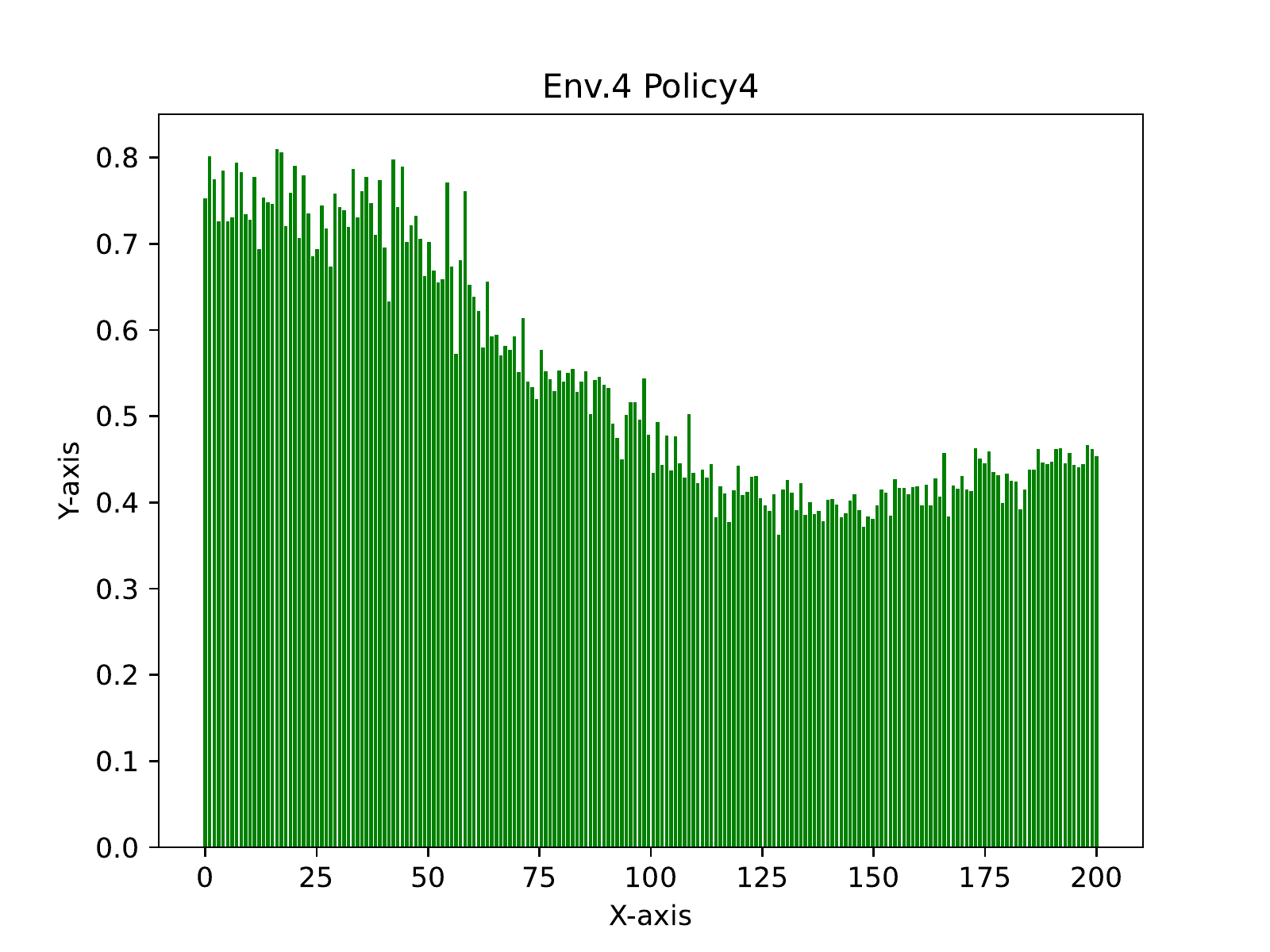}
\includegraphics[width=0.19\textwidth]{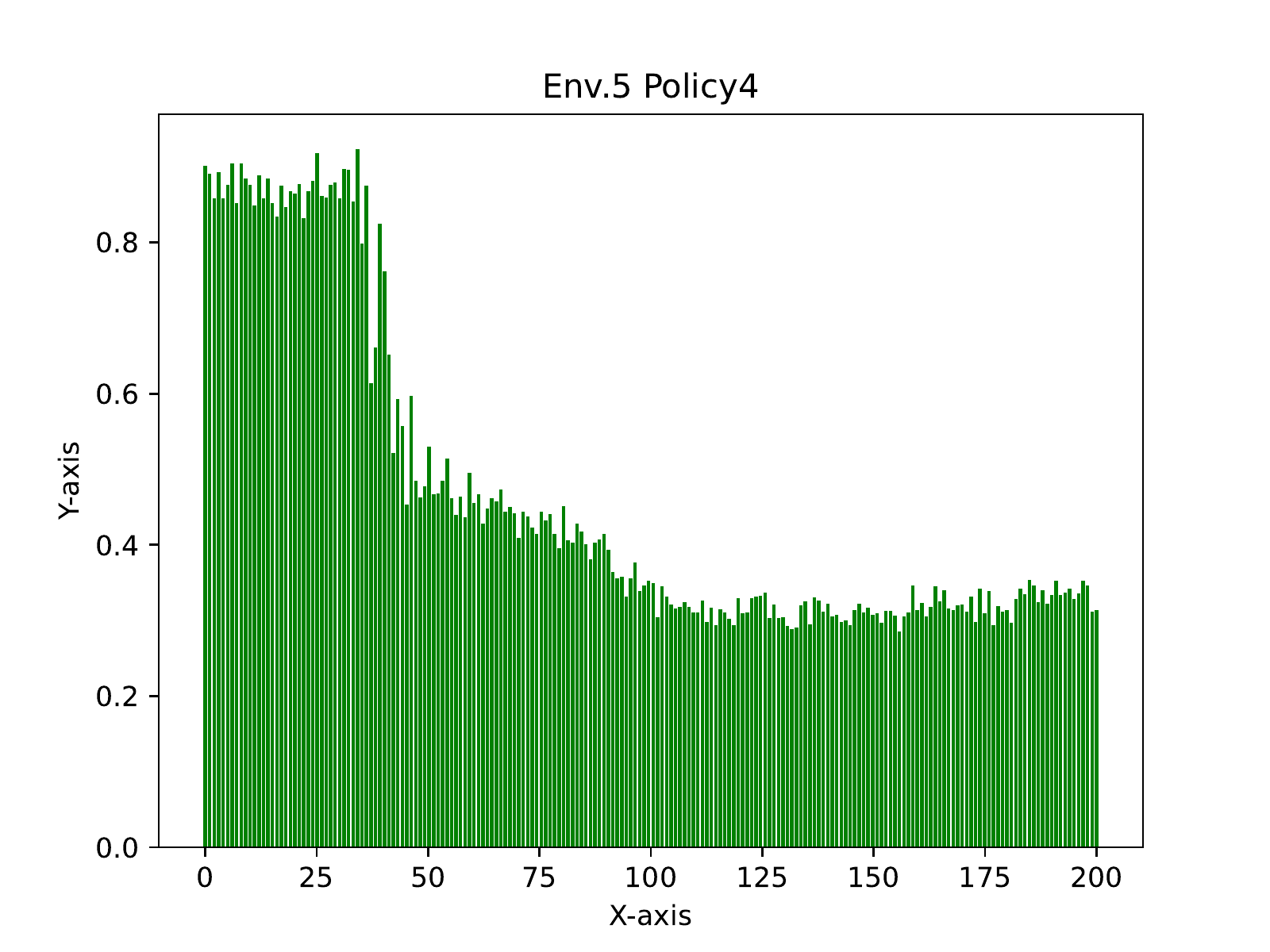}
}
\quad

\subfigure[Policy 5 Env. 1 to Env. 5]{
\includegraphics[width=0.19\textwidth]{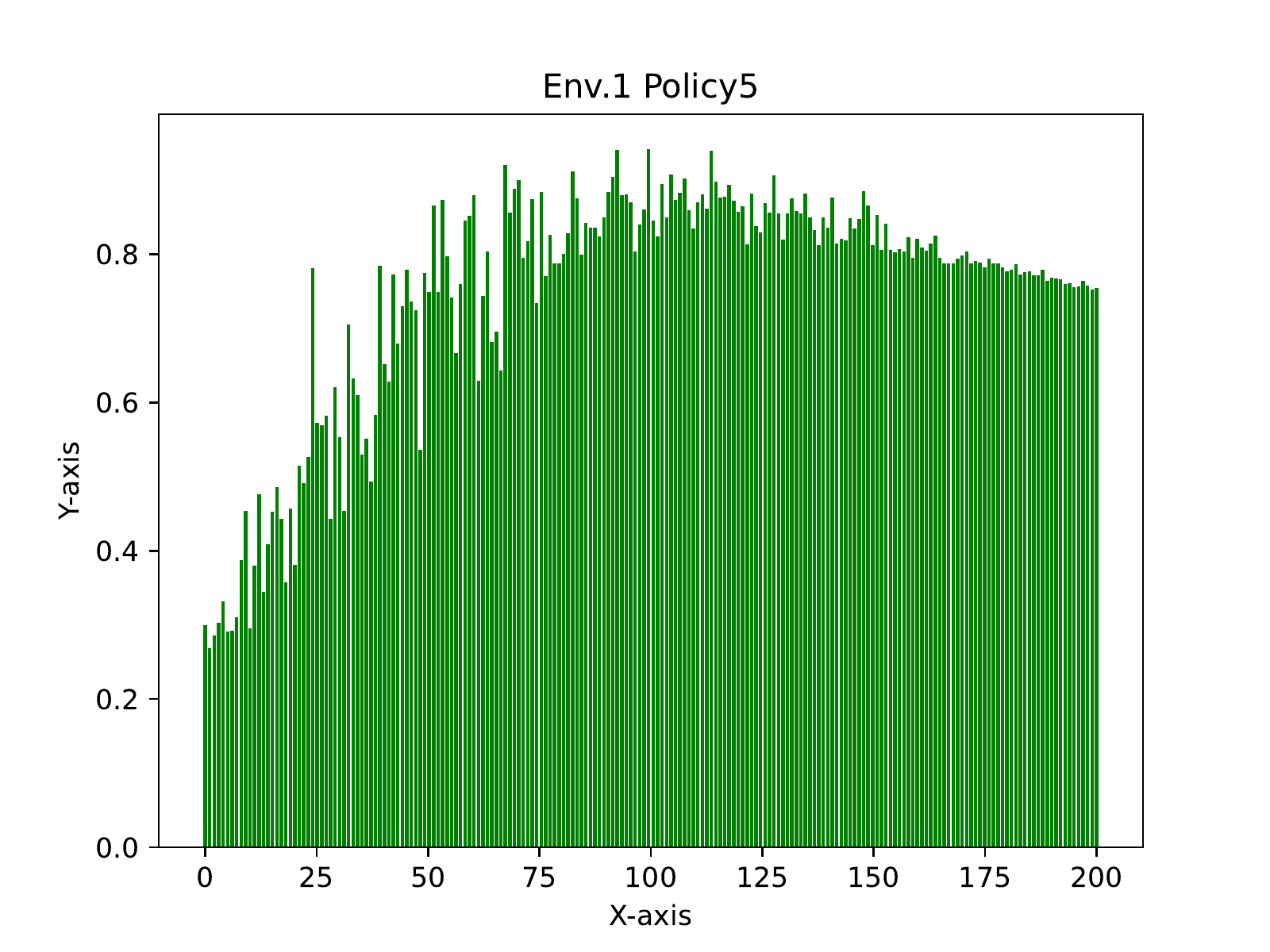}
\includegraphics[width=0.19\textwidth]{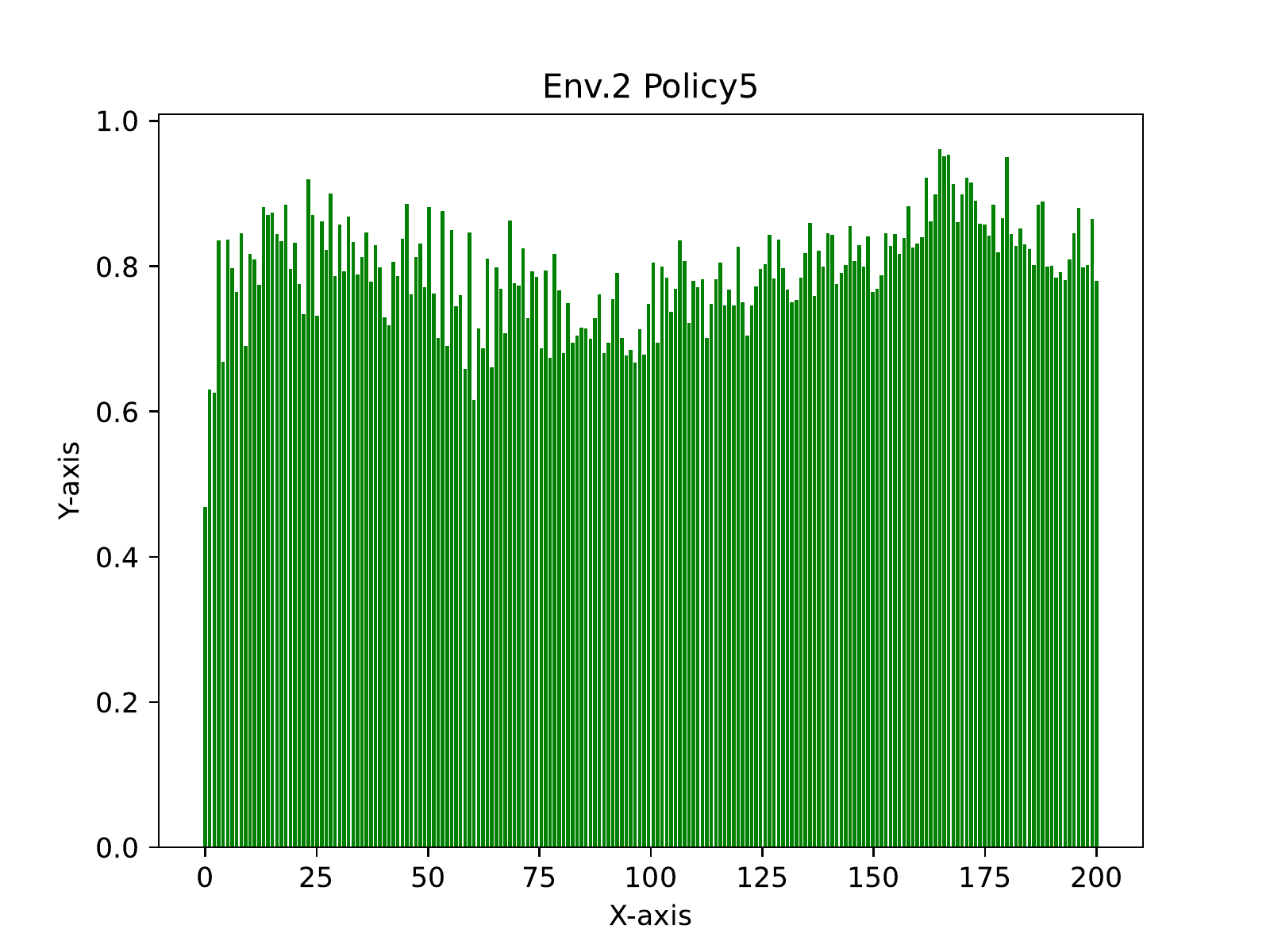}
\includegraphics[width=0.19\textwidth]{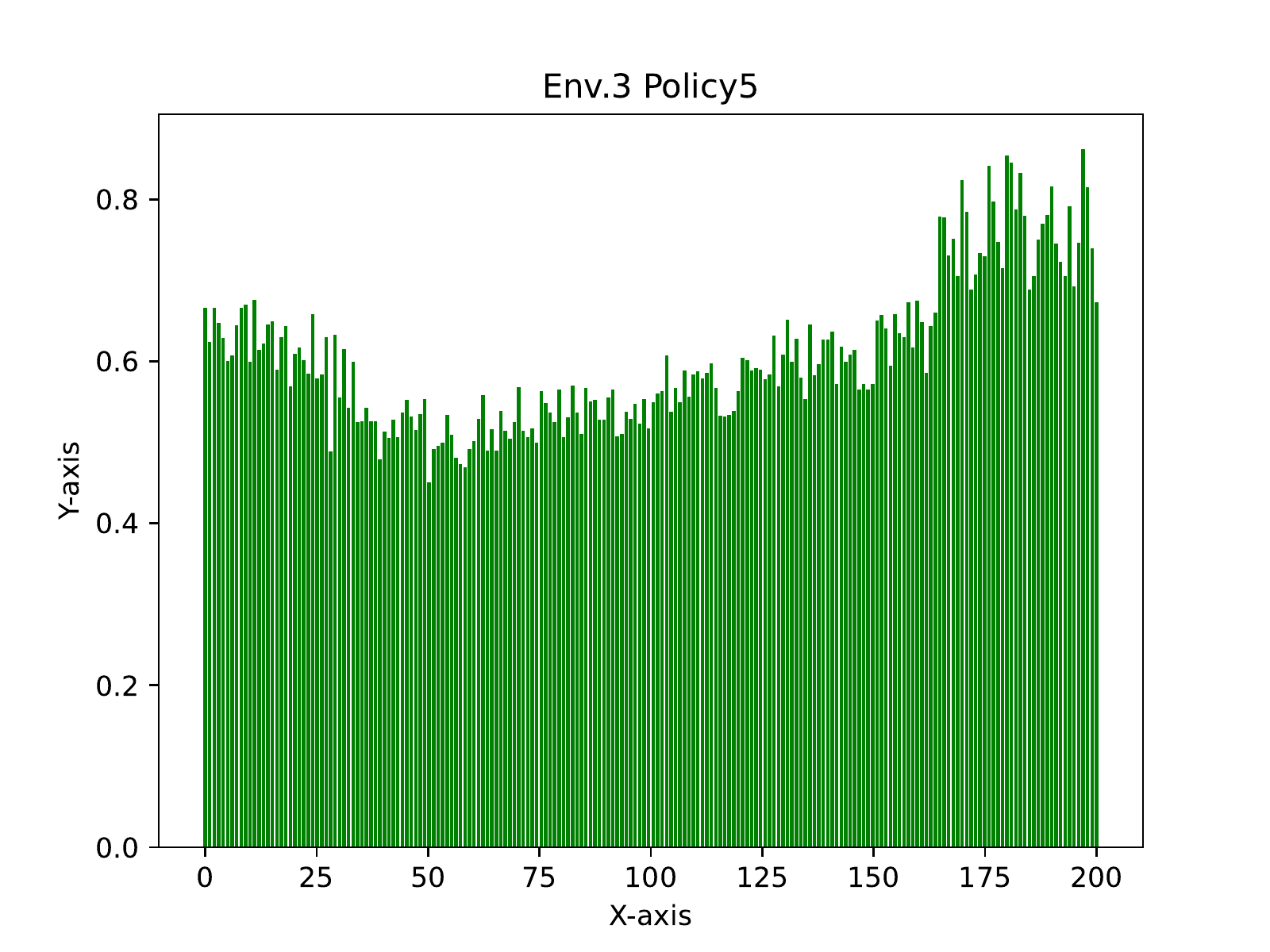}
\includegraphics[width=0.19\textwidth]{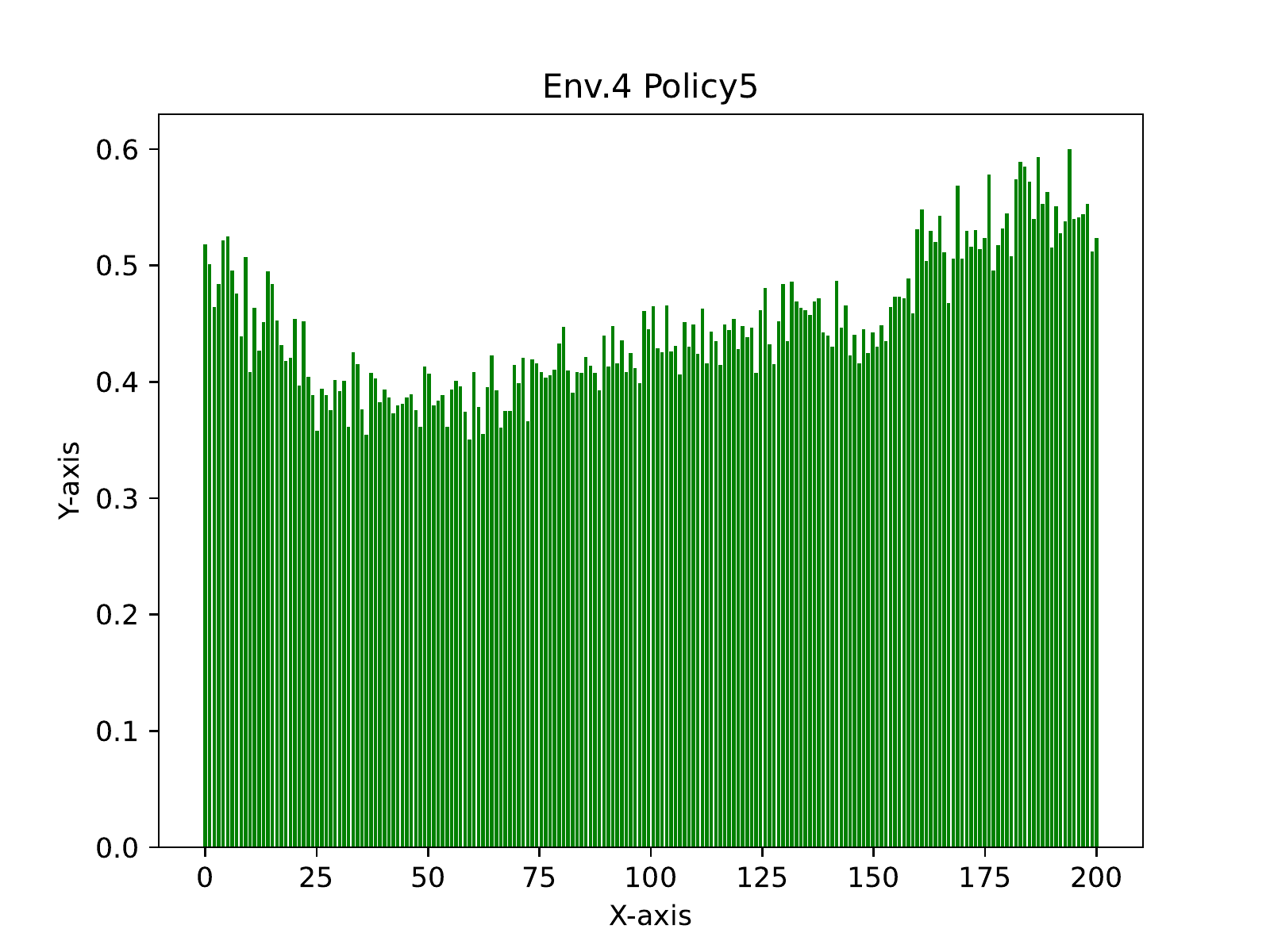}
\includegraphics[width=0.19\textwidth]{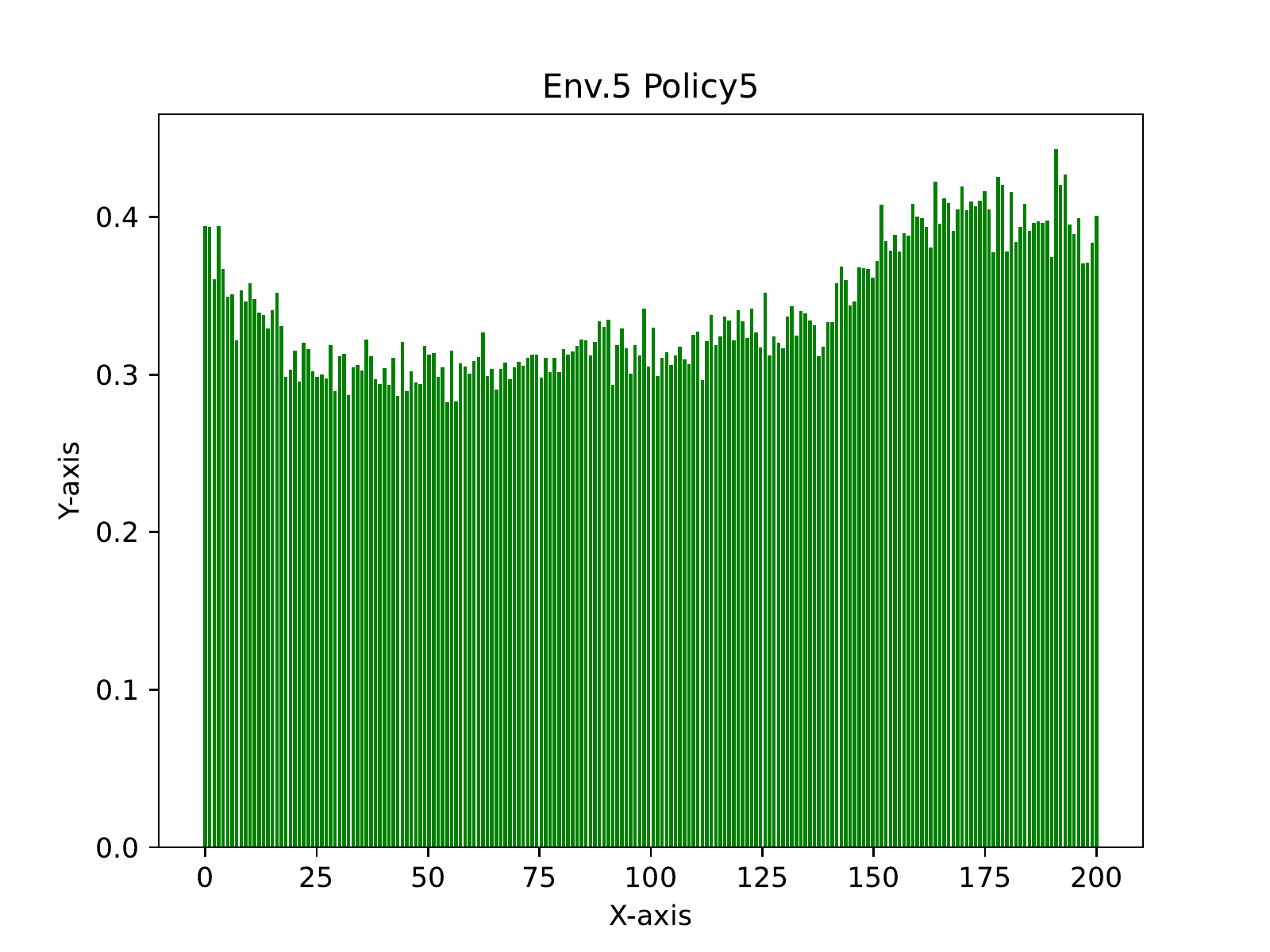}
}
\quad

\subfigure[Policy 6 Env. 1 to Env. 5]{
\includegraphics[width=0.19\textwidth]{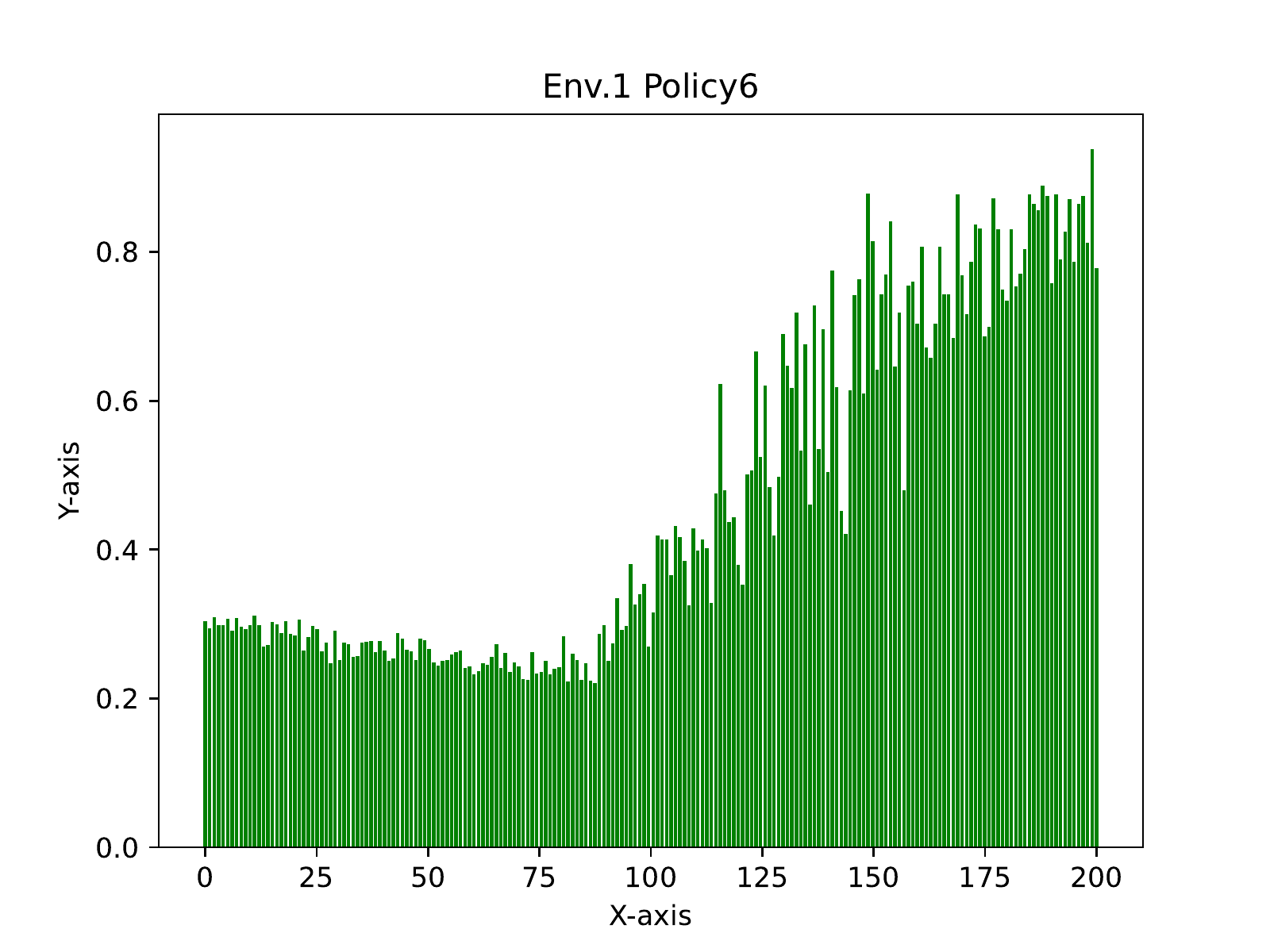}
\includegraphics[width=0.19\textwidth]{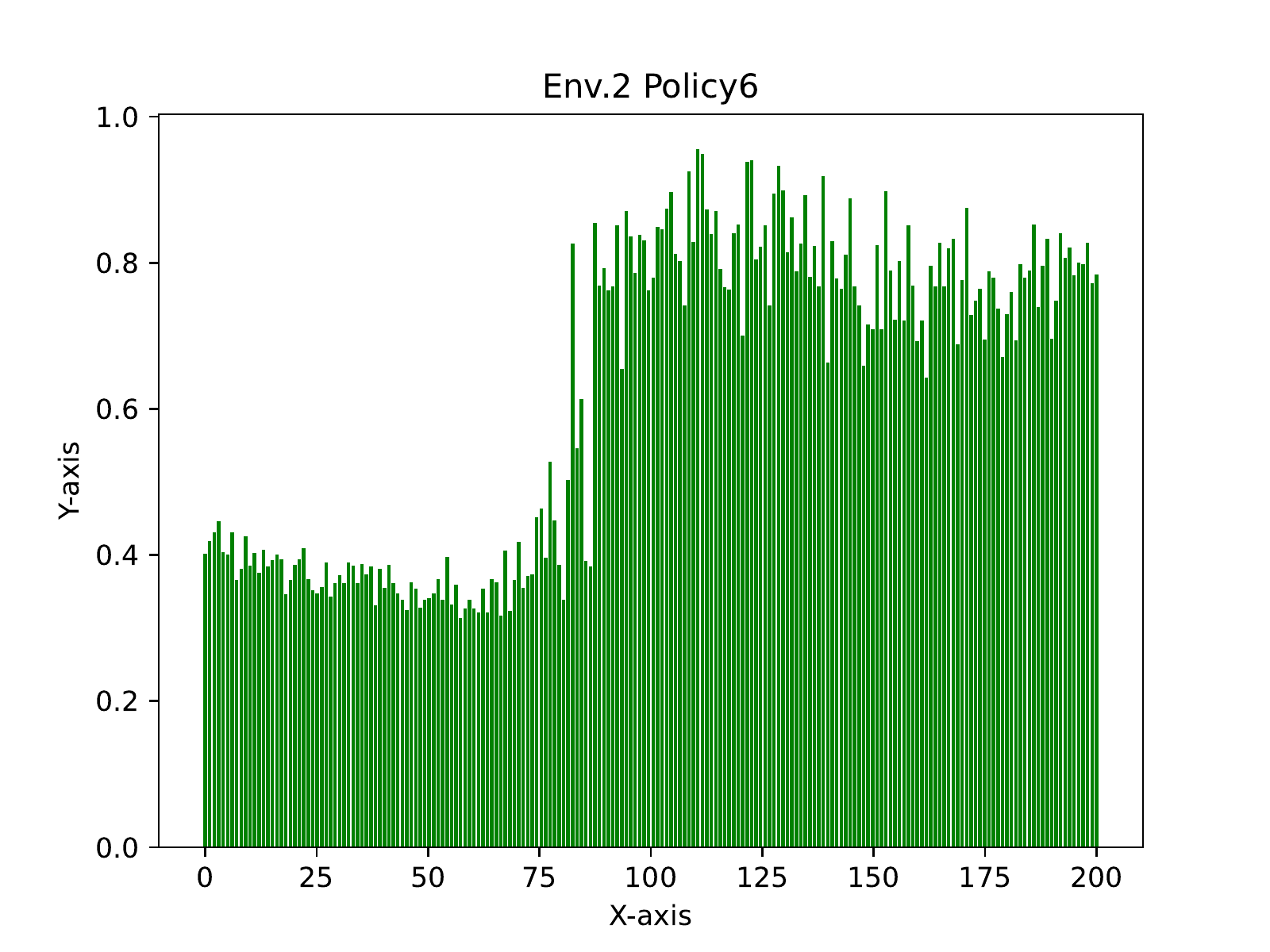}
\includegraphics[width=0.19\textwidth]{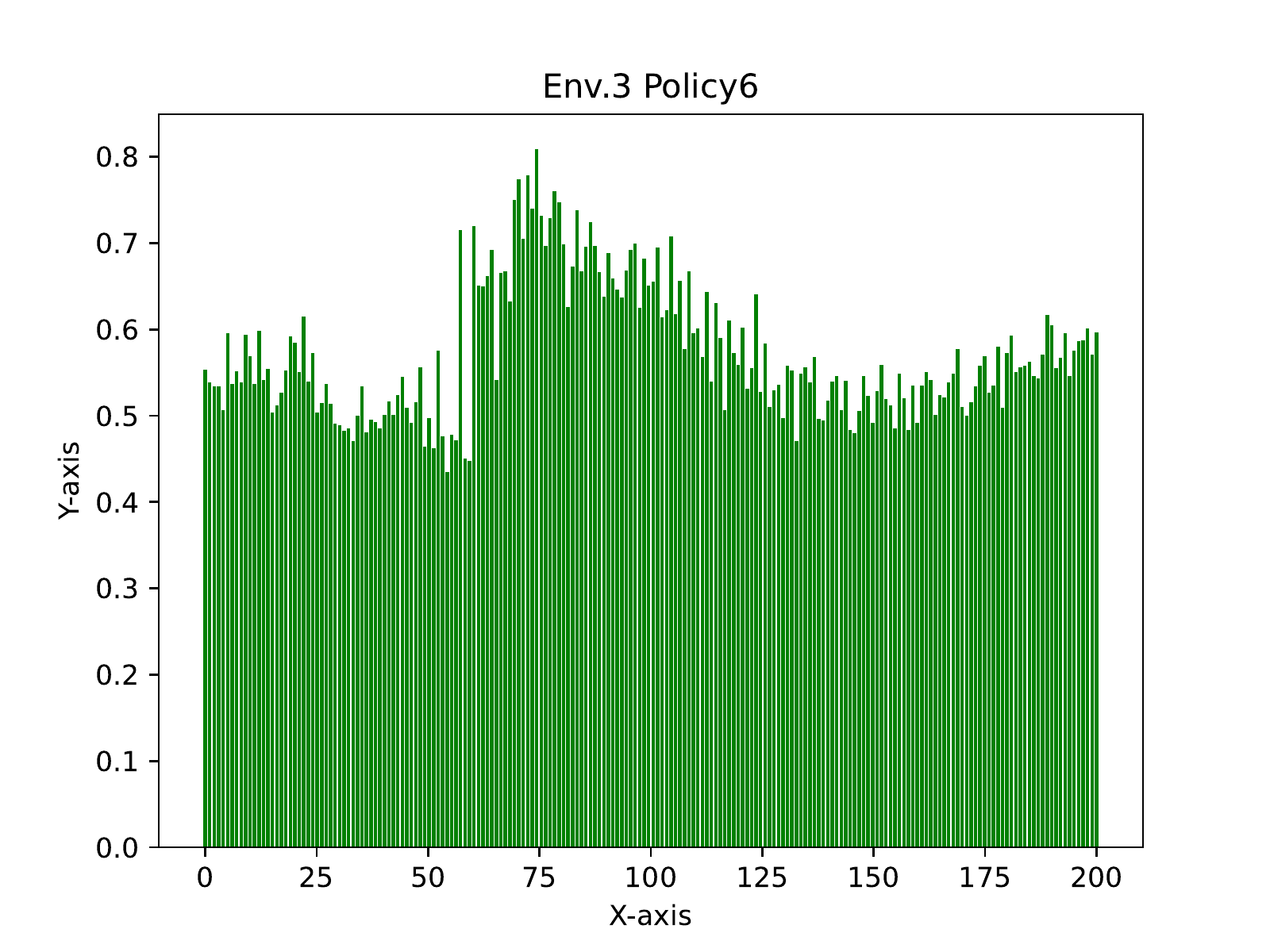}
\includegraphics[width=0.19\textwidth]{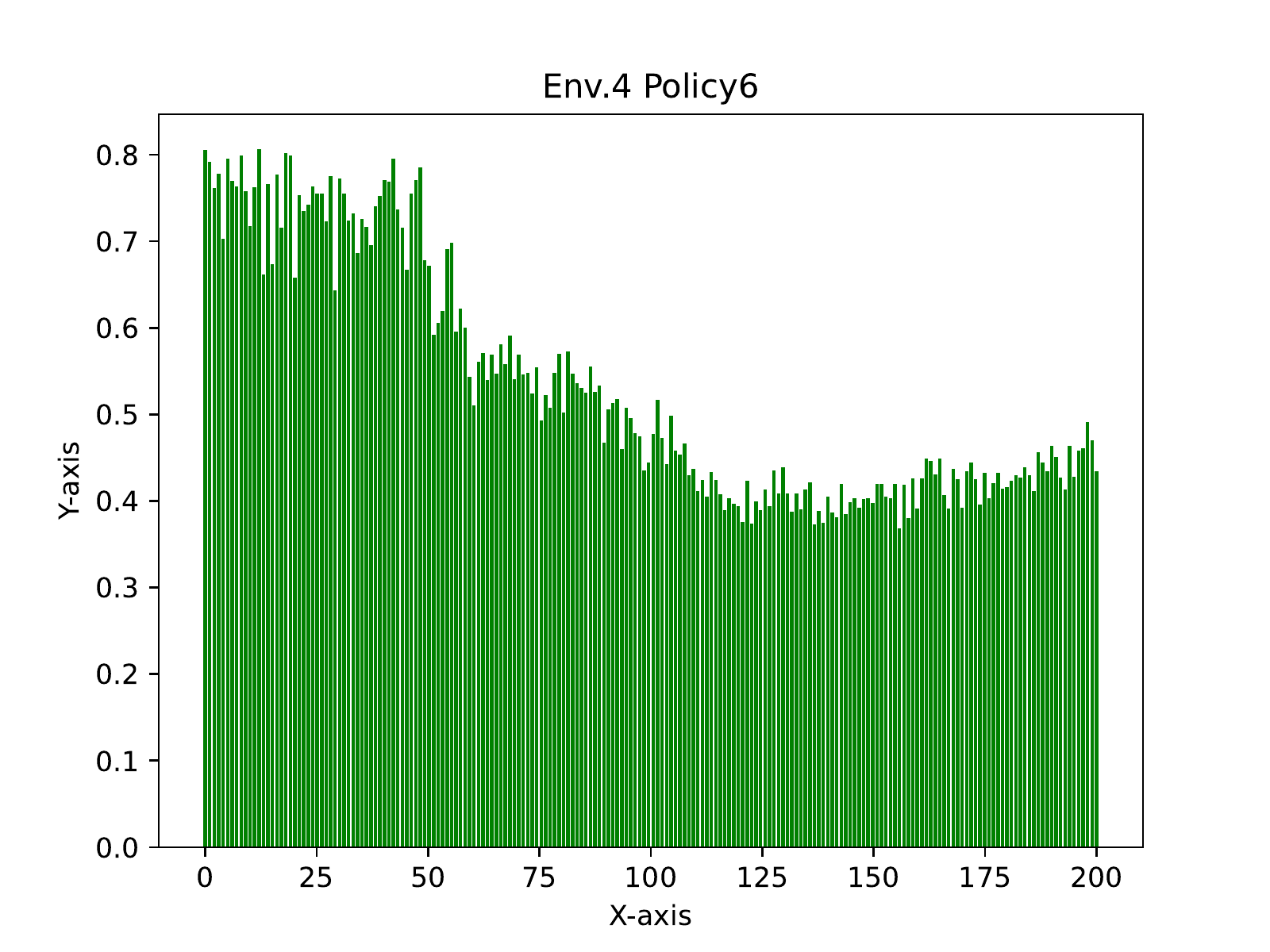}
\includegraphics[width=0.19\textwidth]{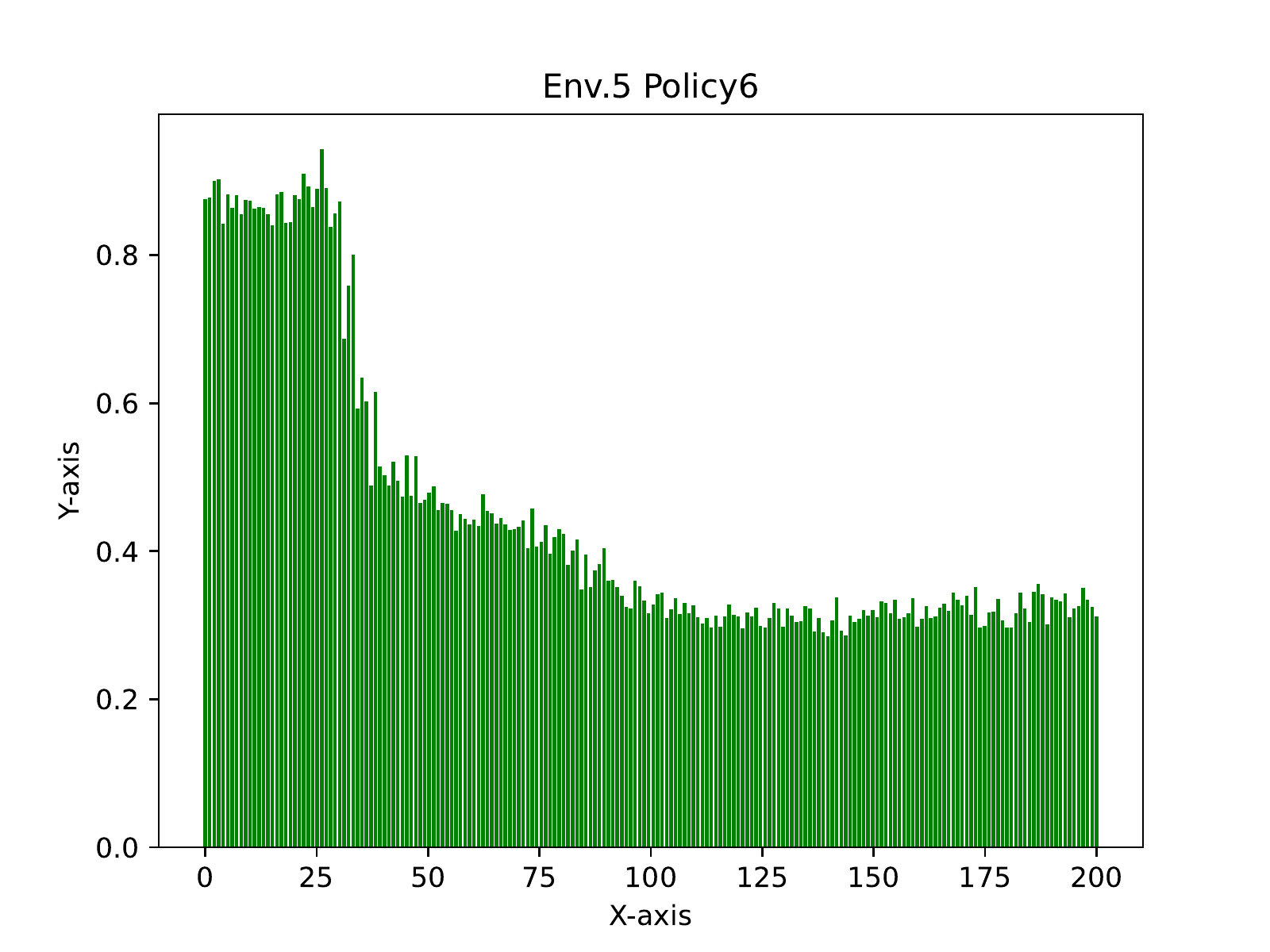}
}
\quad

\caption{Episodic MC return distribution of adapted policy during GA process.}
\label{fig::Para Tune Space 1}
\end{figure*}

\begin{figure*}[t]
\addtocounter{subfigure}{6}
\ContinuedFloat
\centering
    
\subfigure[Policy 7 Env. 1 to Env. 5]{
\includegraphics[width=0.19\textwidth]{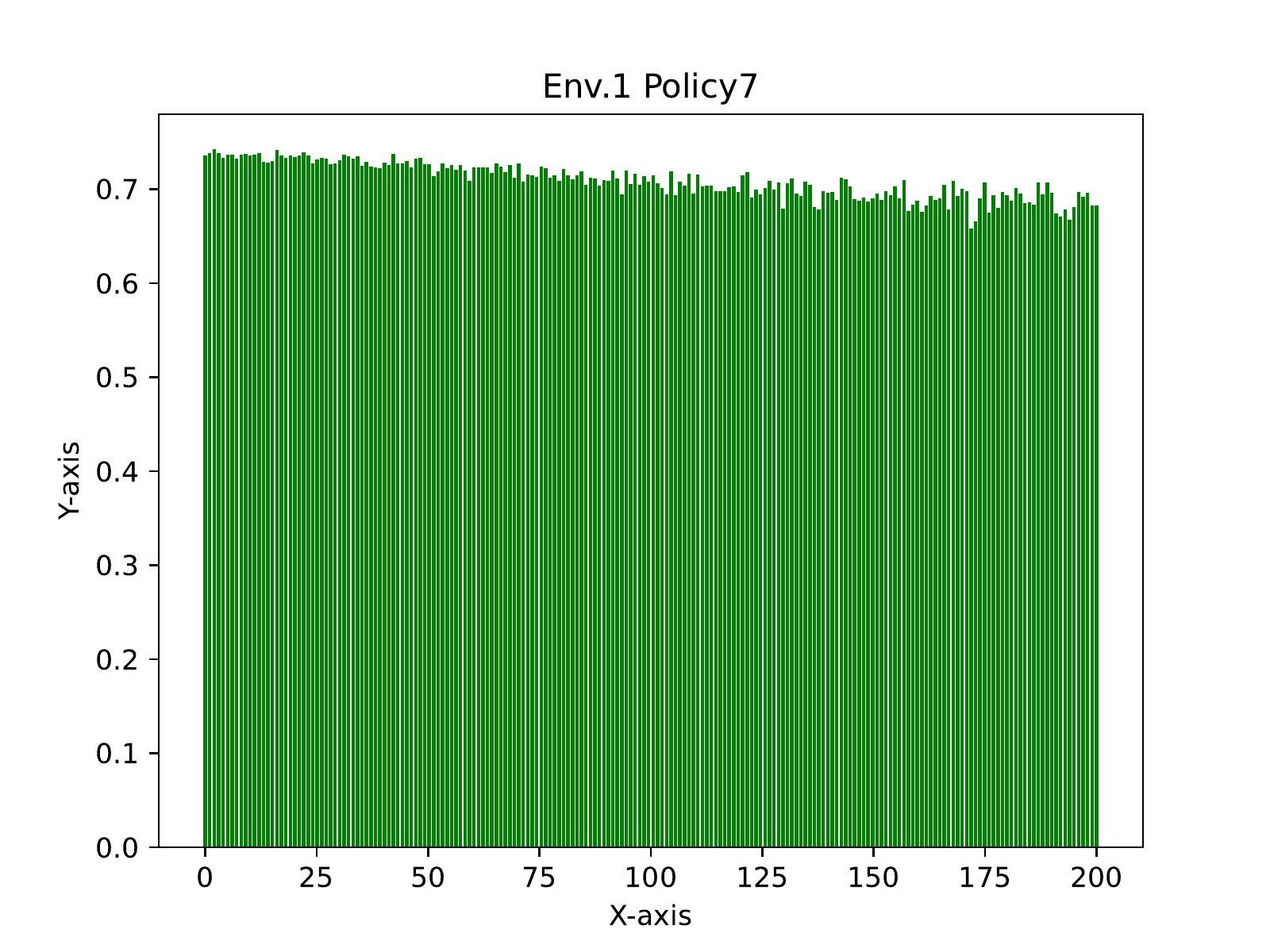}
\includegraphics[width=0.19\textwidth]{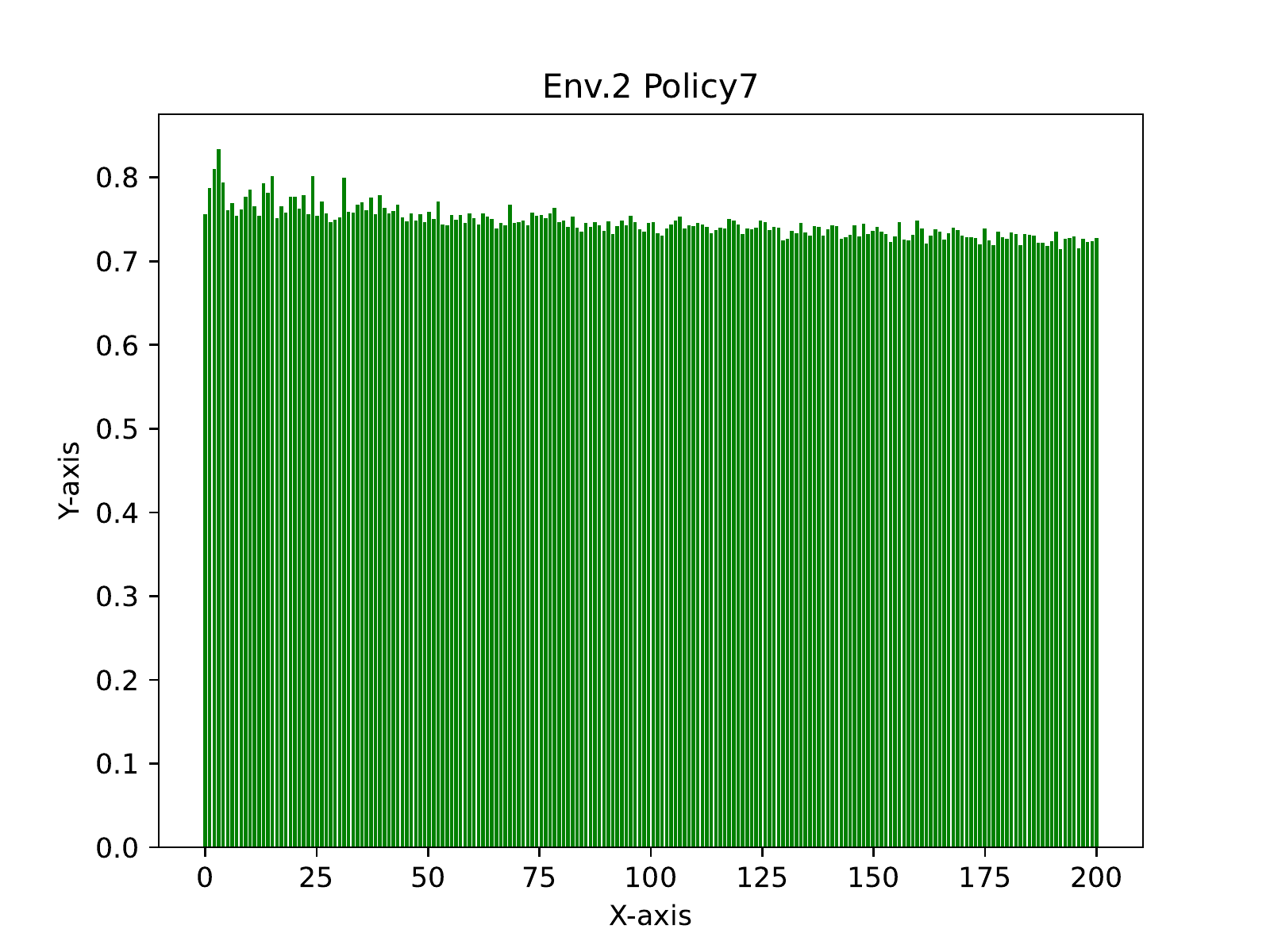}
\includegraphics[width=0.19\textwidth]{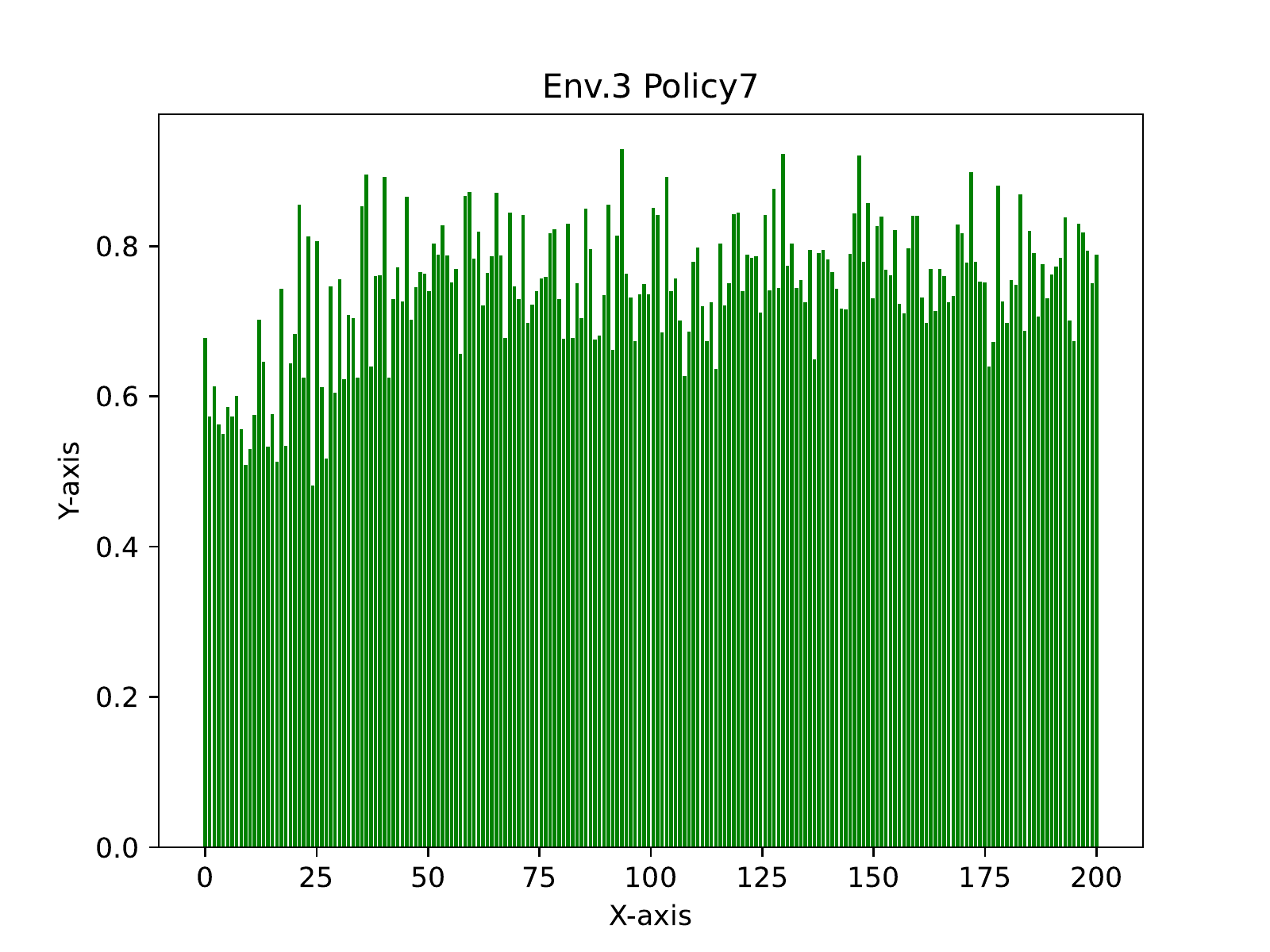}
\includegraphics[width=0.19\textwidth]{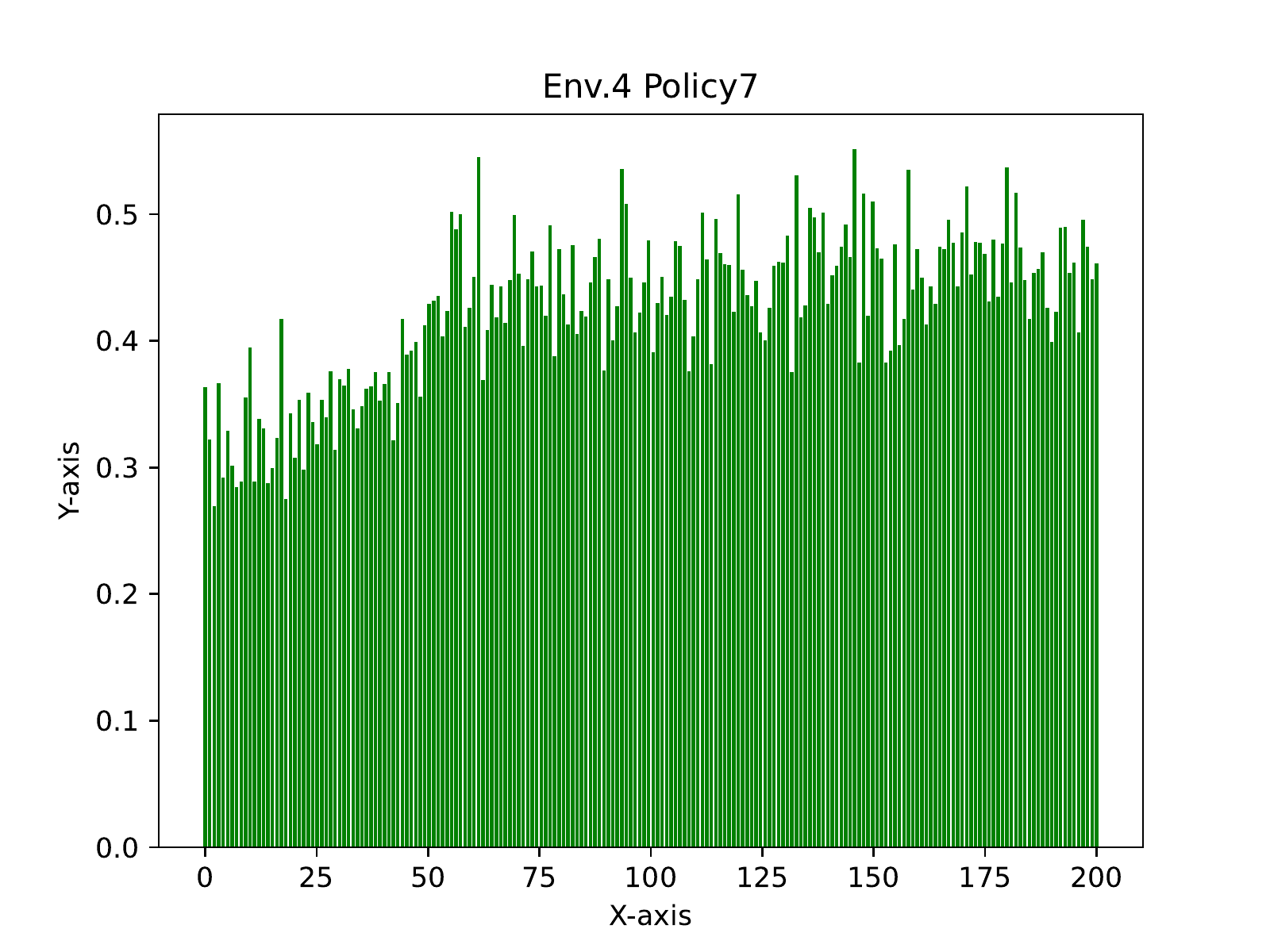}
\includegraphics[width=0.19\textwidth]{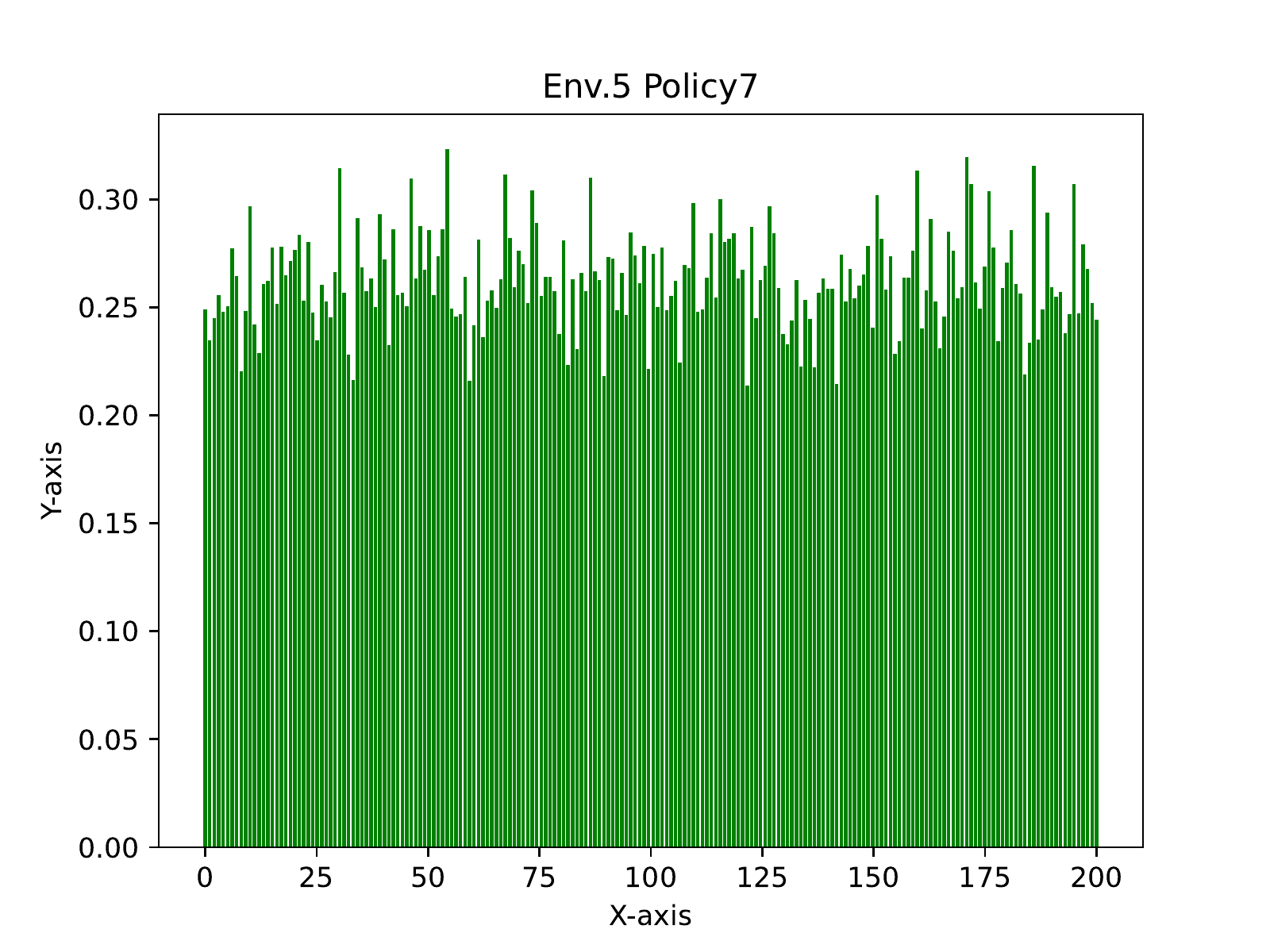}
}
\quad

\subfigure[Policy 8 Env. 1 to Env. 5]{
\includegraphics[width=0.19\textwidth]{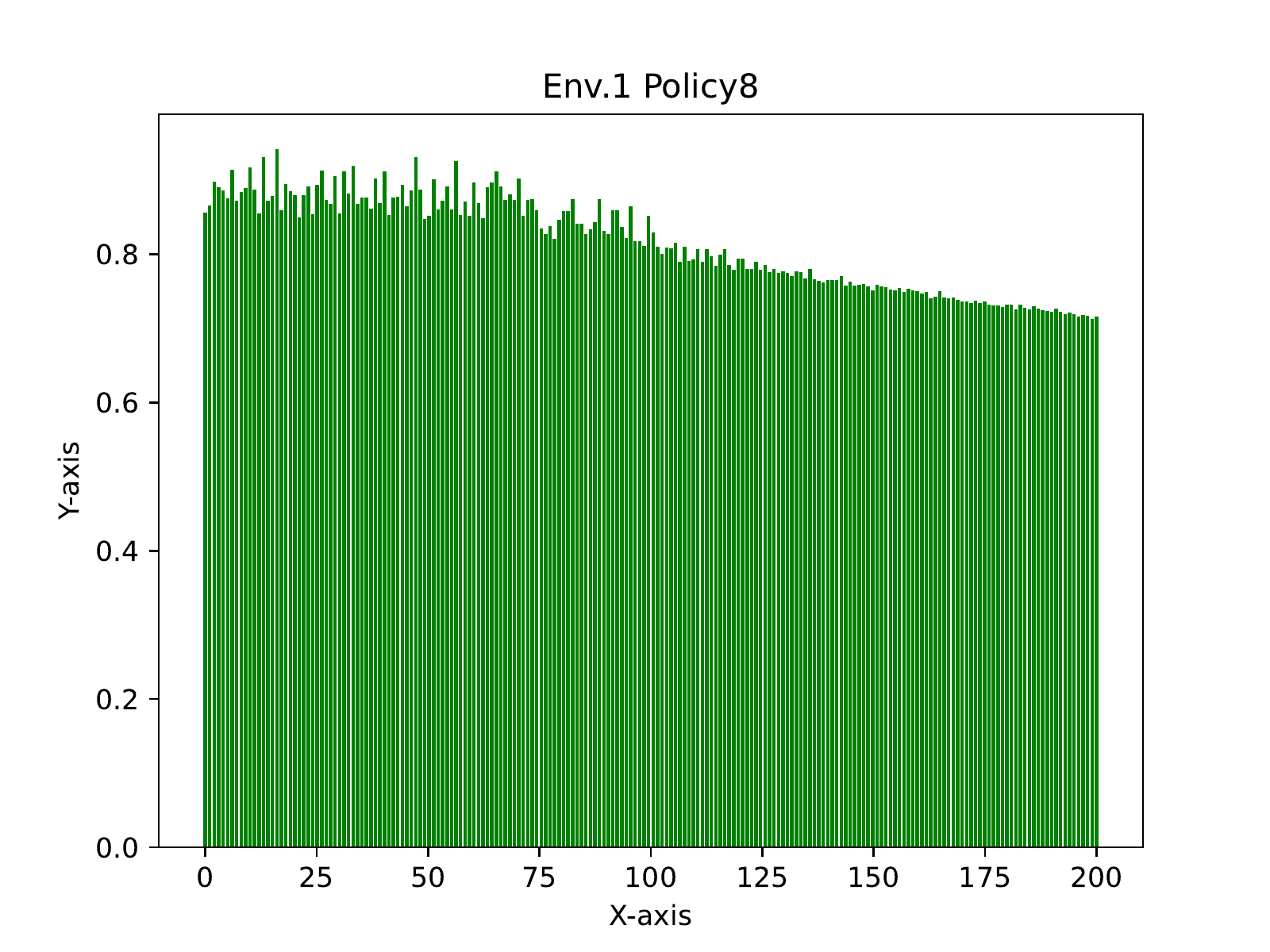}
\includegraphics[width=0.19\textwidth]{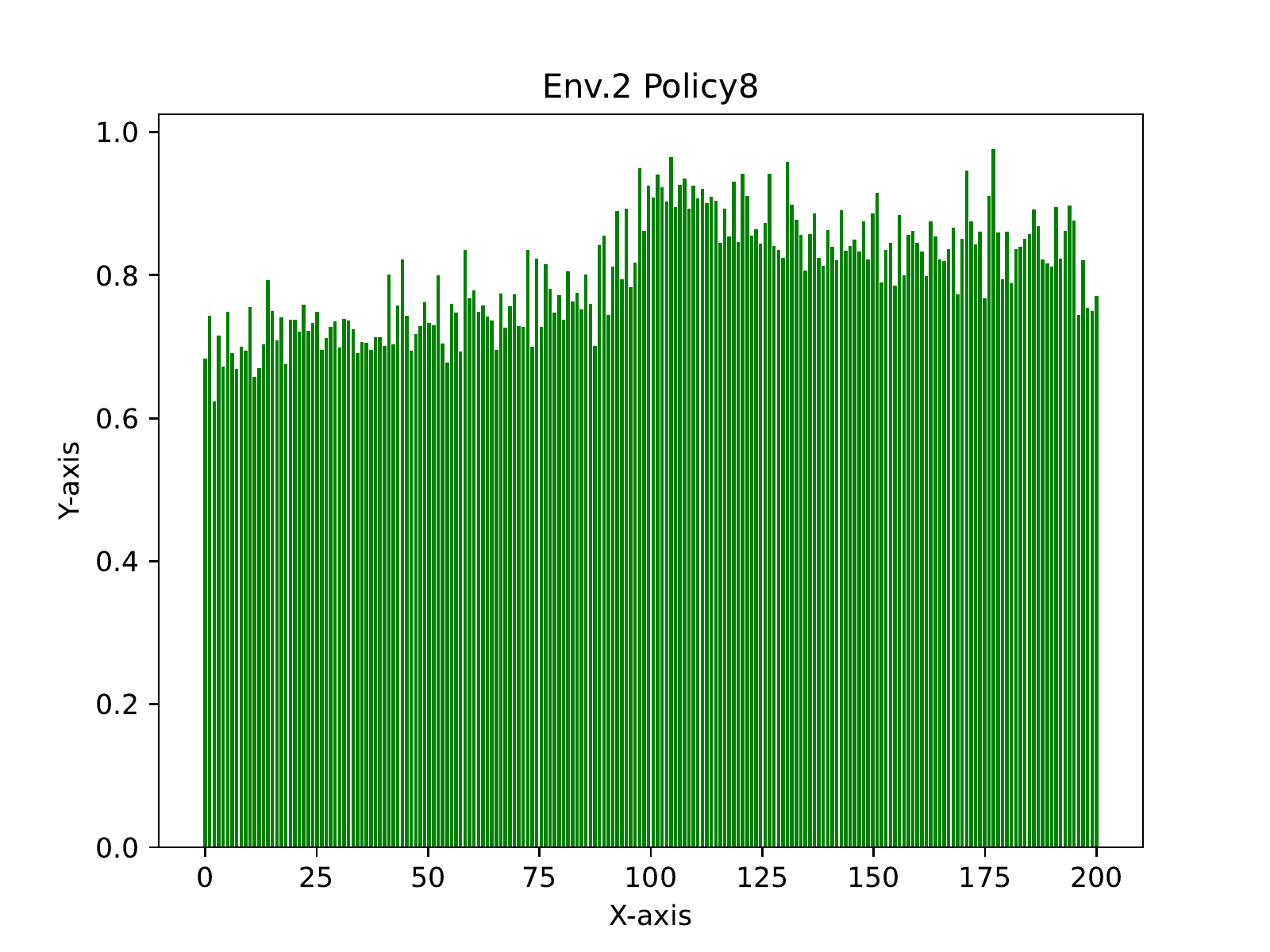}
\includegraphics[width=0.19\textwidth]{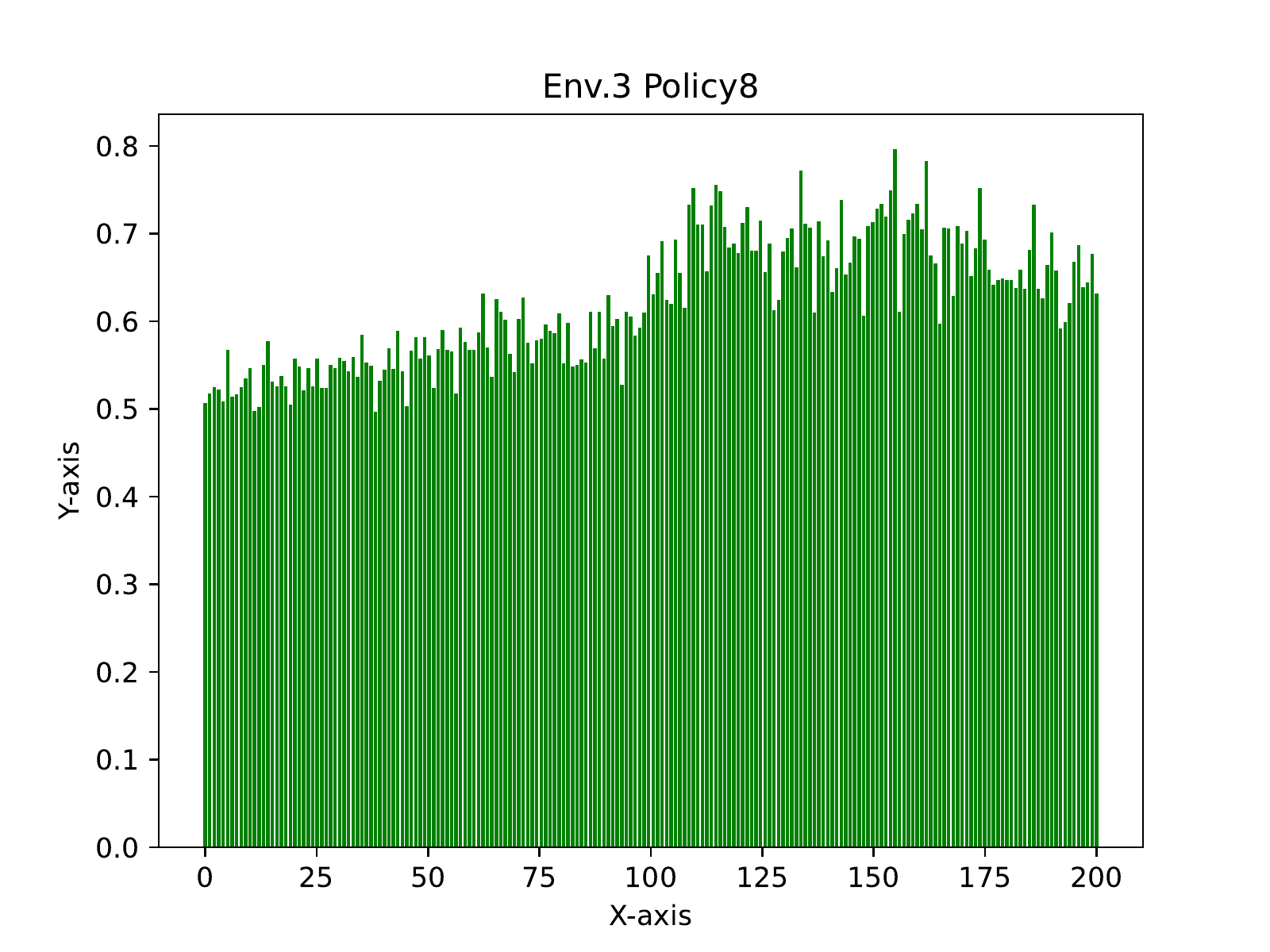}
\includegraphics[width=0.19\textwidth]{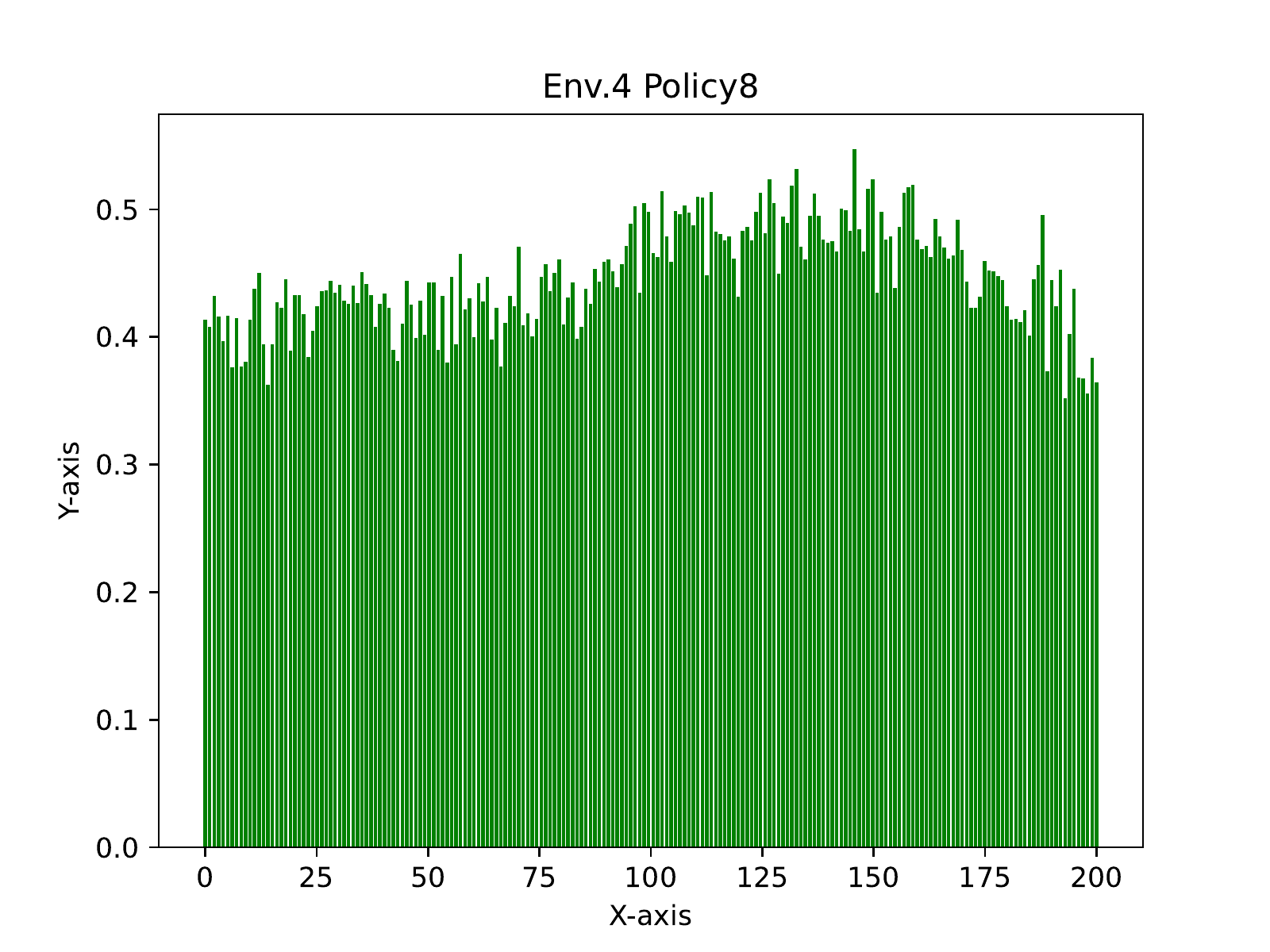}
\includegraphics[width=0.19\textwidth]{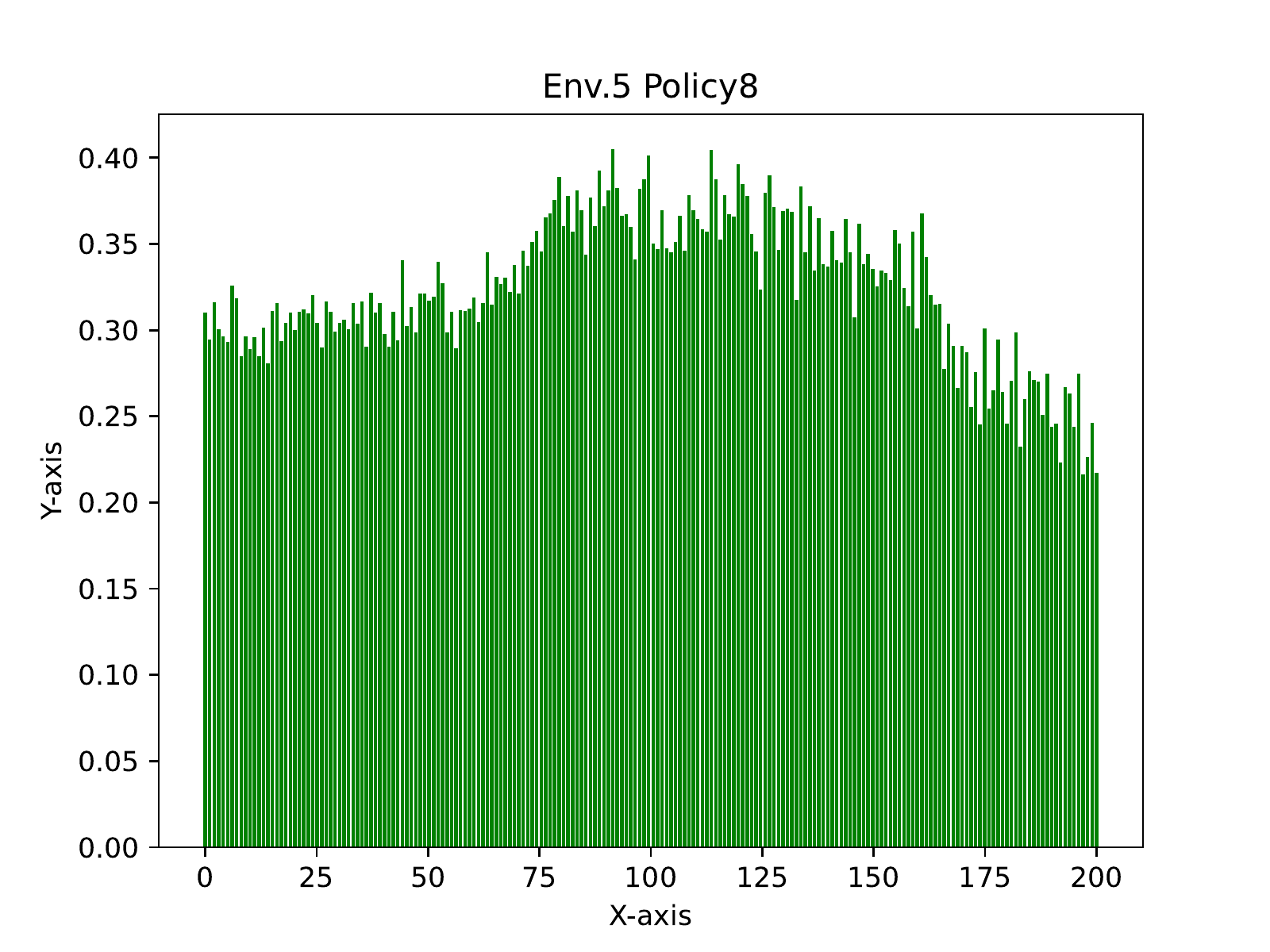}
}
\quad
    
\subfigure[Policy 9 Env. 1 to Env. 5]{
\includegraphics[width=0.19\textwidth]{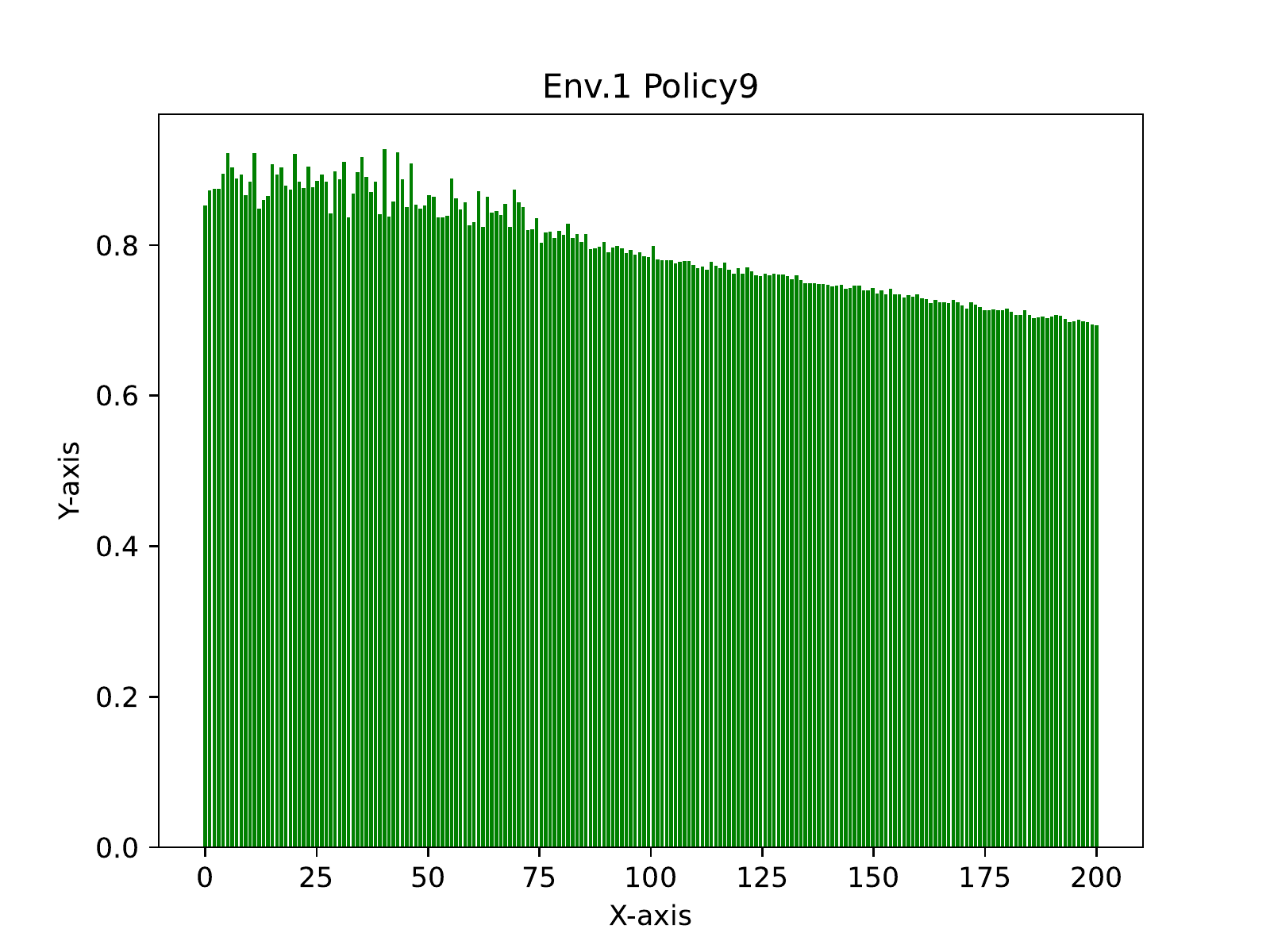}
\includegraphics[width=0.19\textwidth]{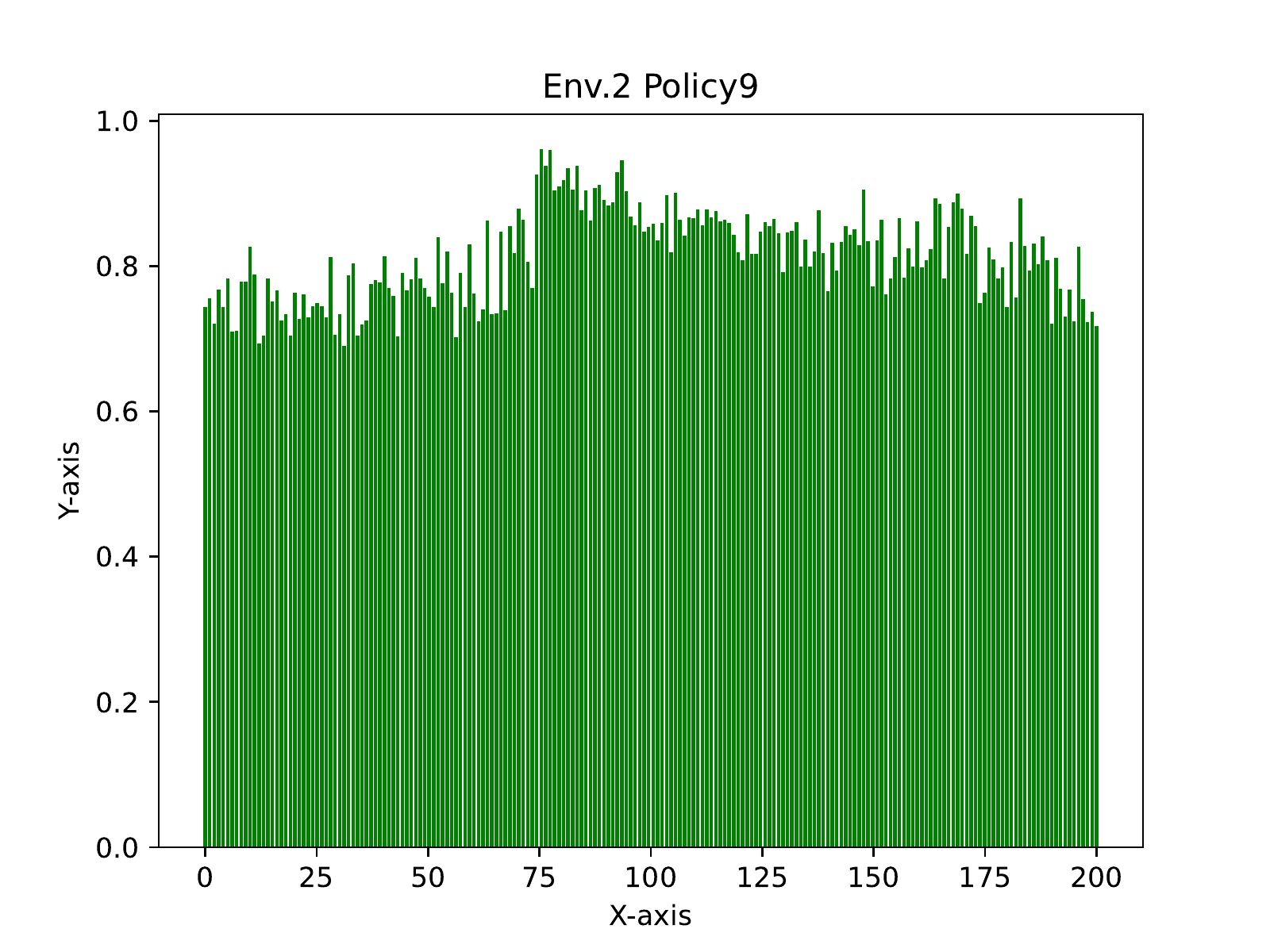}
\includegraphics[width=0.19\textwidth]{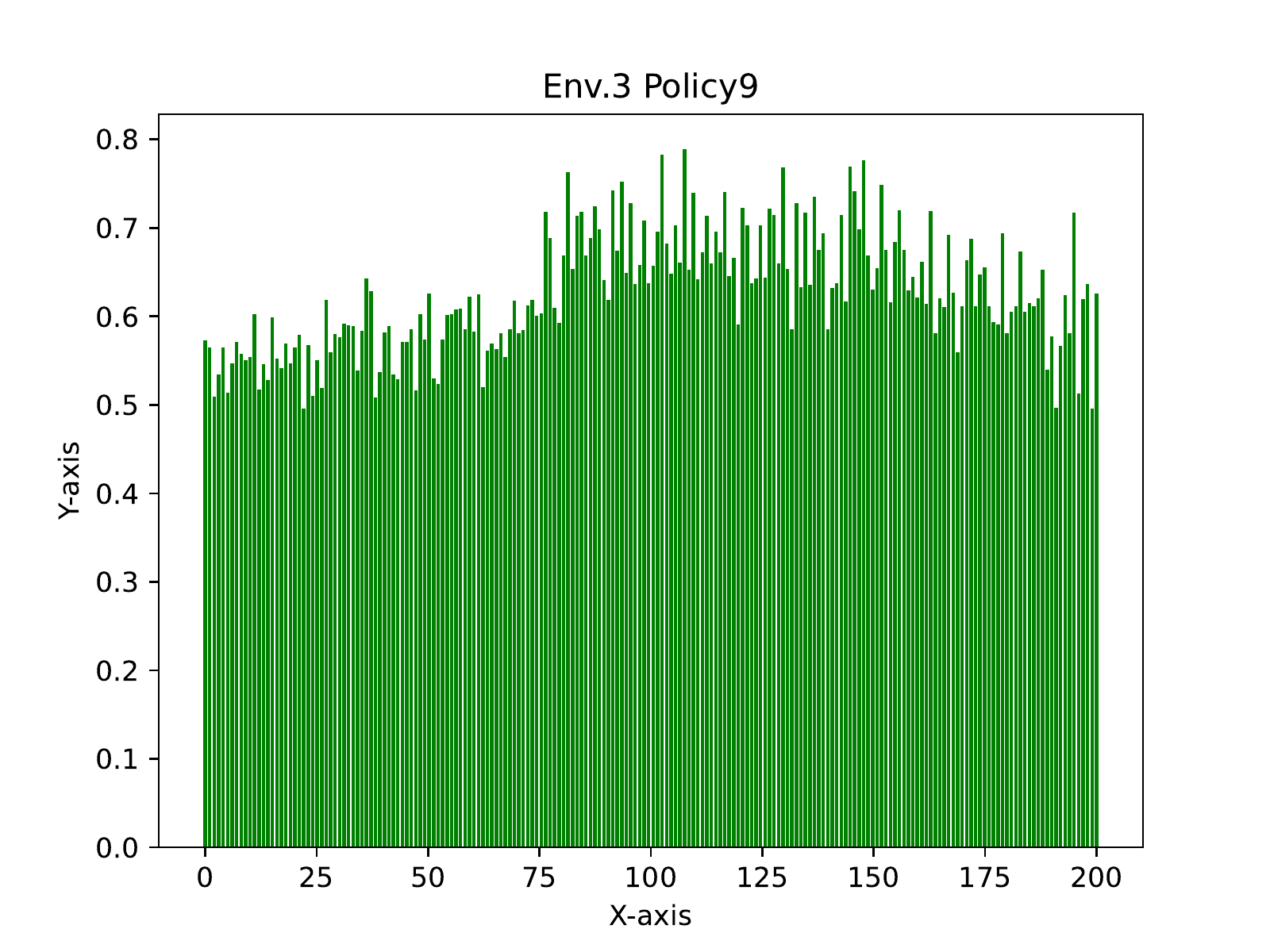}
\includegraphics[width=0.19\textwidth]{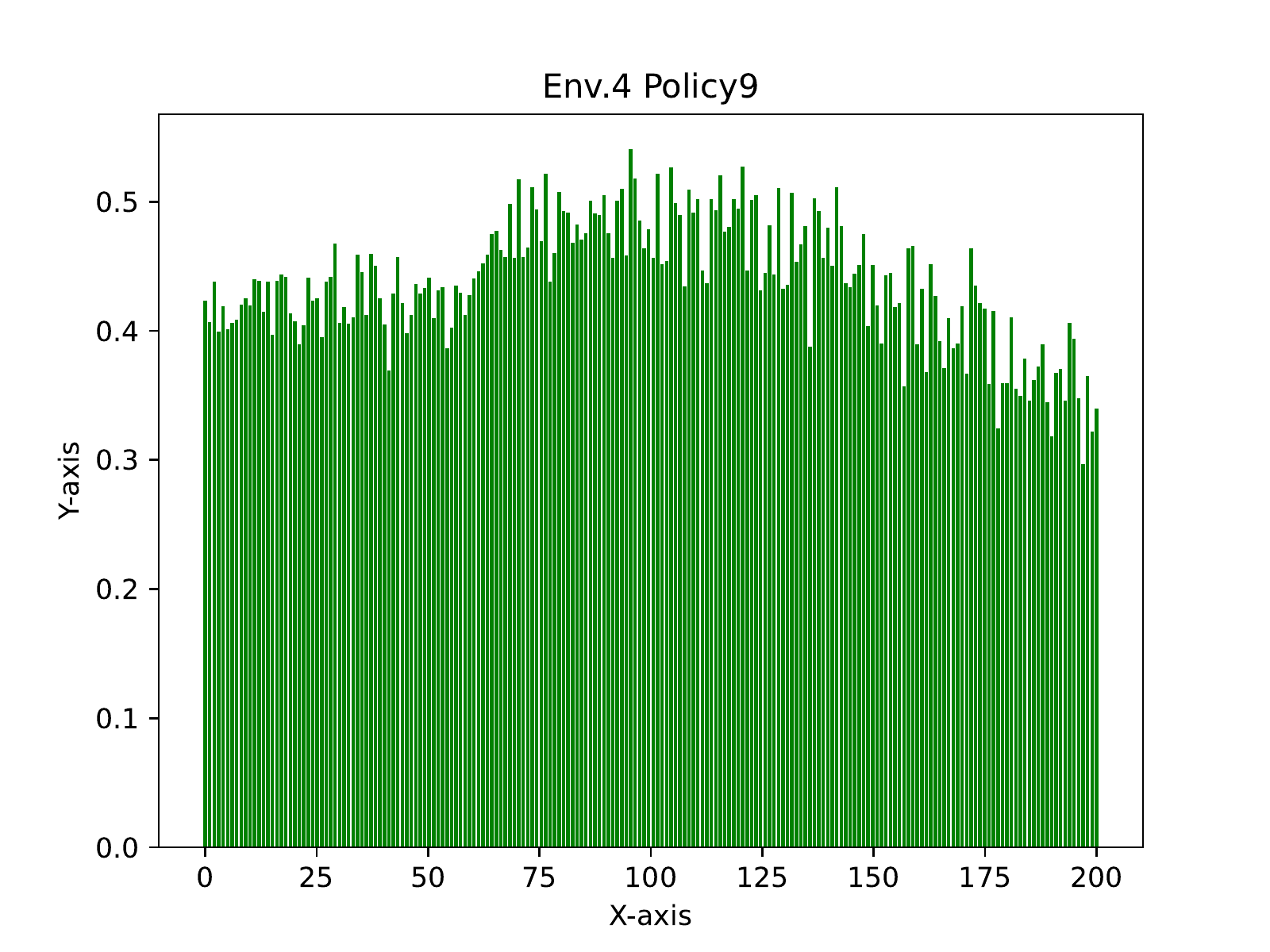}
\includegraphics[width=0.19\textwidth]{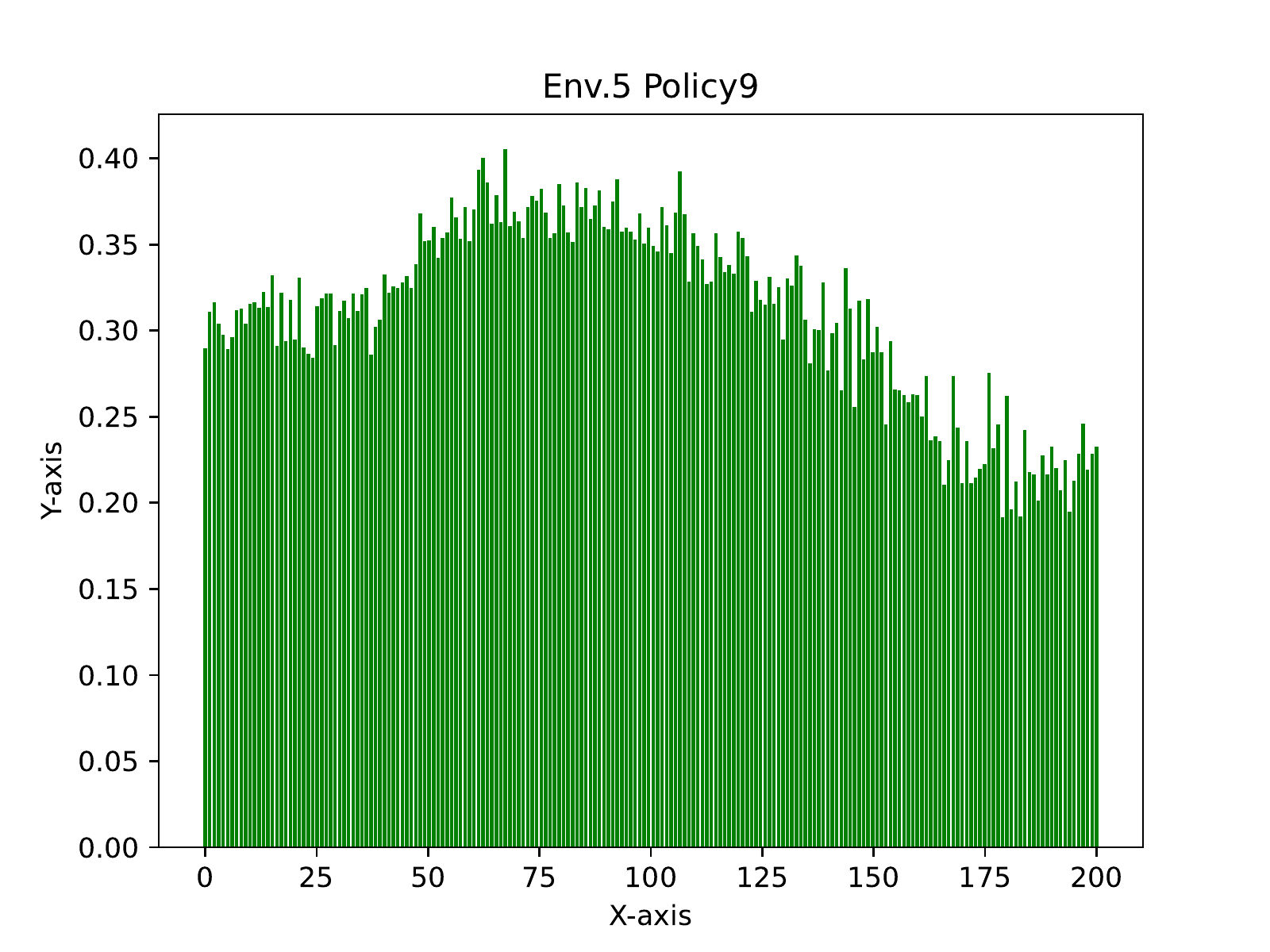}
}
\quad

\subfigure[Policy 10 Env. 1 to Env. 5]{
\includegraphics[width=0.19\textwidth]{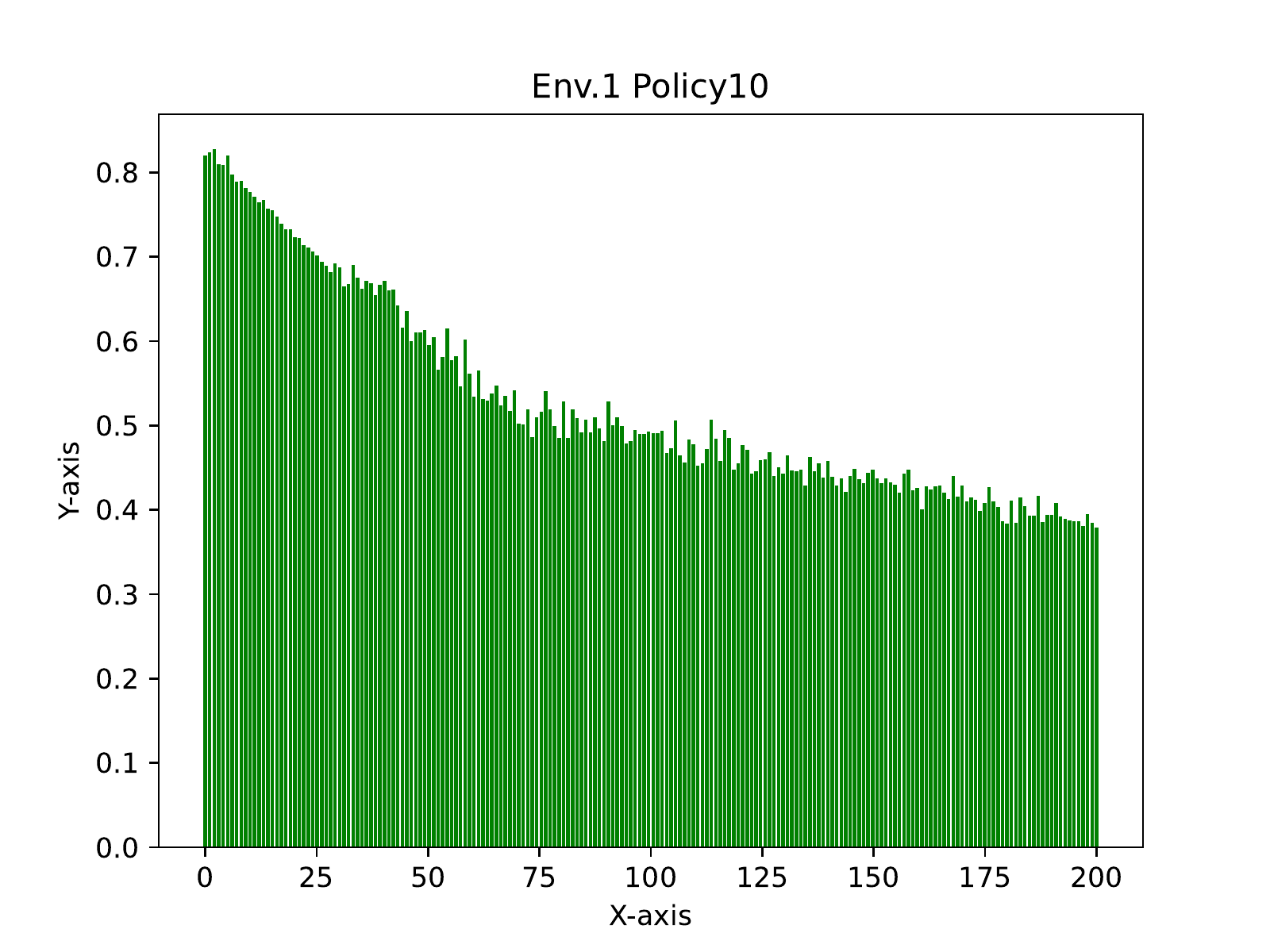}
\includegraphics[width=0.19\textwidth]{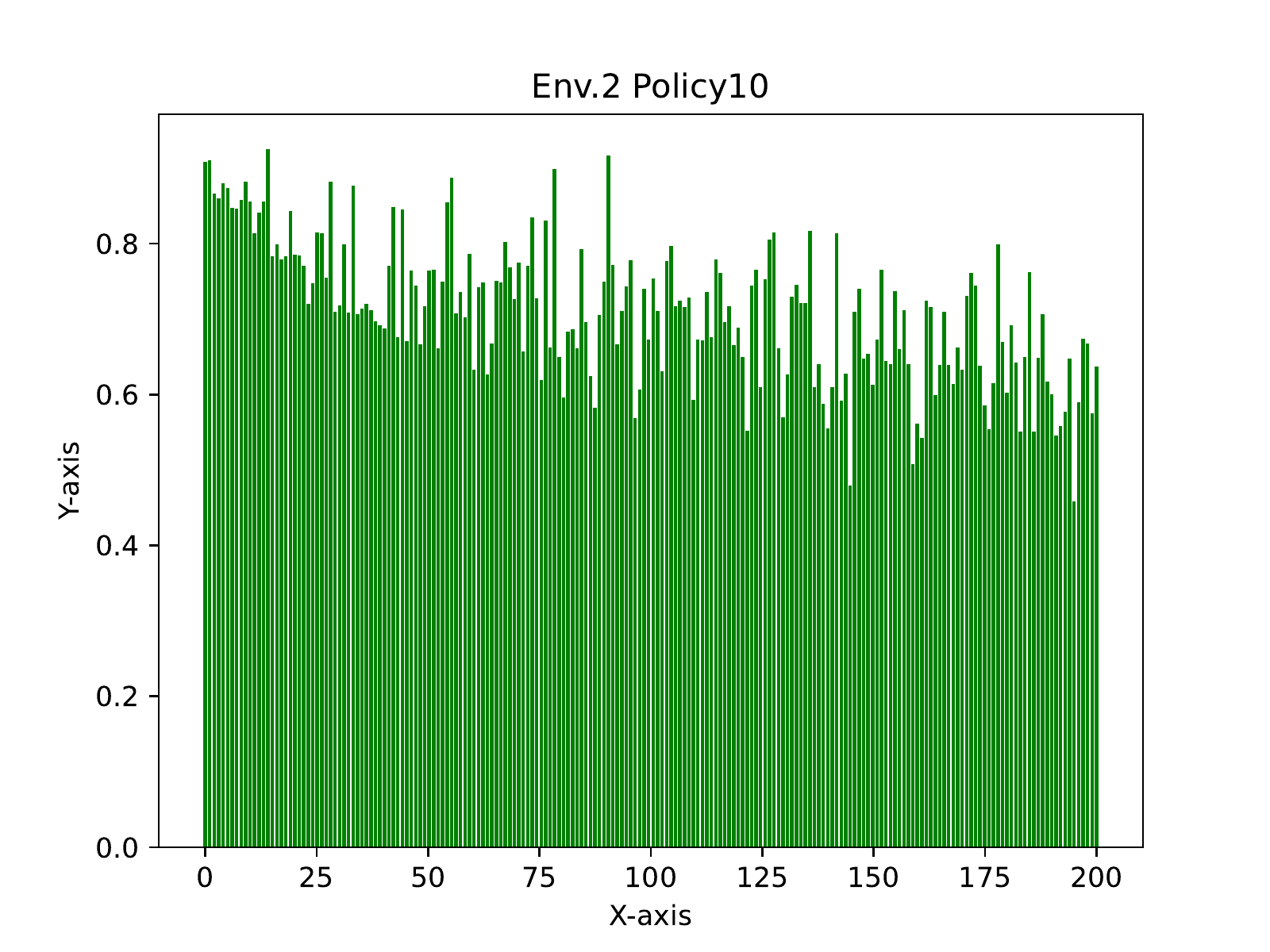}
\includegraphics[width=0.19\textwidth]{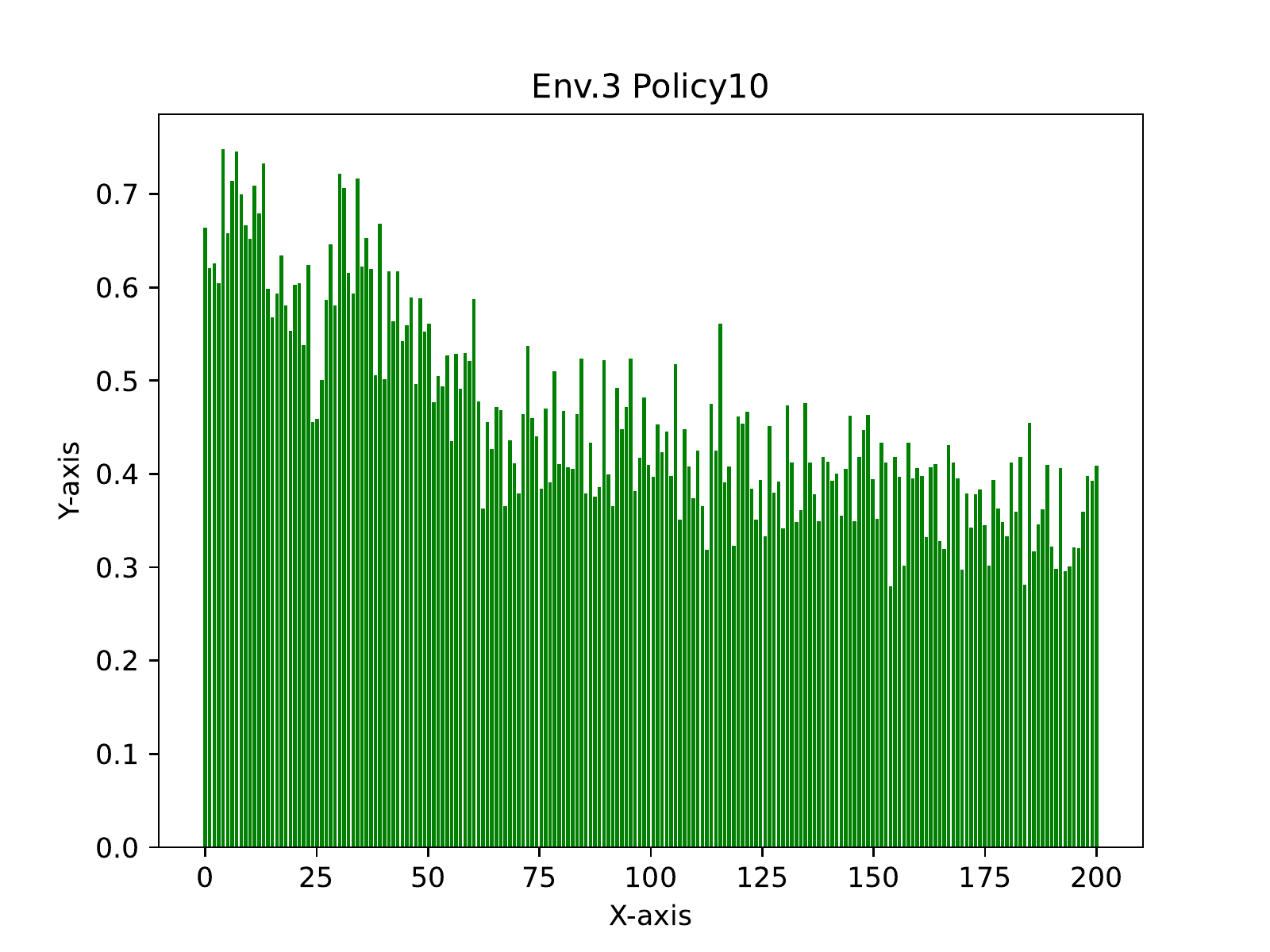}
\includegraphics[width=0.19\textwidth]{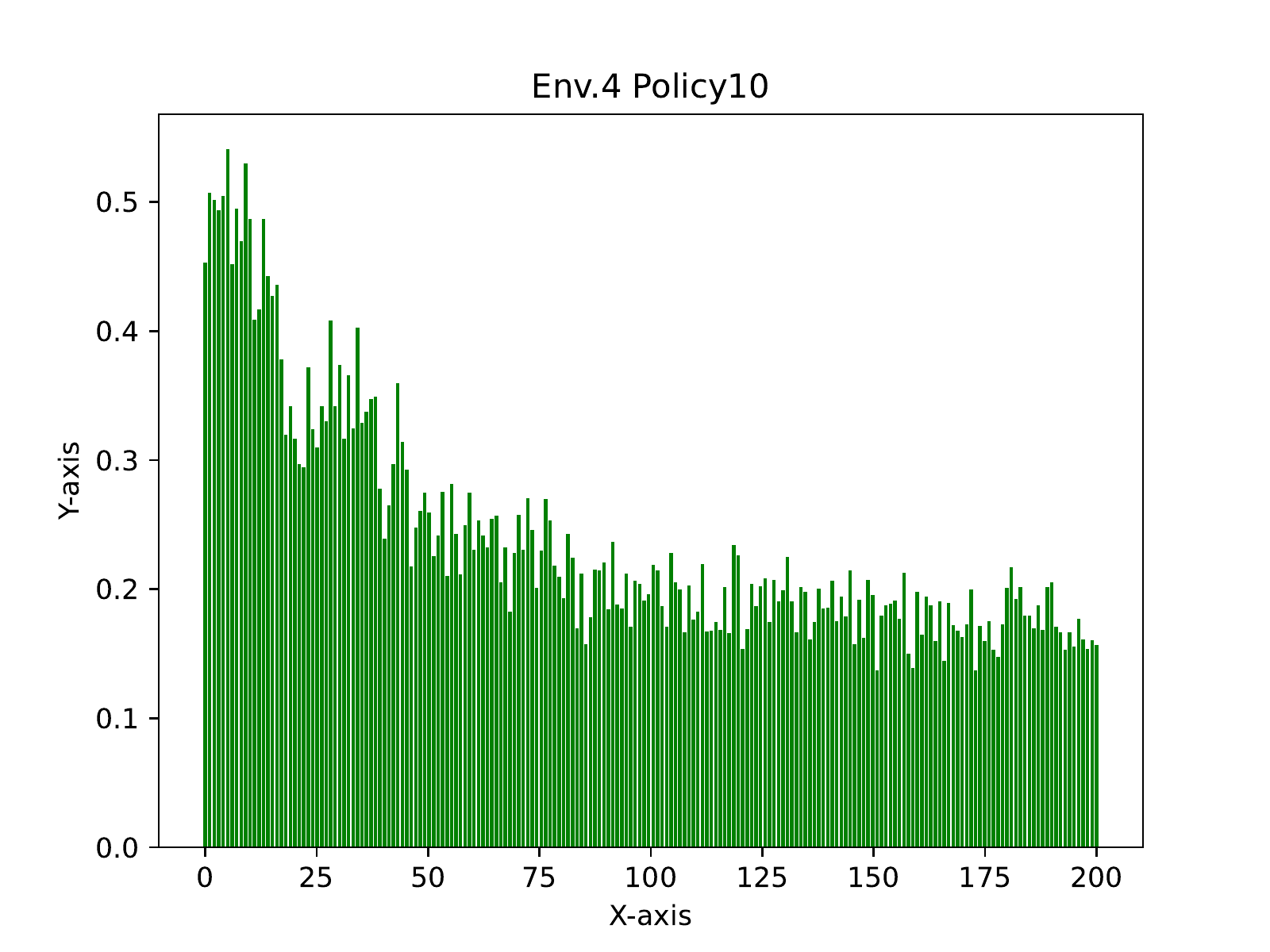}
\includegraphics[width=0.19\textwidth]{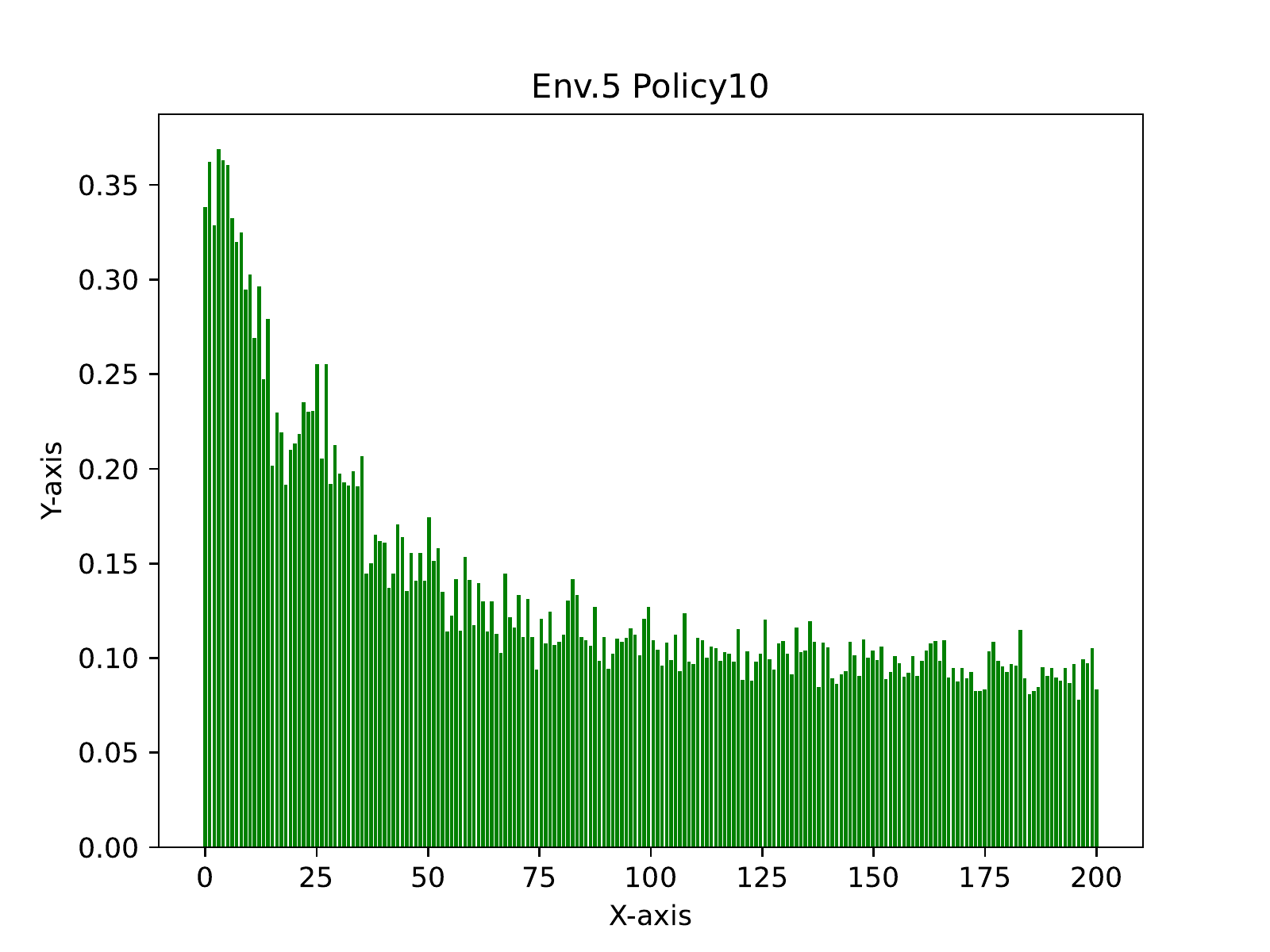}
}
\quad

\subfigure[Policy 11 Env. 1 to Env. 5]{
\includegraphics[width=0.19\textwidth]{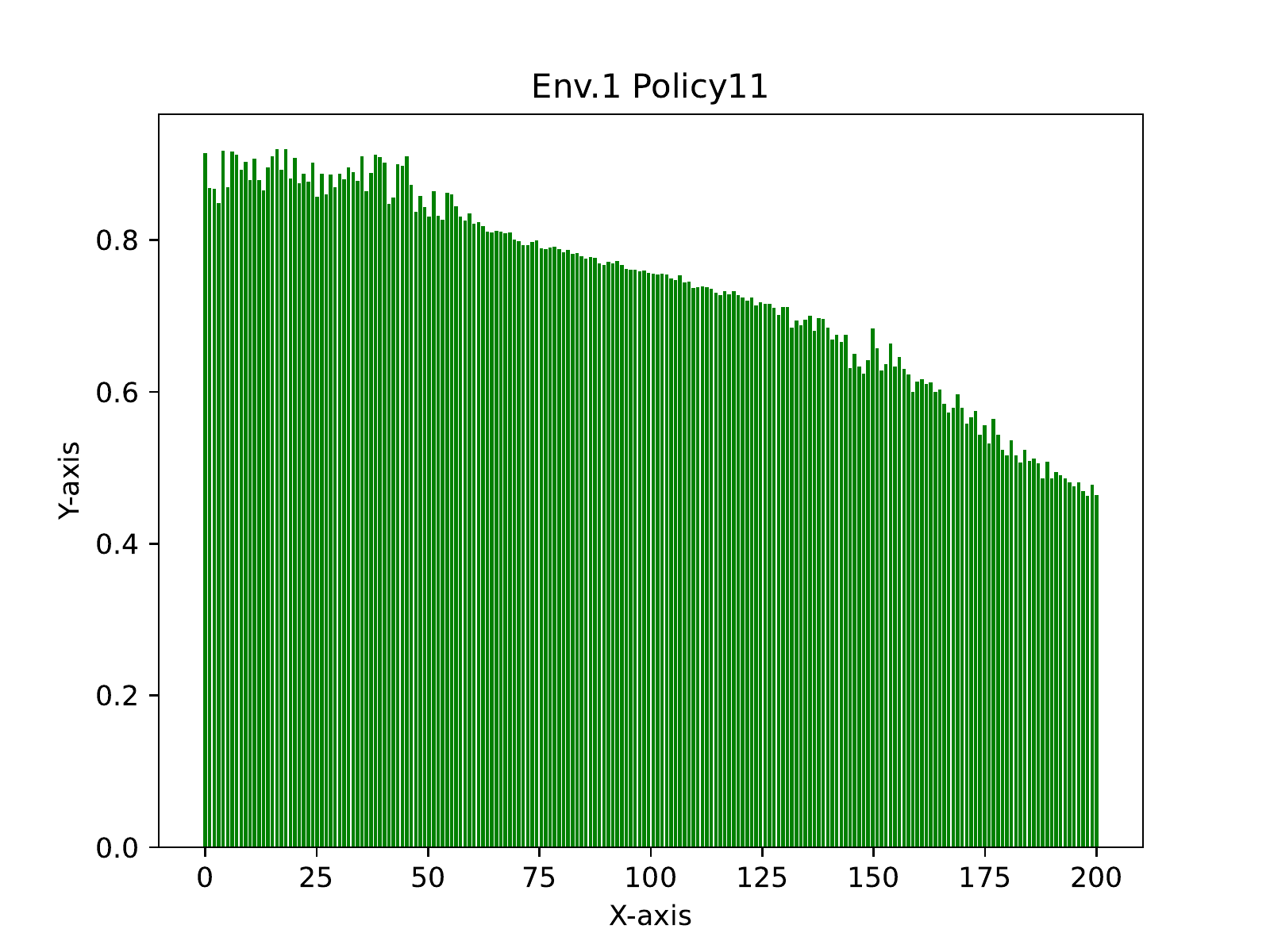}
\includegraphics[width=0.19\textwidth]{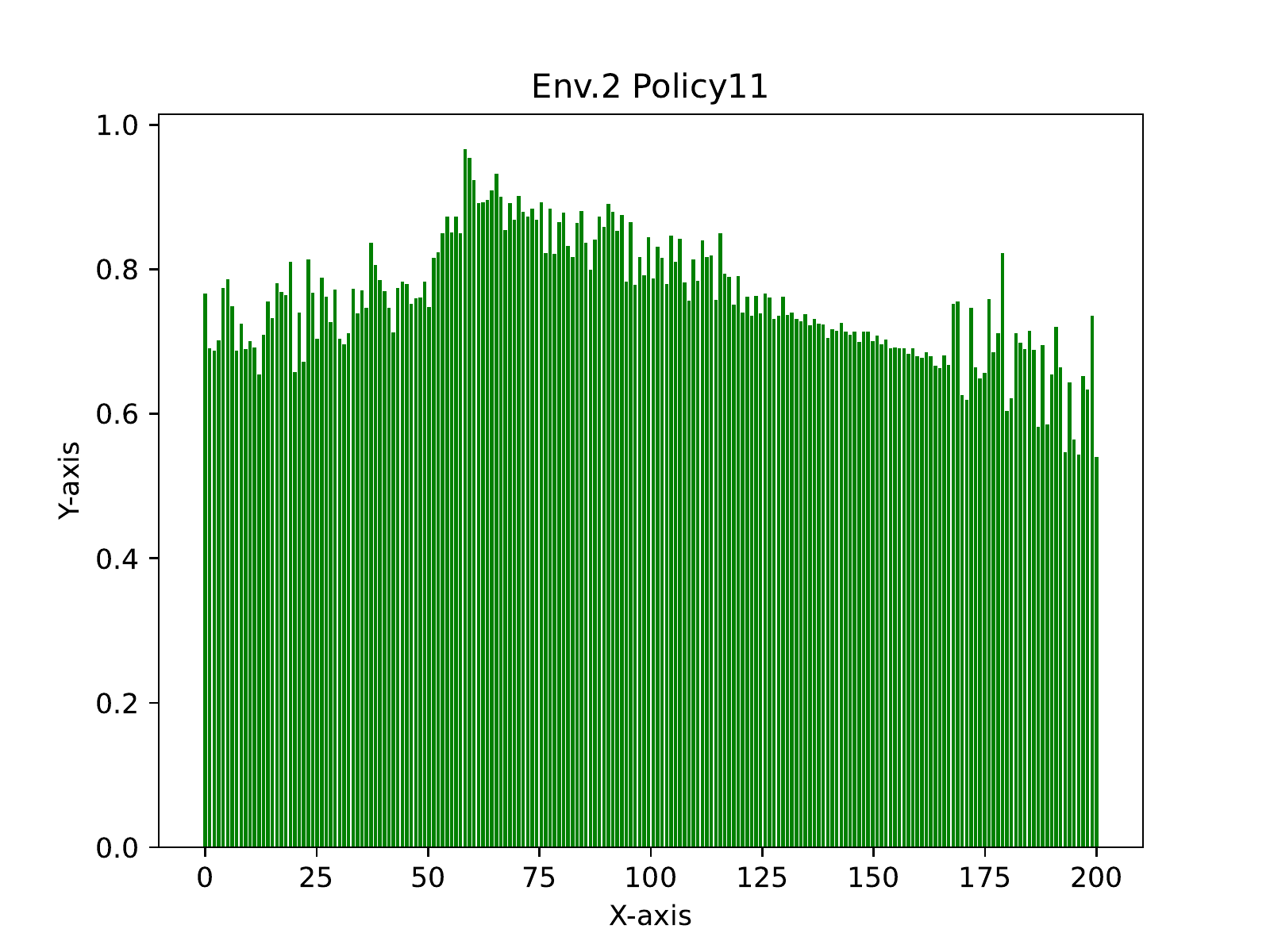}
\includegraphics[width=0.19\textwidth]{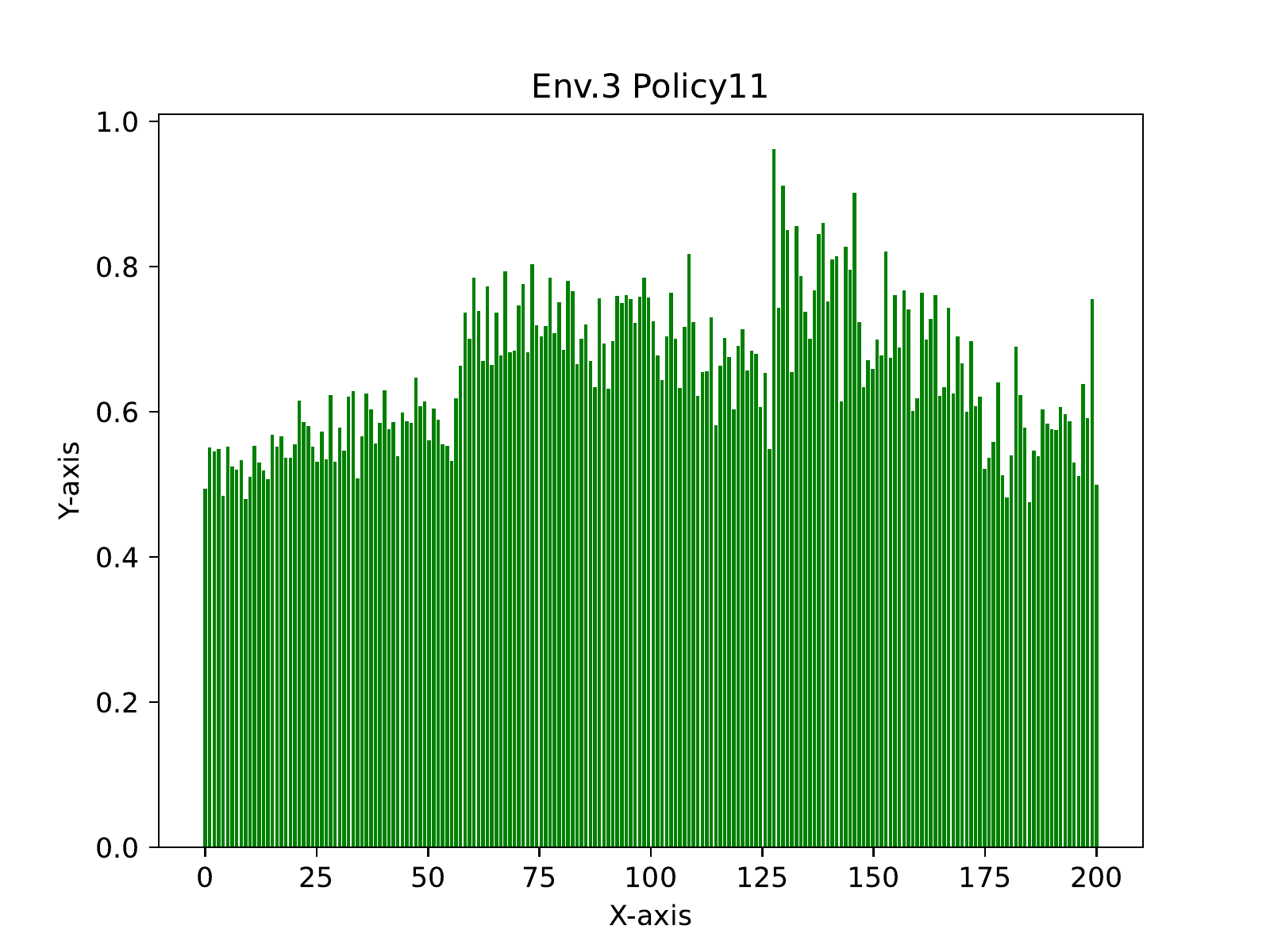}
\includegraphics[width=0.19\textwidth]{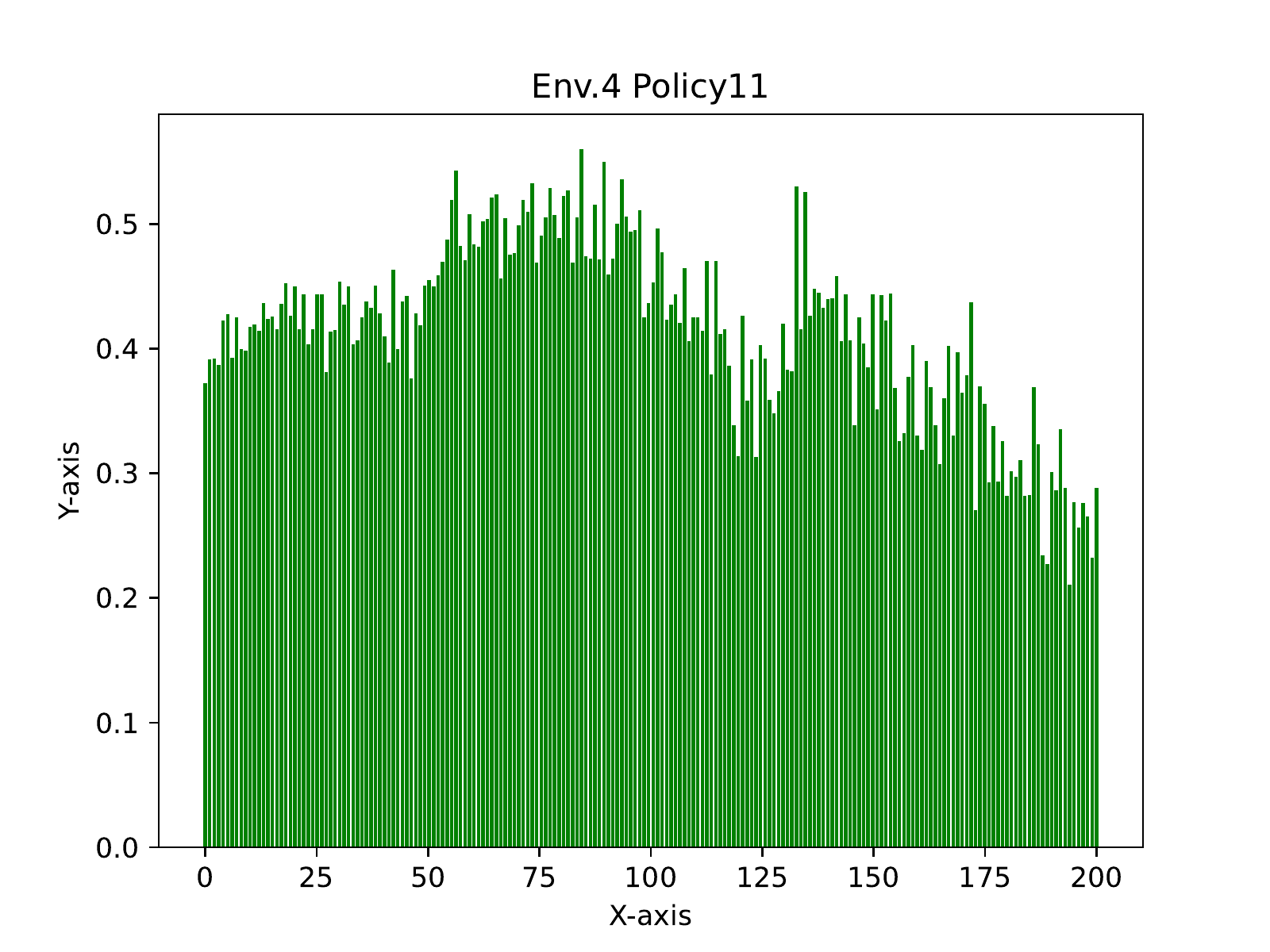}
\includegraphics[width=0.19\textwidth]{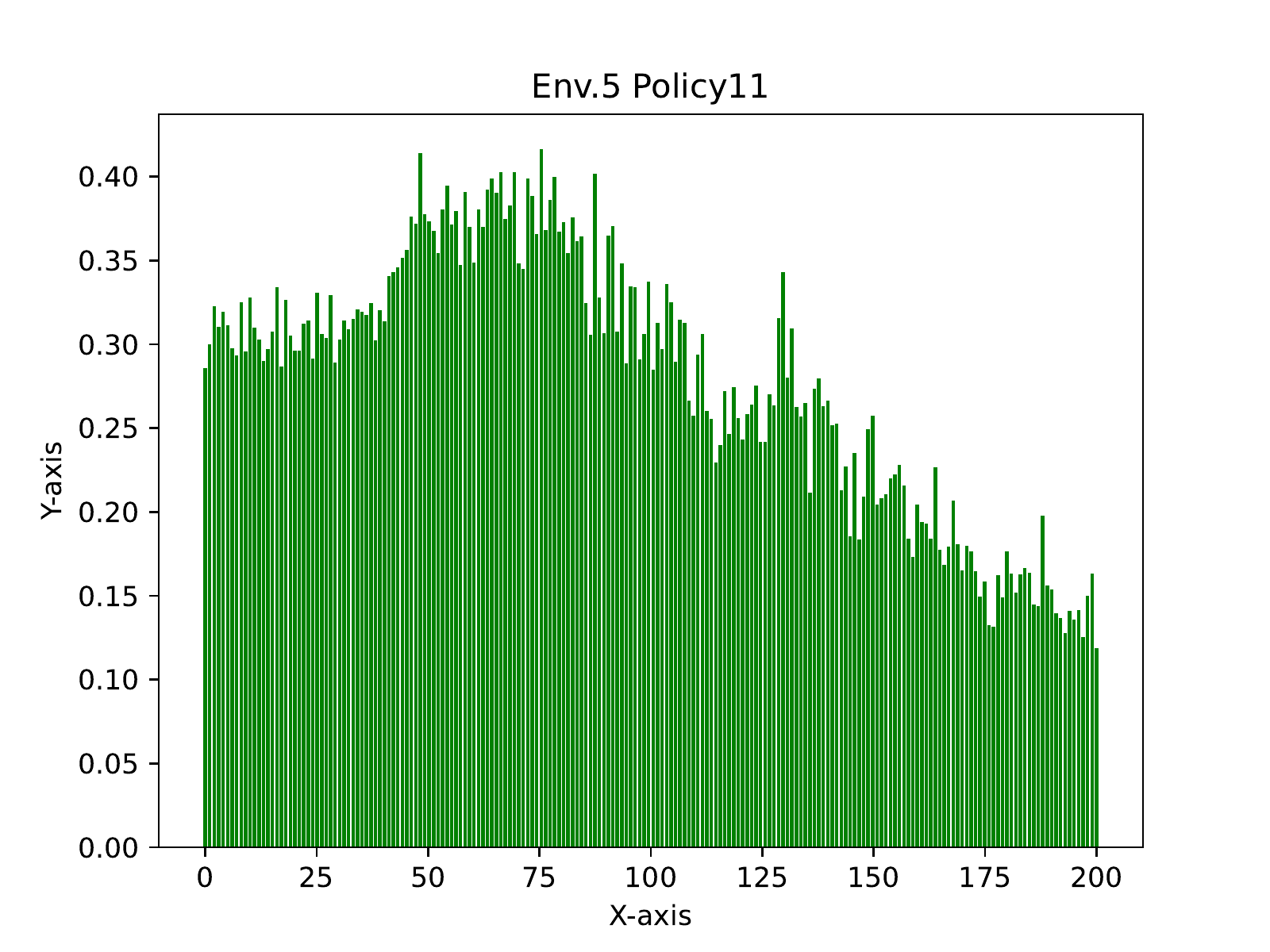}
}
\quad

\subfigure[Policy 12 Env. 1 to Env. 5]{
\includegraphics[width=0.19\textwidth]{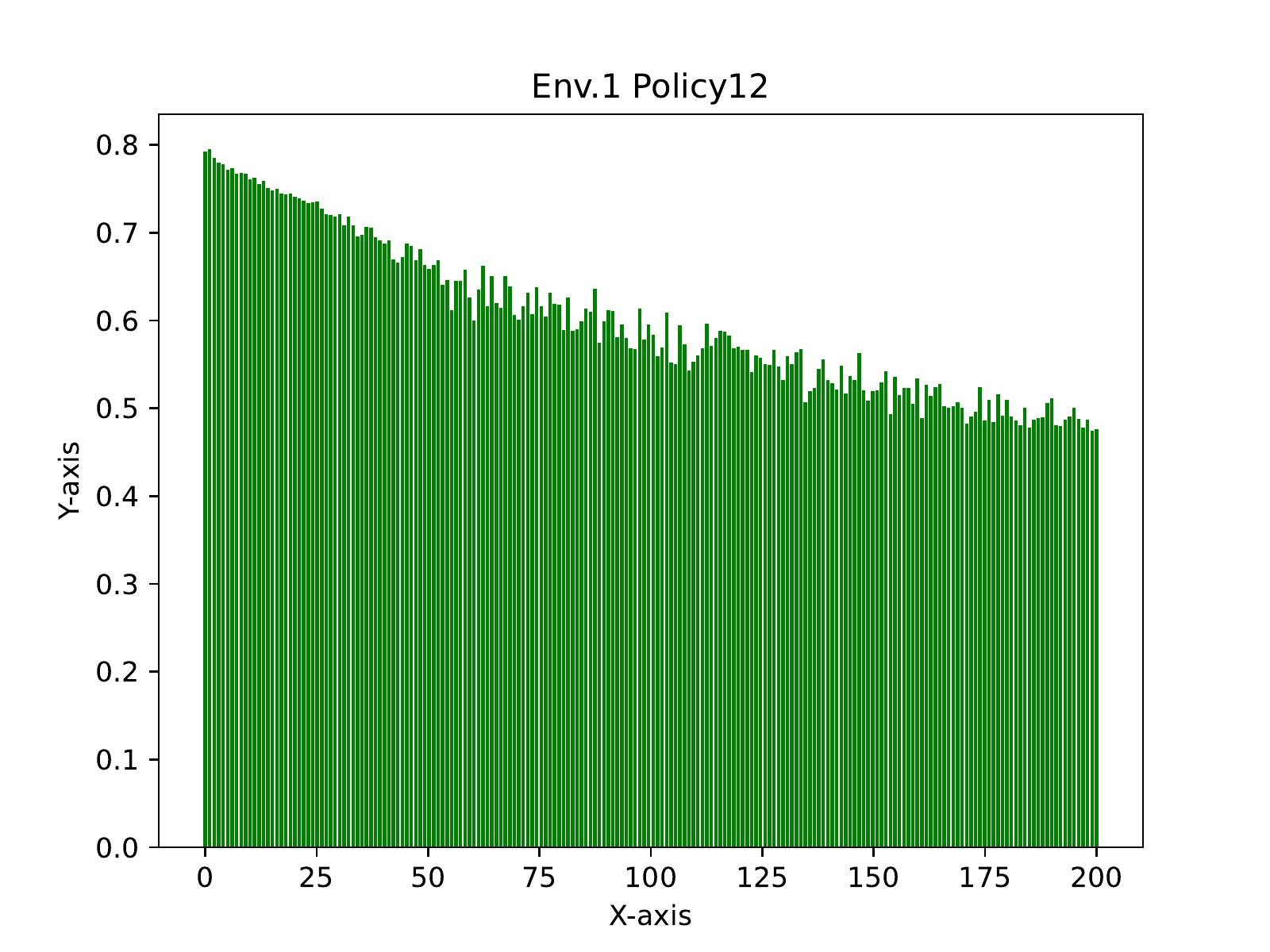}
\includegraphics[width=0.19\textwidth]{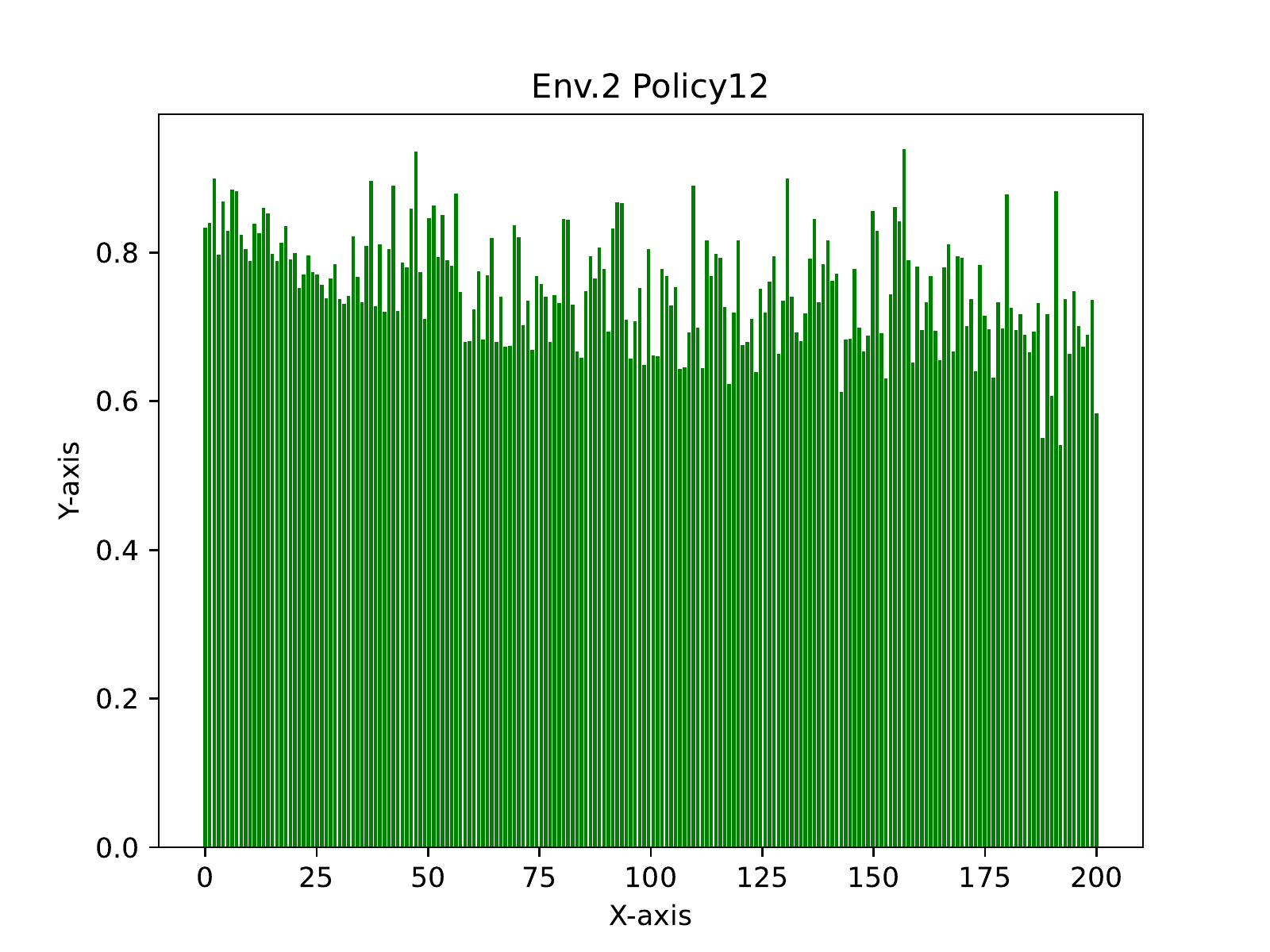}
\includegraphics[width=0.19\textwidth]{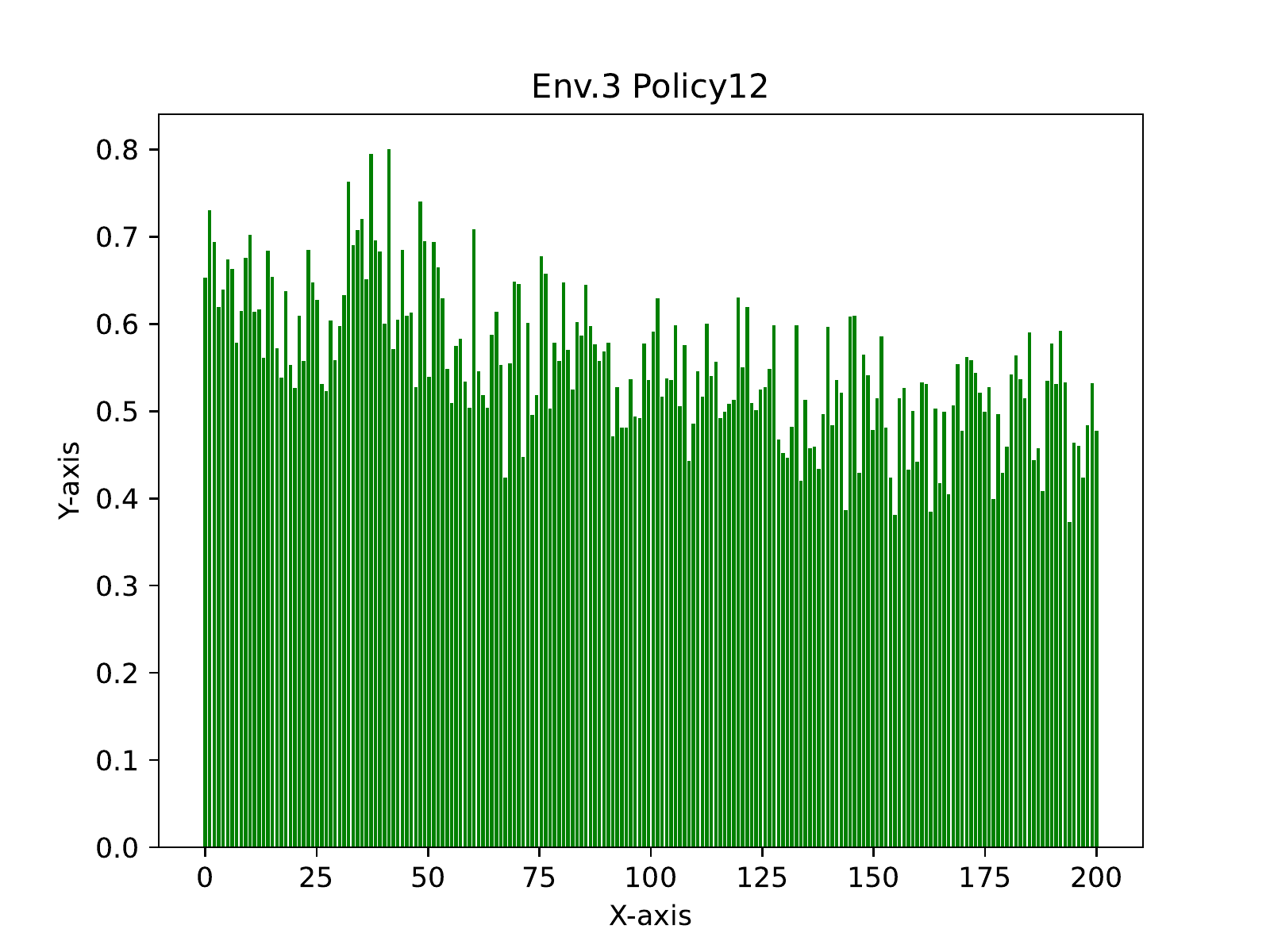}
\includegraphics[width=0.19\textwidth]{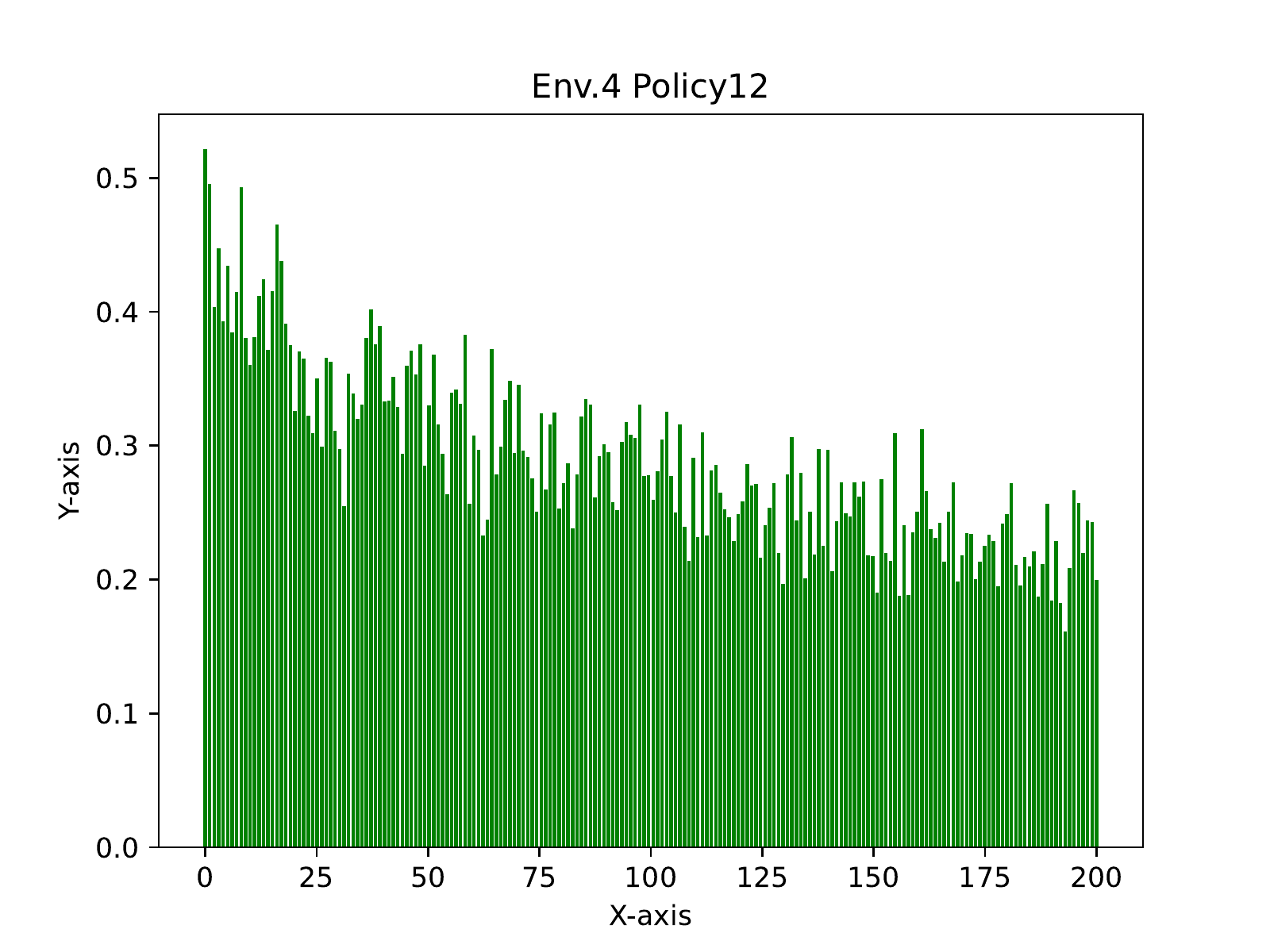}
\includegraphics[width=0.19\textwidth]{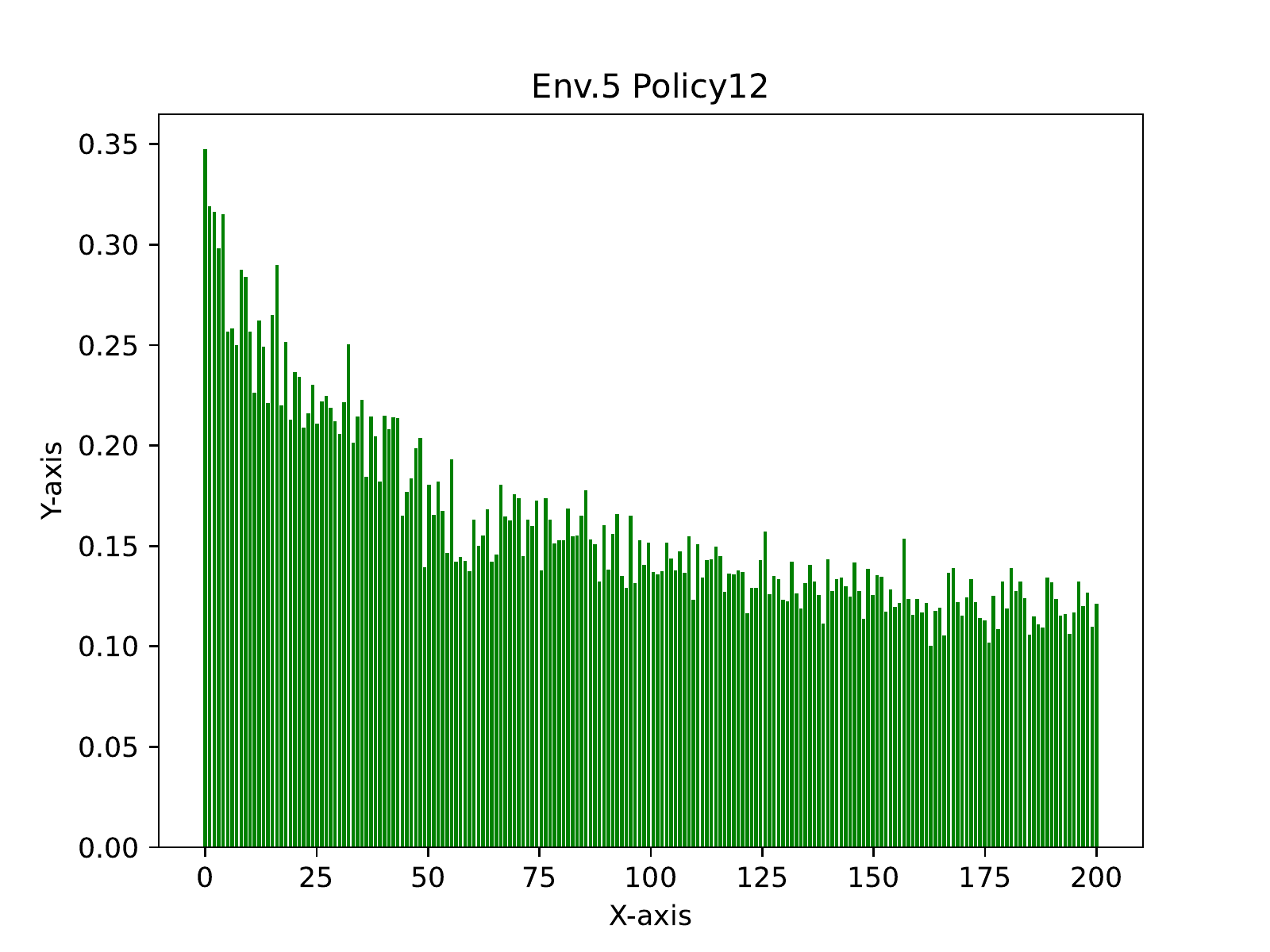}
}
\quad

\caption{Episodic MC return distribution of adapted policy during GA process.}
\label{fig::Para Tune Space 2}
\end{figure*}

\begin{figure*}[t]
\addtocounter{subfigure}{0}
\ContinuedFloat
\centering

\subfigure[Policy 13 Env. 1 to Env. 5]{
\includegraphics[width=0.19\textwidth]{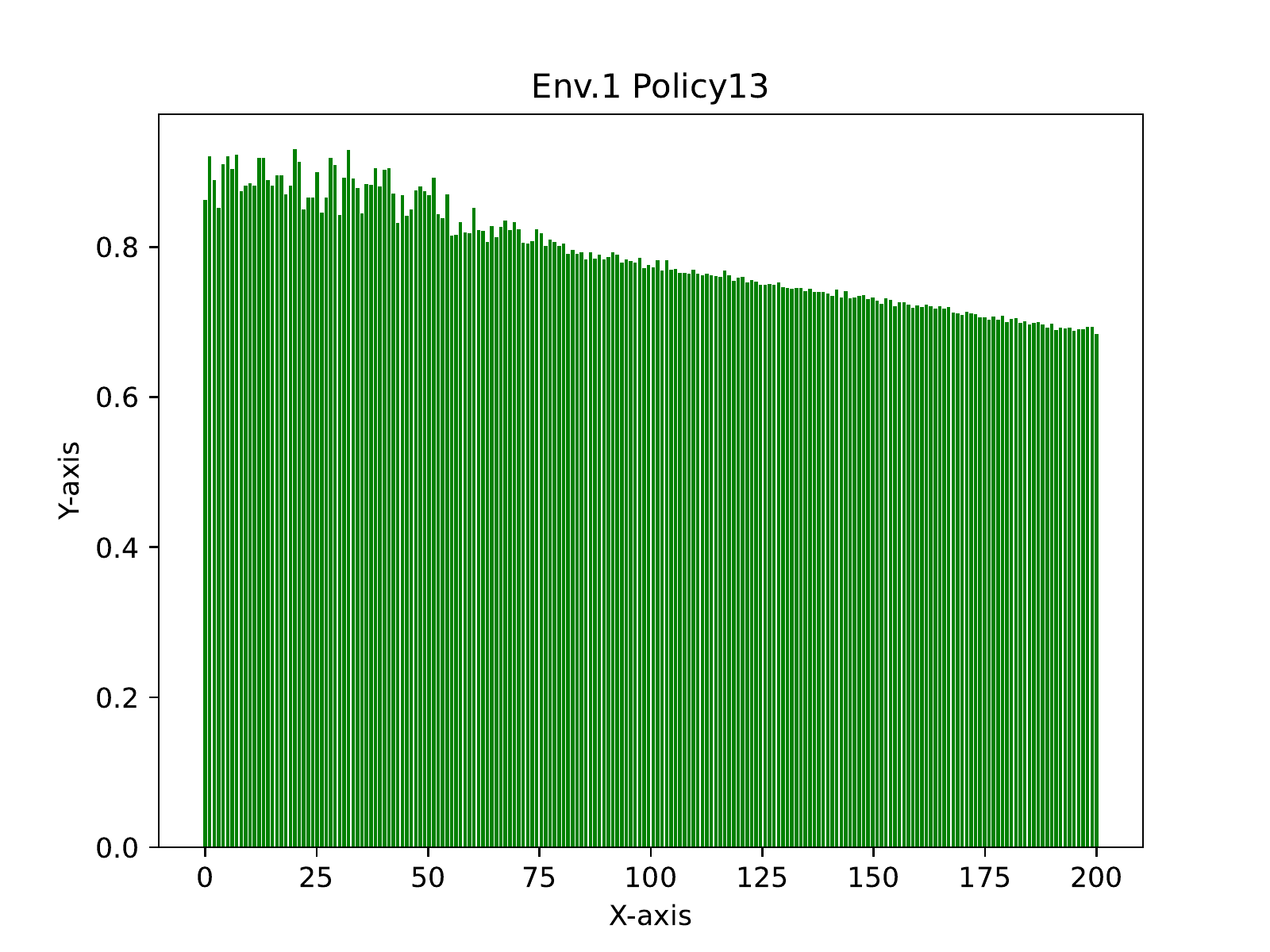}
\includegraphics[width=0.19\textwidth]{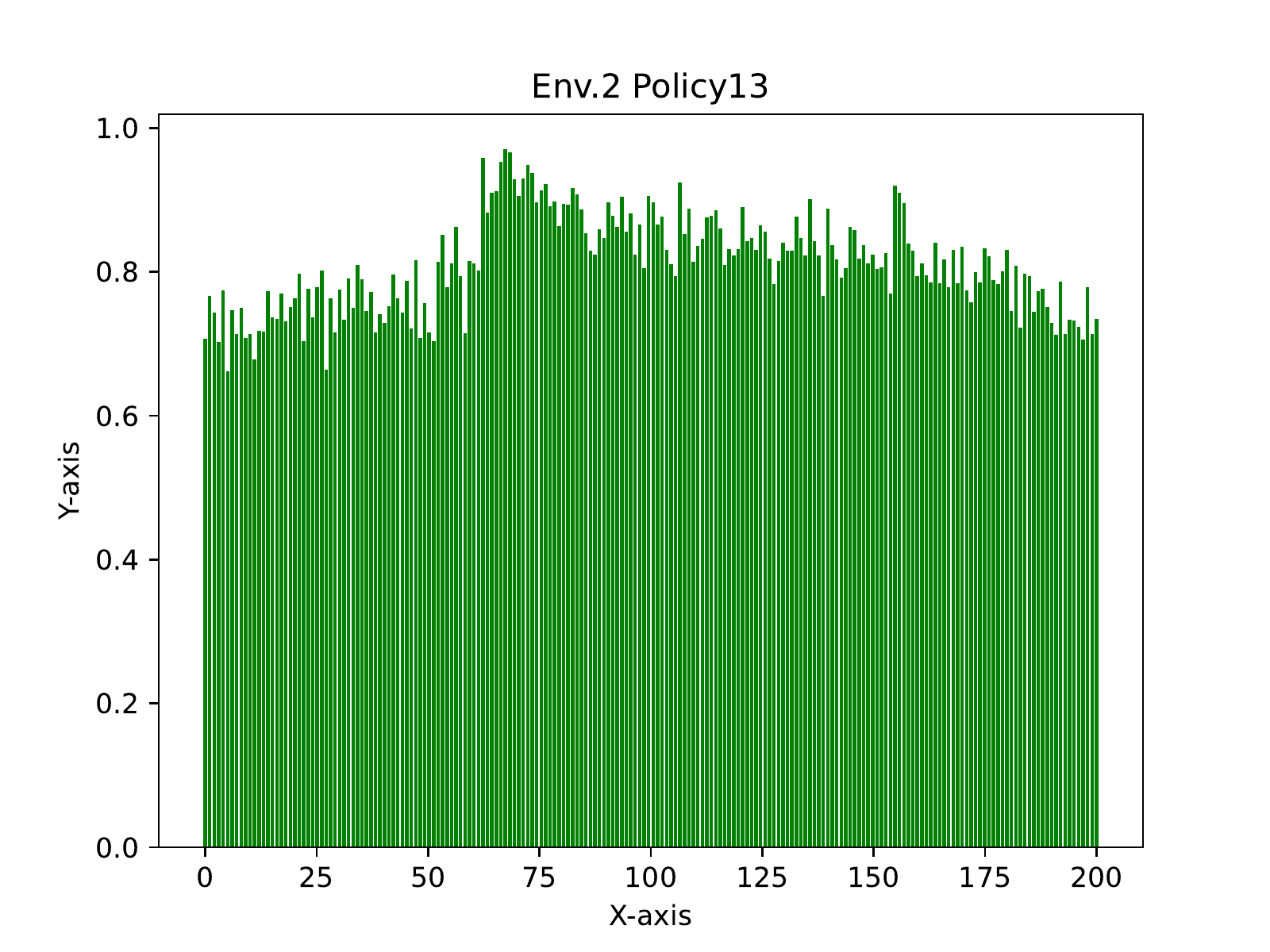}
\includegraphics[width=0.19\textwidth]{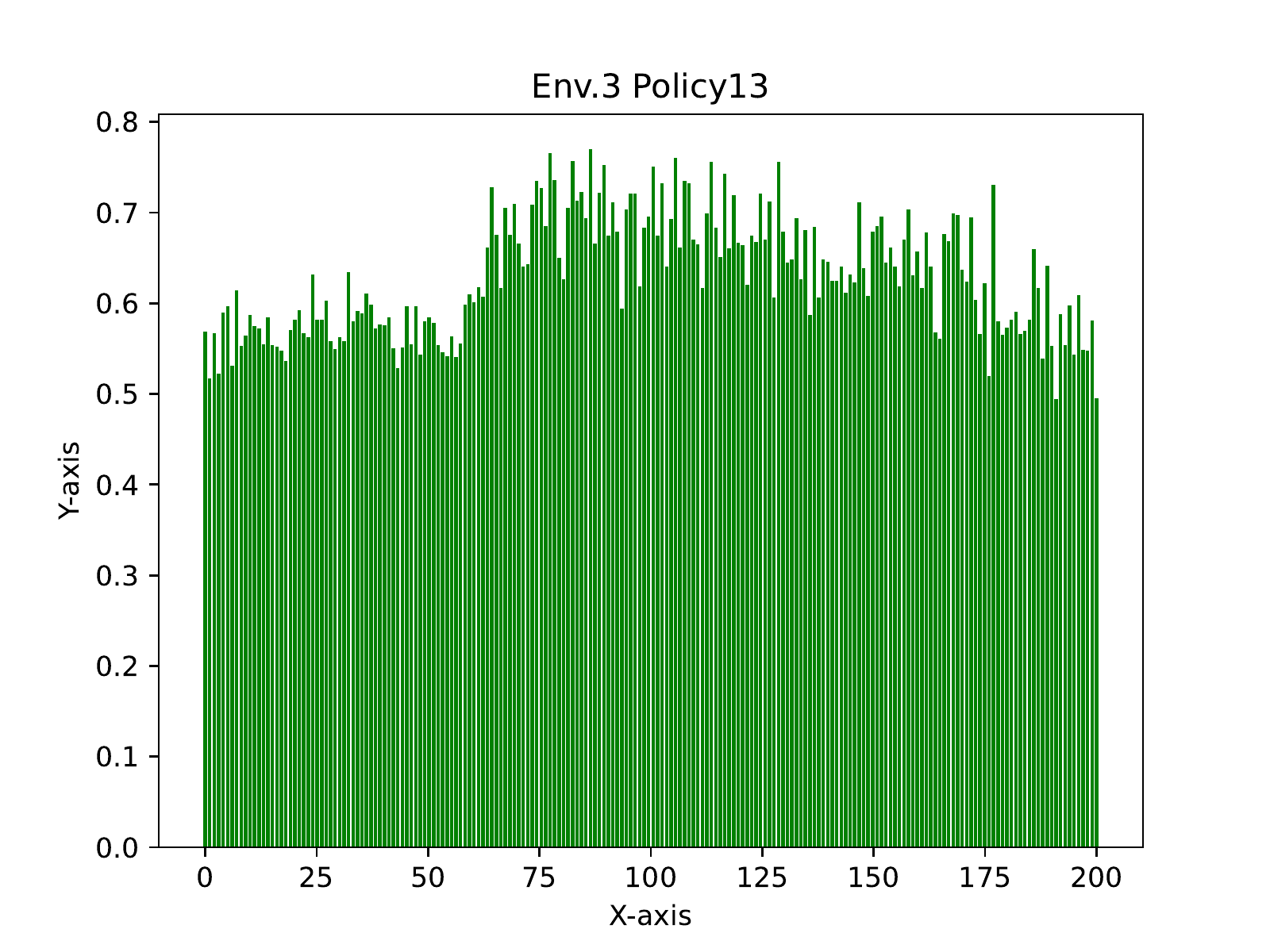}
\includegraphics[width=0.19\textwidth]{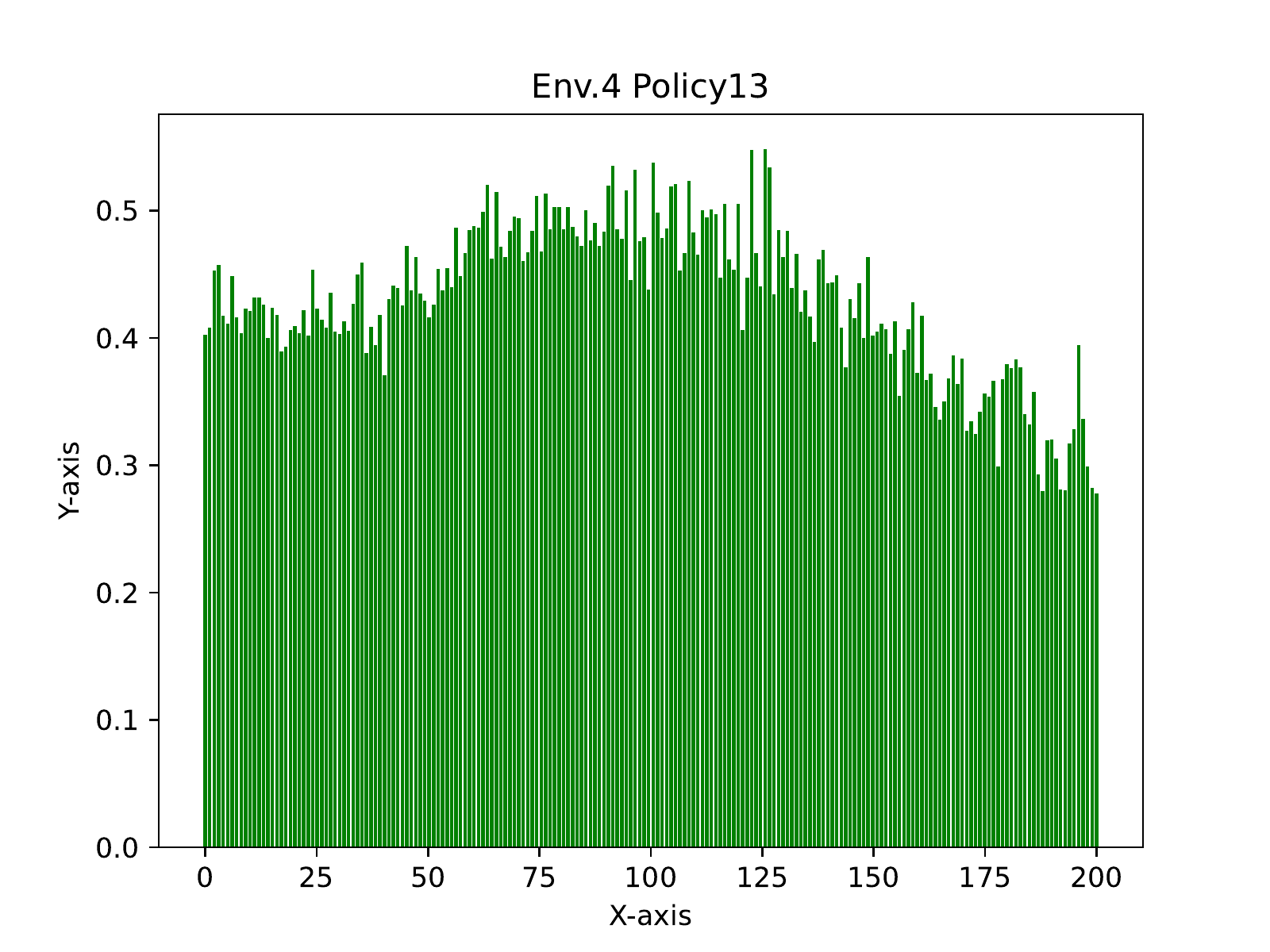}
\includegraphics[width=0.19\textwidth]{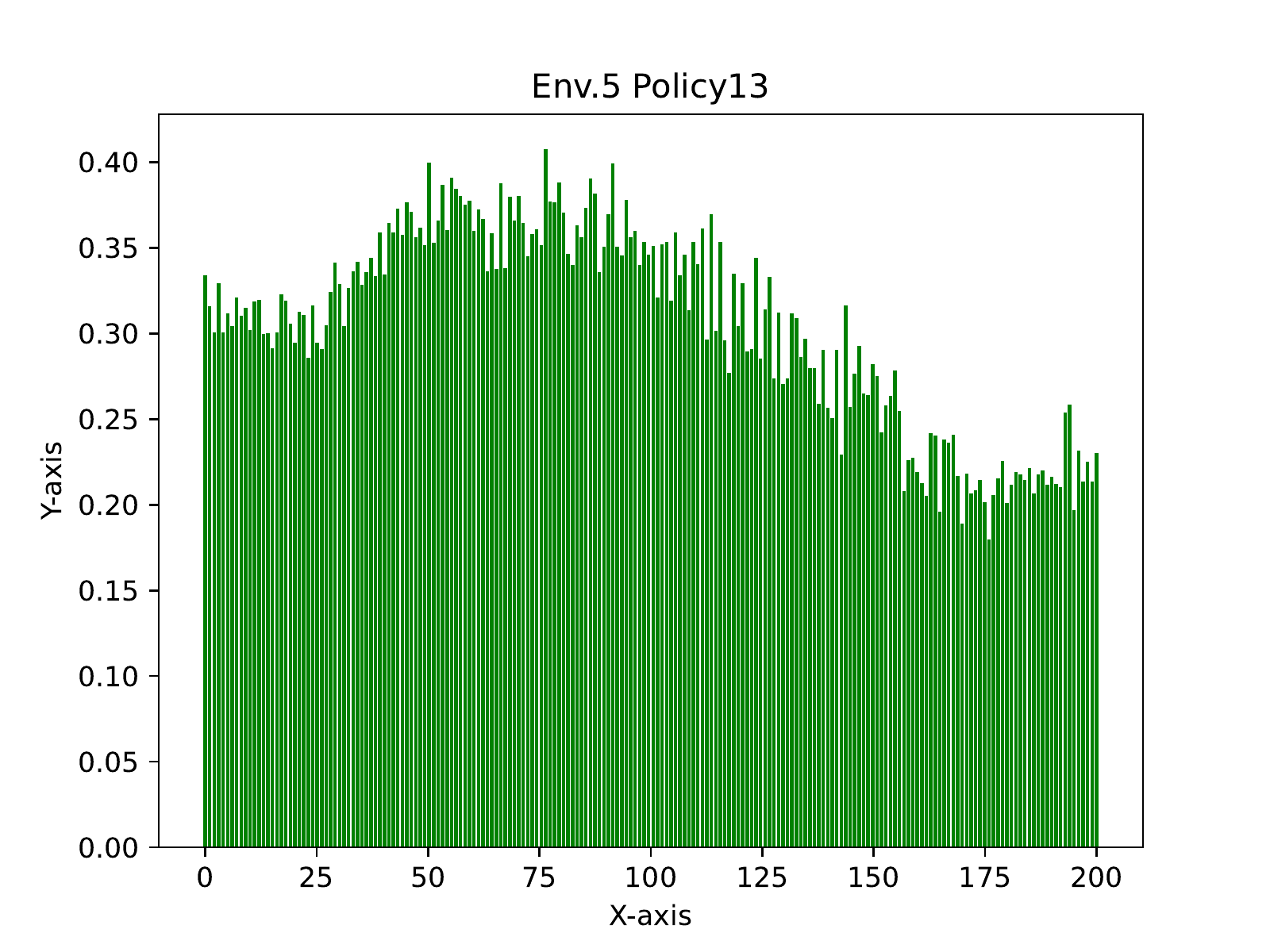}
}
\quad

\subfigure[Policy 14 Env. 1 to Env. 5]{
\includegraphics[width=0.19\textwidth]{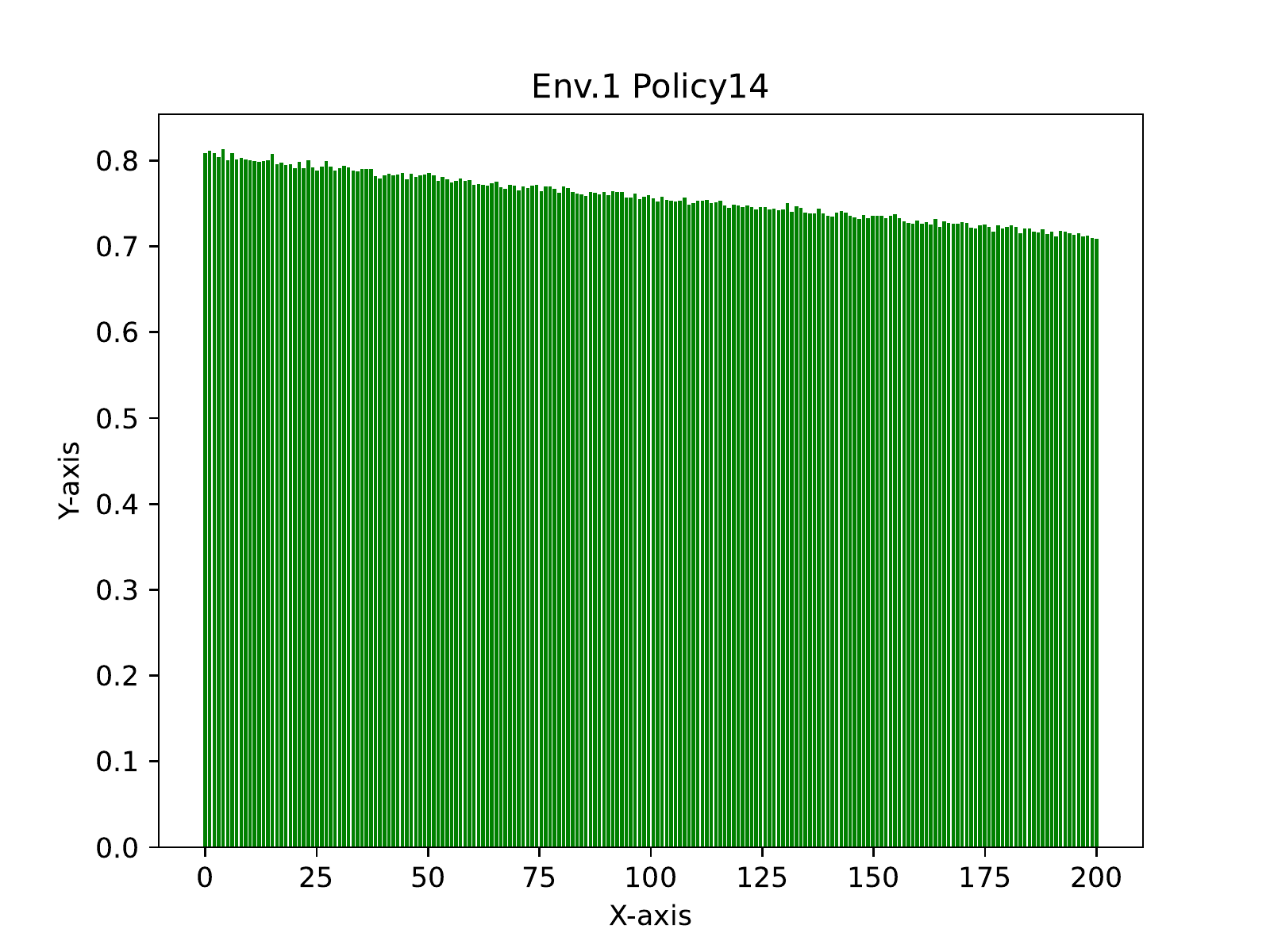}
\includegraphics[width=0.19\textwidth]{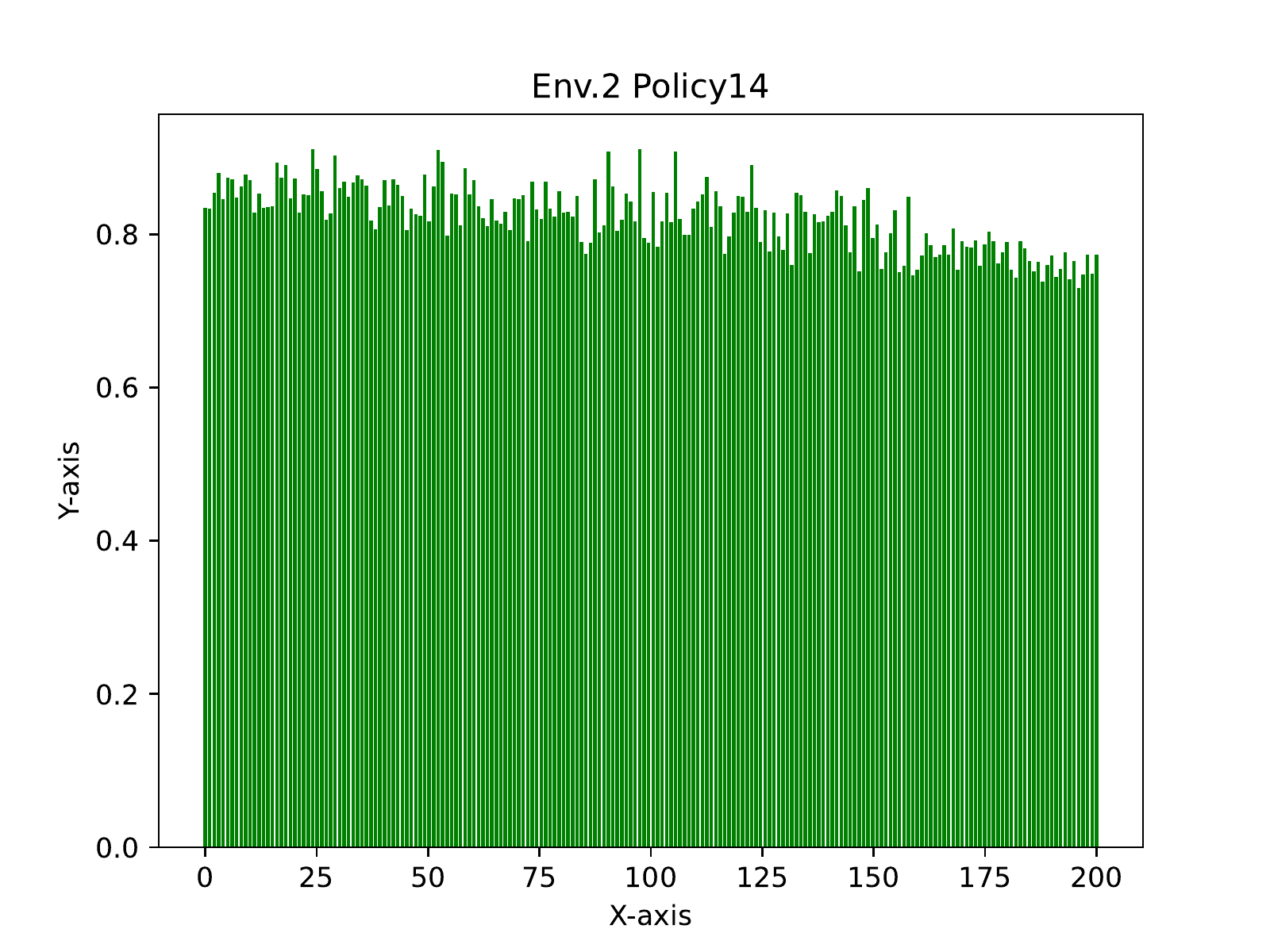}
\includegraphics[width=0.19\textwidth]{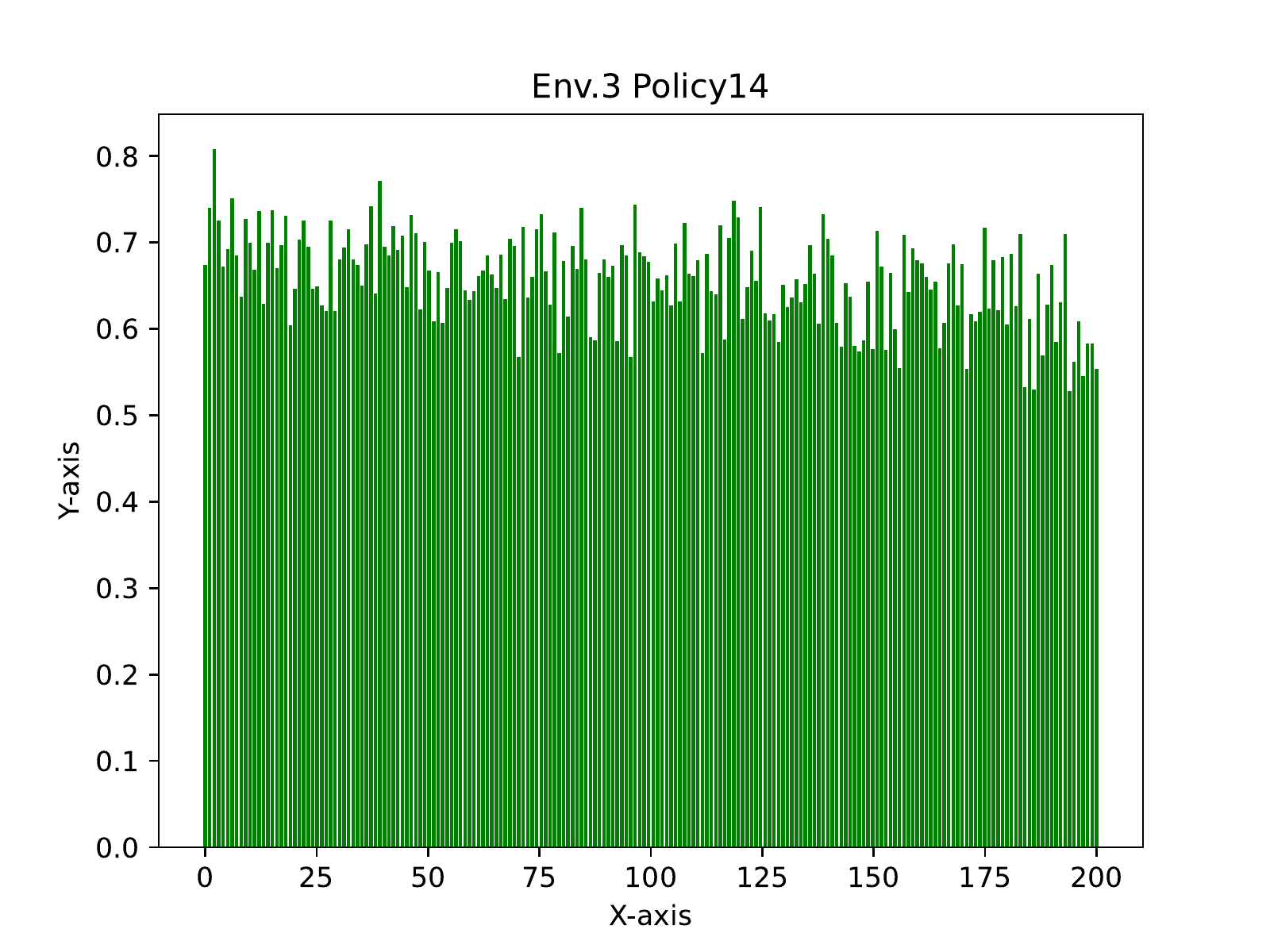}
\includegraphics[width=0.19\textwidth]{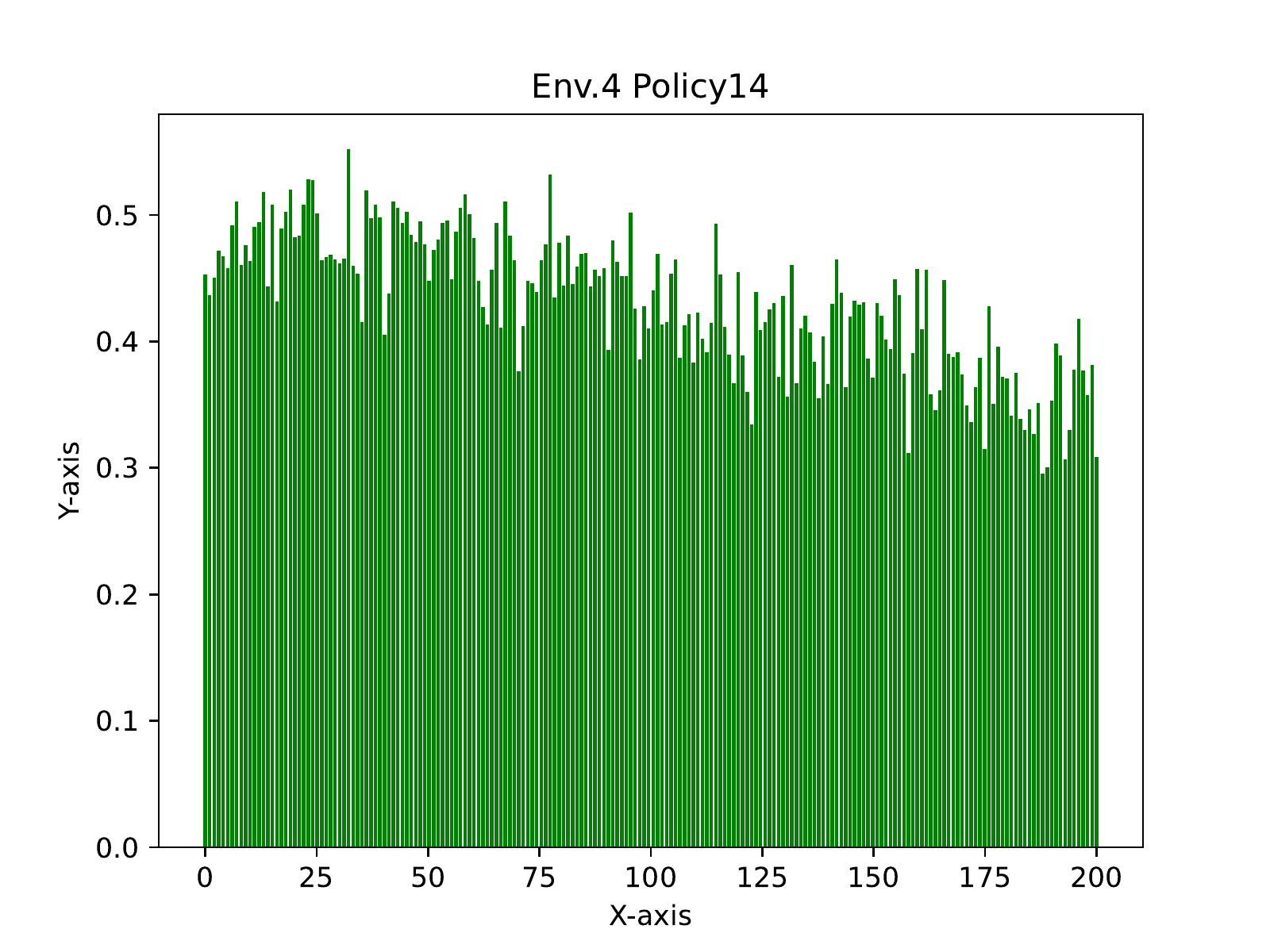}
\includegraphics[width=0.19\textwidth]{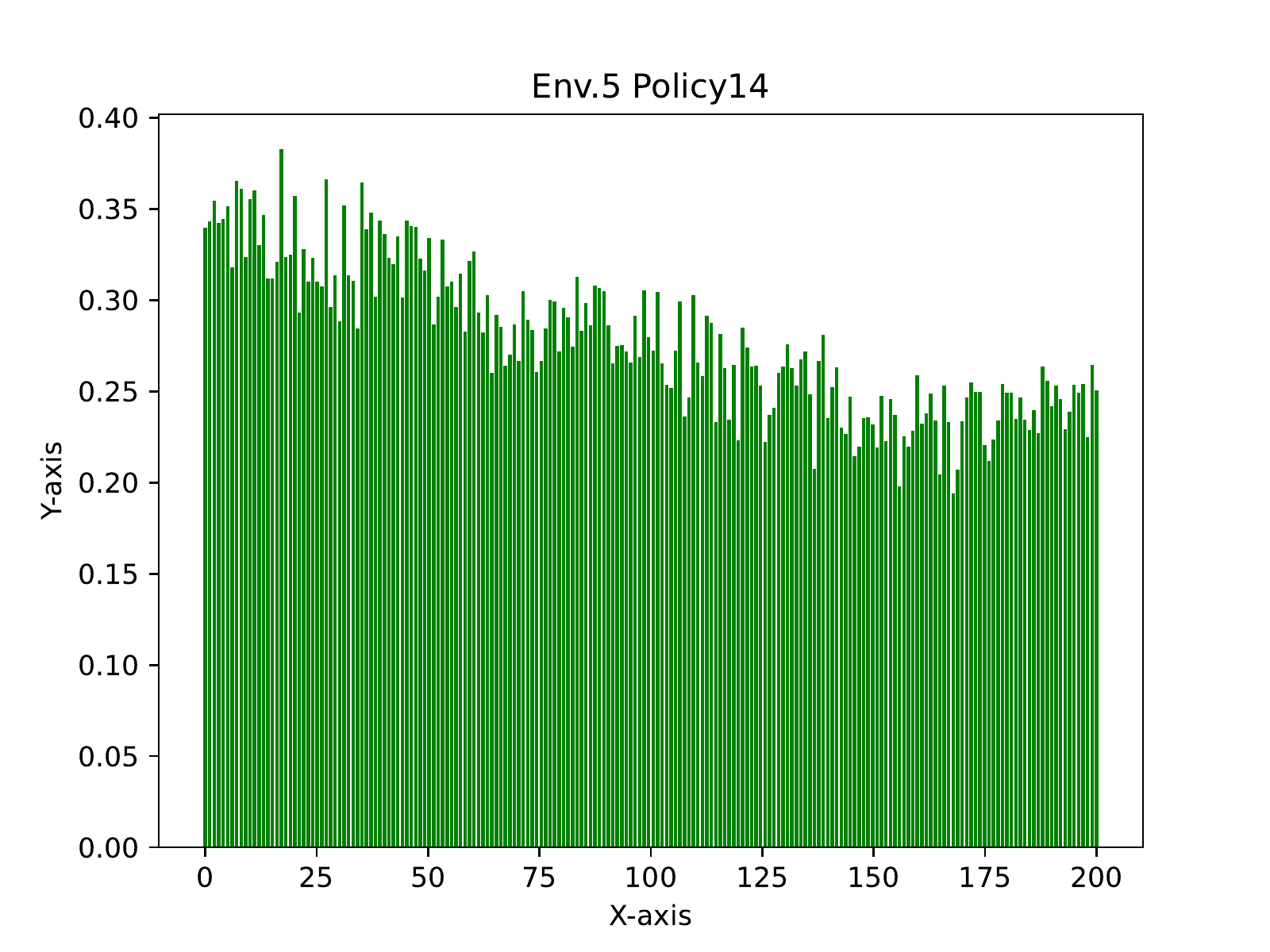}
}
\quad

\subfigure[Policy 15 Env. 1 to Env. 5]{
\includegraphics[width=0.19\textwidth]{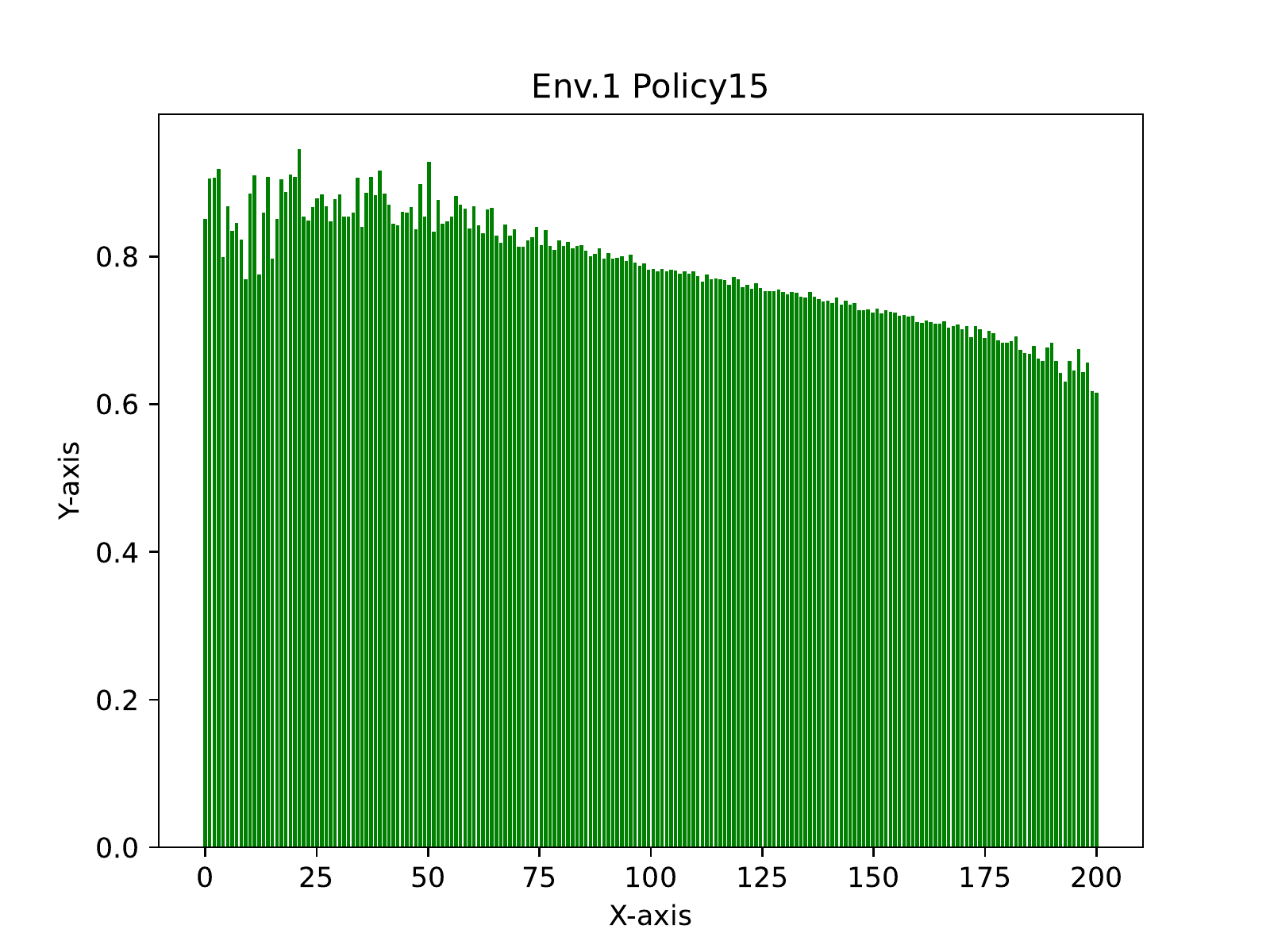}
\includegraphics[width=0.19\textwidth]{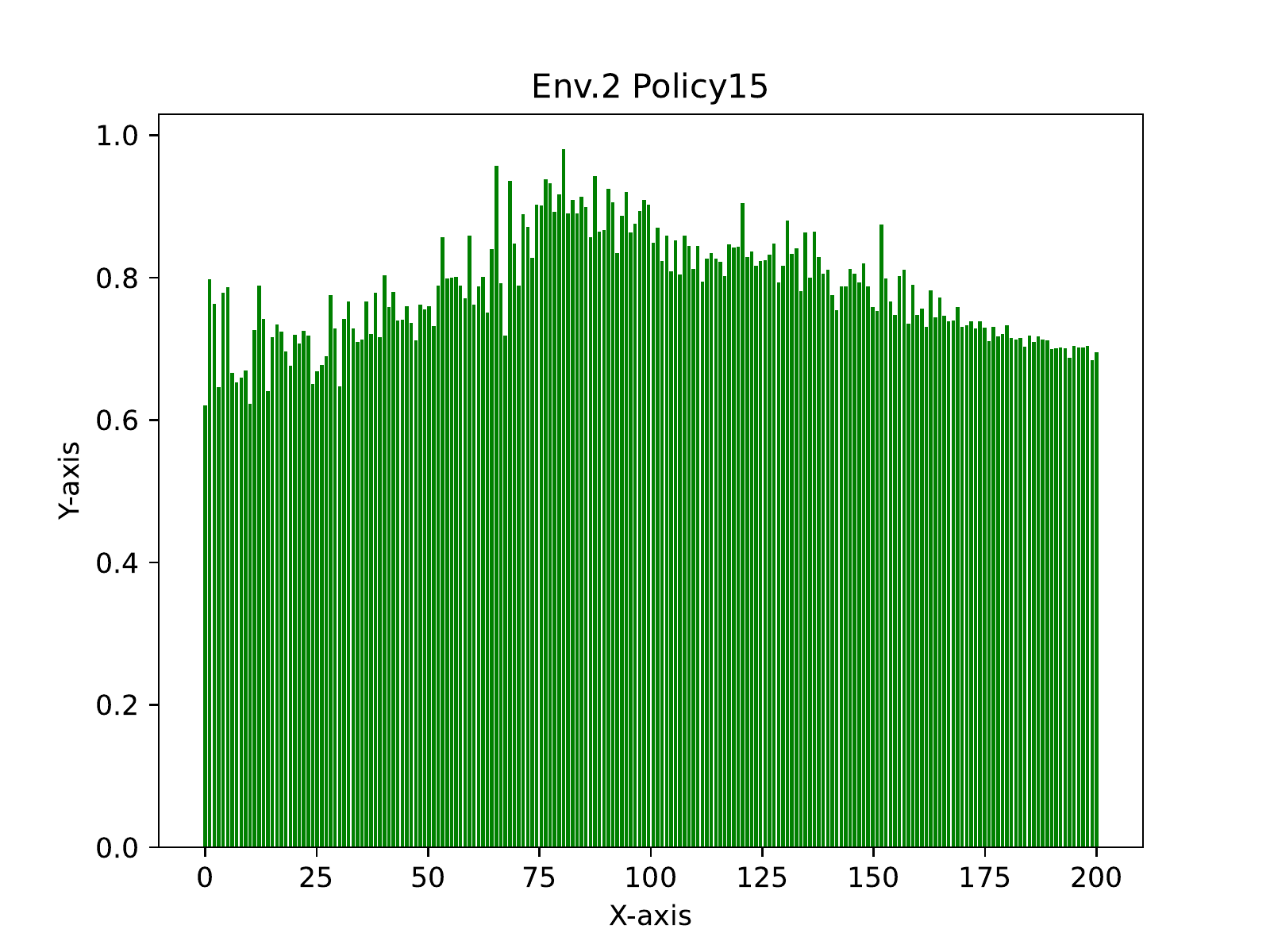}
\includegraphics[width=0.19\textwidth]{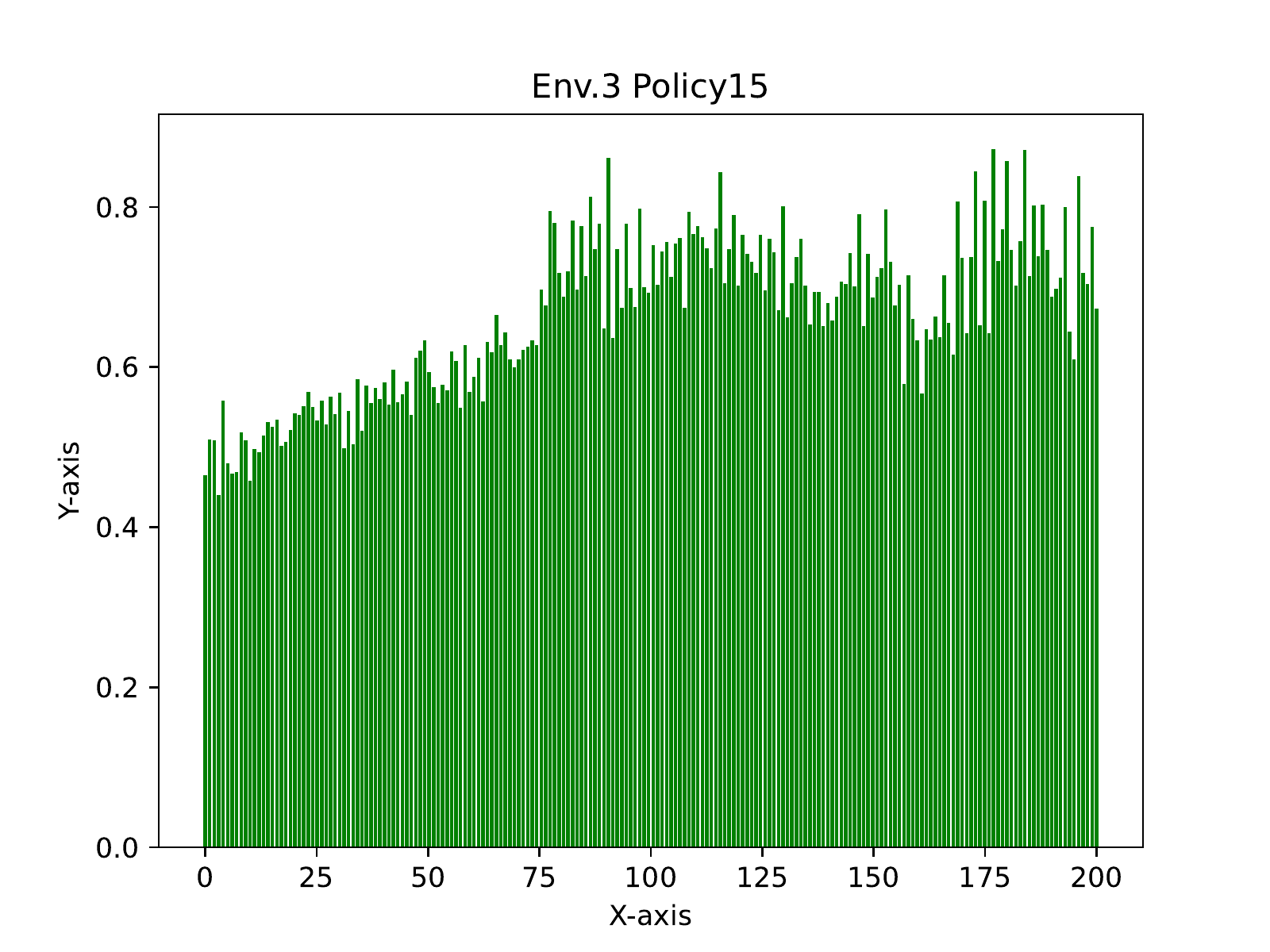}
\includegraphics[width=0.19\textwidth]{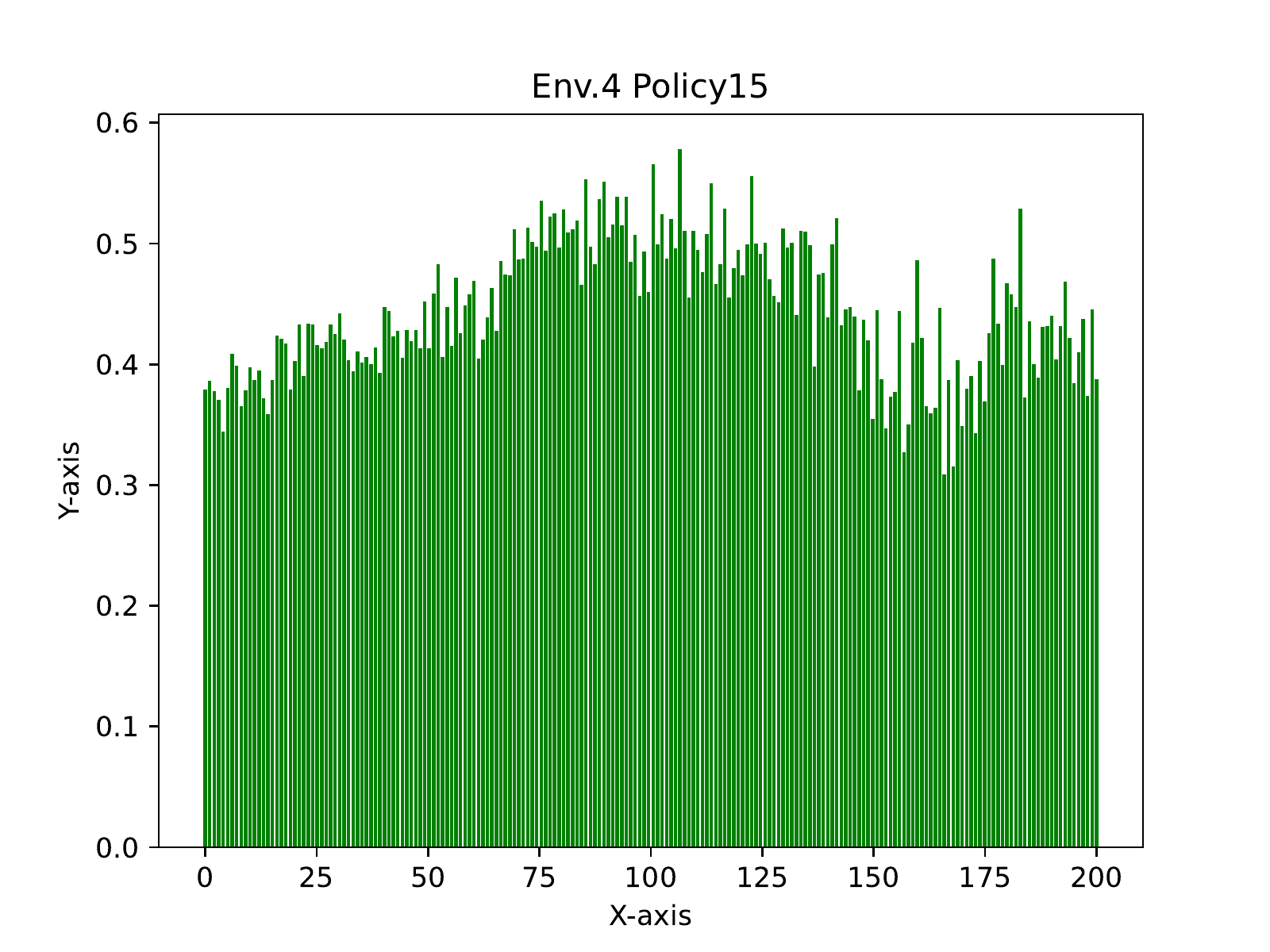}
\includegraphics[width=0.19\textwidth]{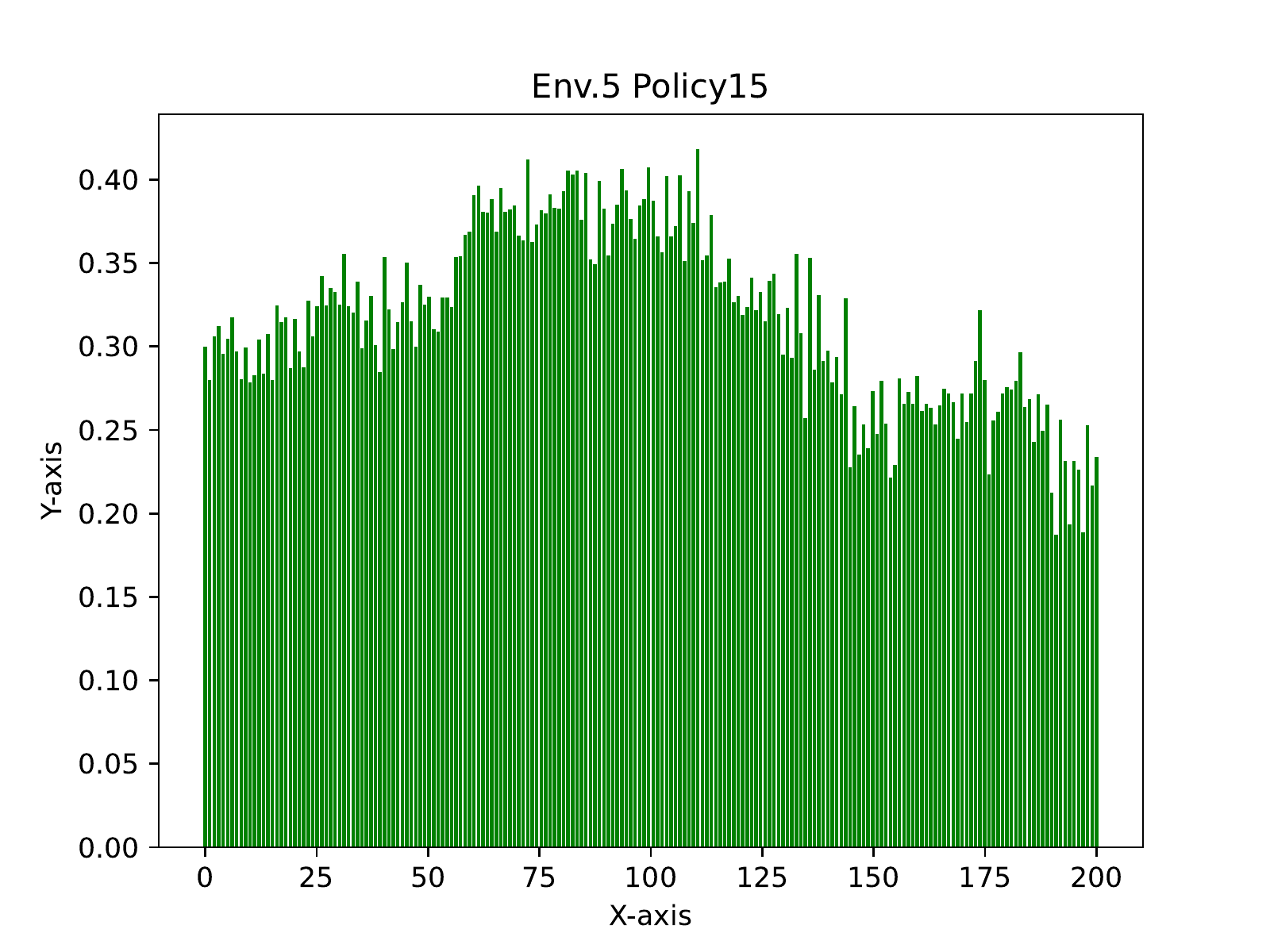}
}
\quad
    
\caption{Episodic MC return distribution of adapted policy during GA process.}
\label{fig::Para Tune Space 3}
\end{figure*}

\end{document}